\documentclass{article}

\usepackage[main, final]{neurips_2026}
\usepackage[utf8]{inputenc} % allow utf-8 input
\usepackage[T1]{fontenc}    % use 8-bit T1 fonts
\usepackage{hyperref}       % hyperlinks
\usepackage{url}            % simple URL typesetting
\usepackage{booktabs}       % professional-quality tables
\usepackage{amsfonts}       % blackboard math symbols
\usepackage{nicefrac}       % compact symbols for 1/2, etc.
\usepackage{microtype}      % microtypography
\usepackage{xcolor}         % colors
\usepackage[table]{xcolor}  % Add to preamble

\usepackage{amsmath}
\usepackage{stmaryrd} 
\usepackage{tikz}
\usetikzlibrary{shapes.geometric, arrows.meta, positioning, calc, backgrounds}
\usepackage{enumitem}

\usepackage{booktabs}   % Pour les jolies lignes (\toprule, \cmidrule, etc.)
\usepackage{multirow}   % Pour fusionner les cellules verticalement
\usepackage{graphicx}   % Pour \rotatebox
\usepackage{makecell}   % Pour les sauts de ligne dans les cellules si besoin
\usepackage{rotating}
\usepackage[labelfont=bf]{caption}

\usepackage{placeins}

\usepackage{threeparttable}
\usepackage[dvipsnames]{xcolor}
\newcommand{\cmark}{\textcolor{ForestGreen}{\checkmark}}
\newcommand{\xmark}{\textcolor{Maroon}{\texttimes}}
\newcommand{\phantomcmid}{\noalign{\vspace{\dimexpr\aboverulesep+\cmidrulewidth+\belowrulesep\relax}}}
\newcommand{\phantommid}{\noalign{\vspace{\dimexpr\aboverulesep+\lightrulewidth+\belowrulesep\relax}}}

\title{HyperNSDE: Personalized Neural SDEs for Joint Static--Longitudinal Clinical Data Generation}

\author{%
  Perrine Chassat 
    \\
  Inria, Université Paris Cité, Inserm, HeKA \\
  F-75015 Paris, France \\
  \texttt{perrine.chassat@inria.fr} \\
  \And
  Agathe Guilloux \\
  Inria, Université Paris Cité, Inserm, HeKA \\
  F-75015 Paris, France \\
  \texttt{agathe.guilloux@inria.fr} \\
}

\begin{document}

\maketitle

\begin{abstract}
Synthetic patient data generation is a promising solution to the dual challenge of data scarcity and privacy constraints in healthcare machine learning.
Realistic synthesis of patient-level clinical data requires jointly modeling heterogeneous static covariates, irregularly sampled longitudinal trajectories, and informative observation times --- three tightly coupled components in practice yet rarely addressed together. We propose \emph{HyperNSDE}, a continuous-time generative model that conditions a latent Neural SDE on static patient representations through a hypernetwork, allowing baseline characteristics to shape trajectory evolution beyond the initial condition without requiring a trajectory encoder, while stochastic latent dynamics capture realistic variability in generated paths. 
Observation times are modeled jointly through a latent-state-dependent intensity process, and training on irregular stochastic paths is stabilized via a deterministic--stochastic path decomposition with a non-adversarial signature-kernel objective. 
Experiments on simulated and real clinical datasets show improved observation-time fidelity and competitive performance, while matched-grid analyses reveal that forecasting and correlation metrics are affected by observation-grid regularity and trajectory smoothness.
% \perrine{Experiments on simulated and real clinical datasets demonstrate consistent improvements over strong baselines across fidelity, privacy, and utility metrics.}
% Experiments on simulated and real clinical datasets show improved observation-time fidelity and strong performance on several distributional and privacy metrics, while forecasting and correlation results are more mixed. Additional matched-grid analyses show that these latter metrics are strongly affected by observation-grid regularity and trajectory smoothness, highlighting the difficulty of evaluating stochastic irregular clinical trajectories.
\end{abstract}

\section{Introduction}

Clinical cohorts are expensive to collect, difficult to share due to privacy regulations, and often too small to support reliable machine learning models~\cite{chen2021synthetic}. Yet many healthcare applications require realistic patient-level data at scale: benchmarking predictive methods, simulating trial populations, studying rare subgroups, or augmenting underrepresented cohorts~\cite{thorlund2020synthetic, chen2021synthetic, peltner2025eu}. Synthetic data generation is therefore a promising approach to this concrete clinical challenge, provided that generated samples preserve the multimodal structure, heterogeneity, and variability of real patient records.
% Synthetic patient data generation offers a principled solution: by learning the joint distribution of patient-level records, a generative model can support privacy-aware data sharing, data augmentation, and clinical trial simulation~\cite{thorlund2020synthetic, chen2021synthetic, peltner2025eu}.

% Clinical cohorts are expensive to collect, difficult to share due to privacy regulations and governance requirements, and often too small to support reliable machine learning models. Synthetic patient data generation offers a promising way to mitigate these limitations by supporting privacy-aware data sharing, data augmentation, and clinical trial simulation \perrine{refs}. \agathe{tu as besoin d'aide pour ces ref ?} \agathe{s'il y a la place, je pense qu'il faut un peu insister ici}

The structure of clinical data makes this a fundamentally challenging generative modeling problem, going beyond the generic time-series generation. Clinical records combine heterogeneous static baseline covariates --- demographics, comorbidities, treatment history --- with longitudinal measurements collected over irregular, patient-specific time grids. 
% These components are not independent: baseline characteristics influence disease progression and the nature of temporal dynamics, while observation times may themselves be informative of patient health status (i.e., informative sampling and censoring).
These components are not independent: baseline characteristics influence disease progression, observation times are often informative of the underlying health state, and part of the trajectory variability can reflect latent or unmeasured factors that are not fully explained by observed covariates. 
A realistic generative model should therefore capture the joint distribution of static features, longitudinal trajectories, and observation patterns, while accounting for both cross-subject heterogeneity and residual temporal variability.

Existing approaches only partially address this setting. Most generative models for time series operate in discrete time and assume regularly sampled sequences~\cite{yoon2019time,esteban2017real,li2023generating,nicholas2023synthetic}, making them ill-suited to irregular clinical trajectories. Continuous-time models~\cite{rubanova2019latent,chen2018neural} provide a natural framework for irregular data, but generally do not address joint static--longitudinal generation. The closest prior work in this setting, MultiNODEs~\cite{wendland2022generation}, combines a variational model for static covariates with a latent Neural ODE. However, static information affects the model mainly through the initial latent state rather than the full dynamics, latent initialization depends on the full observed trajectory rather than only baseline information, the latent dynamics remain deterministic, and informative observation times are not modeled. These limitations motivate a different design for realistic synthetic clinical trajectory generation.

We propose \textbf{HyperNSDE} (\textbf{Hyper}network \textbf{N}eural \textbf{SDE}), a unified continuous-time generative model that addresses these limitations jointly,
by combining a generative model for static covariates with a personalized Neural SDE whose dynamics are conditioned on the patient’s static latent representation through a hypernetwork.  
% allowing baseline characteristics to modulate the full trajectory dynamics rather than only the initial condition, without requiring a trajectory encoder. 
Observation times are modeled through a latent-state-dependent intensity process, and training is stabilized via a deterministic--stochastic path decomposition with a signature-kernel objective. 
% We evaluate HyperNSDE on simulated and real clinical datasets and show consistent improvements over strong baselines across fidelity, privacy, and utility metrics. 
% Across simulated and real clinical datasets, HyperNSDE improves observation-time fidelity and strong performance on several distributional and privacy metrics, while performance on forecasting and correlation metrics is more heterogeneous; we investigate the source of these differences through controlled matched-grid analyses.
Across simulated and real clinical datasets, HyperNSDE improves observation-time fidelity and several distributional and privacy metrics, while forecasting and correlation performance is more heterogeneous; controlled matched-grid analyses investigate these differences.

\textbf{Contributions}
\begin{itemize}[leftmargin=0.5cm]
    \item \textbf{Unified generative model for clinical data:} HyperNSDE is a holistic continuous-time model for the joint generation of heterogeneous static covariates, irregular longitudinal trajectories, and informative observation times, without requiring trajectory encoding or pre-imputation.
    \item \textbf{Stochastic latent dynamics for realistic trajectory variability:} We use stochastic latent dynamics, capturing an unmeasured source of variability not explained by baseline covariates alone and yielding richer and more realistic trajectory diversity than deterministic latent dynamics. For stable training on irregular paths, we introduce a deterministic--stochastic path decomposition optimized via a non-adversarial signature-kernel-based training objective.
    % We use stochastic latent dynamics, 
    % % enabling the model to capture intrinsic temporal uncertainty and richer inter-individual variability.  
    % allowing the model to capture an unmeasured latent source of variability not explained by baseline covariates alone, and yields richer, more realistic trajectory diversity than deterministic latent dynamics.
    % For stable training on irregular paths, we introduce a deterministic--stochastic path decomposition optimized via a non-adversarial signature-kernel-based training objective.  
    \item \textbf{Hypernetwork-based personalization of latent dynamics:} 
    % We propose a hypernetwork-conditioned latent Neural SDE that allows baseline attributes to modulate the entire temporal vector field and enables patient-specific progression rates rather than just shifting initial conditions.
    % We condition the latent dynamics on static patient representations through a hypernetwork, so that baseline
    We propose a hypernetwork-conditioned latent Neural SDE that allows baseline
    covariates to modulate the full temporal vector field rather than only the initial condition. This explicitly links cross-subject heterogeneity in baseline profiles to variability in longitudinal evolution.
    % \item \textbf{Modeling informative sampling:} We explicitly model observation patterns as a latent-state-dependent intensity process. This captures the "informative sampling" bias inherent in healthcare records.
\end{itemize}

\section{Background and related work}
\label{sec:background}

Table~\ref{tab:model_comparison} compares representative generative models along five properties central to our setting: generation of static covariates, support for irregular time series, continuous-time latent dynamics, joint modeling of static and longitudinal variables, and explicit modeling of the observation intensity.

\begin{table*}[h]
\centering
\footnotesize
\caption{\textbf{Comparison of generative models for joint static and irregular longitudinal data.} $^\text{1}$~Irregular timing handled externally. $^\text{2}$~Static covariates conditioned on, not generated. $^\text{3}$~Static variables represented as time-series channels.}
\label{tab:model_comparison}
\definecolor{rowgray}{gray}{0.9}
\setlength{\tabcolsep}{5pt}
\setlength{\heavyrulewidth}{1pt}
\setlength{\lightrulewidth}{1pt}
% \begin{threeparttable}
\begin{tabular}{lccccc}
\toprule
\multirow{2}{*}{\textbf{Model}} & \textbf{Static} & \textbf{Irregular} & \textbf{Continuous} 
    & \textbf{Joint modeling} & \textbf{Modeled} \\
\noalign{\vspace{-1pt}}
 & \textbf{data} & \textbf{time-series} & \textbf{latent} & \textbf{static/TS} 
    & \textbf{intensity} \\
\midrule
TimeGAN \cite{yoon2019time}               & \xmark & \xmark & \xmark & \xmark & \xmark \\
\rowcolor{rowgray}
EHR-M-GAN \cite{li2023generating}        & \xmark & \xmark & \xmark & \xmark & \xmark \\
TimeDiff \cite{tian2024reliable}
& \xmark\footnotemark[3] & \xmark & \xmark & \xmark\footnotemark[3] & \xmark \\
\rowcolor{rowgray}
EHR-Safe \cite{yoon2023ehr}              & \cmark & \cmark & \xmark & \cmark & \xmark \\
% \rowcolor{rowgray}
LatentODE \cite{rubanova2019latent}      & \xmark & \cmark & \cmark & \xmark & \cmark \\
\rowcolor{rowgray}
VAMBN \cite{kuhnel2024synthetic}         & \cmark & \cmark\footnotemark[1] & \xmark & \cmark & \xmark \\
% \rowcolor{rowgray}
HL-VAE \cite{ougretir2022variational}    & \xmark\footnotemark[2] & \cmark & \cmark & \xmark & \xmark \\
\rowcolor{rowgray}
DGBFGP \cite{balik2025bayesian}          & \xmark\footnotemark[2] & \cmark & \cmark & \xmark & \xmark \\
% \rowcolor{rowgray}
RTSGAN \cite{pei2021rtsgan}              & \cmark & \cmark & \xmark & \cmark & \cmark \\
\rowcolor{rowgray}
MultiNODEs \cite{wendland2022generation} & \cmark & \cmark & \cmark & \cmark & \xmark \\
\cmidrule(lr){1-6}
% \rowcolor{rowgray}
\textbf{Ours} & \textbf{\cmark} & \textbf{\cmark} & \textbf{\cmark} & \textbf{\cmark} 
    & \textbf{\cmark} \\ [-1pt]
\bottomrule
\end{tabular}
% \begin{tablenotes}
%     \footnotesize
%     \item[1] Treats longitudinal data as ordered discrete sequences; irregular timing must 
%     be handled externally via binning or as a raw input feature.
%     \item[2] Uses static covariates as conditional inputs but does not provide a generative 
%     distribution for the static features themselves.
% \end{tablenotes}
% \end{threeparttable}
\end{table*}

\paragraph{Time-series and joint generative models}
Most generative models for time series target regularly sampled sequences in discrete time~\cite{yoon2019time,esteban2017real,li2023generating,nicholas2023synthetic,yuan2024diffusion}. While effective on standard benchmarks, such models are ill-suited to clinical trajectories without resampling or imputation. Methods designed for irregular time series~\cite{tian2024reliable,kim2024synthesis,jeon2022gt} better address the temporal structure, but typically do not jointly model static covariates and observation intensities. Among joint models, RTSGAN~\cite{pei2021rtsgan} incorporates static covariates and visit-time modeling in an adversarial framework, while HL-VAE~\cite{ougretir2022variational}, VAMBN~\cite{kuhnel2024synthetic}, and EHR-Safe~\cite{yoon2023ehr} combine heterogeneous static data with longitudinal modeling in discrete-time settings. MultiNODEs~\cite{wendland2022generation} is the closest prior work in the continuous-time setting.

\paragraph{Continuous-time latent variable models and Neural SDEs}
Neural ODEs~\cite{chen2018neural} parameterize deterministic latent dynamics through a neural vector field, and Latent ODEs~\cite{rubanova2019latent} embed this idea in a generative framework for irregularly sampled data. Neural SDEs~\cite{kidger2021neural} extend these models by introducing stochastic latent dynamics through learned drift and diffusion terms, allowing the latent process to capture intrinsic temporal uncertainty and to generate richer trajectory variability through the sampled Brownian motion. Training Neural SDEs remains challenging; proposed objectives include variational formulations~\cite{li2020scalable}, adversarial Wasserstein objectives~\cite{kidger2021efficient}, scoring rules~\cite{zhang2024efficient}, and signature-kernel scores on path space~\cite{Issa2023,salvi2021signature,bonnier2025proper}.

\paragraph{Hypernetworks}
Hypernetworks~\cite{ha2016hypernetworks} generate the parameters of a target network as a function of a conditioning input, providing a flexible individualization mechanism beyond concatenation or feature-wise modulation. They have been applied to personalized modeling in scientific and clinical settings~\cite{de2021hyperpinn,belogolovsky2023individualized}. 
% \perrine{Peut-être enlever ça: In our model, a hypernetwork conditioned on static latent covariates generates patient-specific SDE parameters, allowing baseline characteristics to shape the full trajectory dynamics.}

\section{Model}

\begin{figure}[h]
    \centering
    \includegraphics[width=1\linewidth]{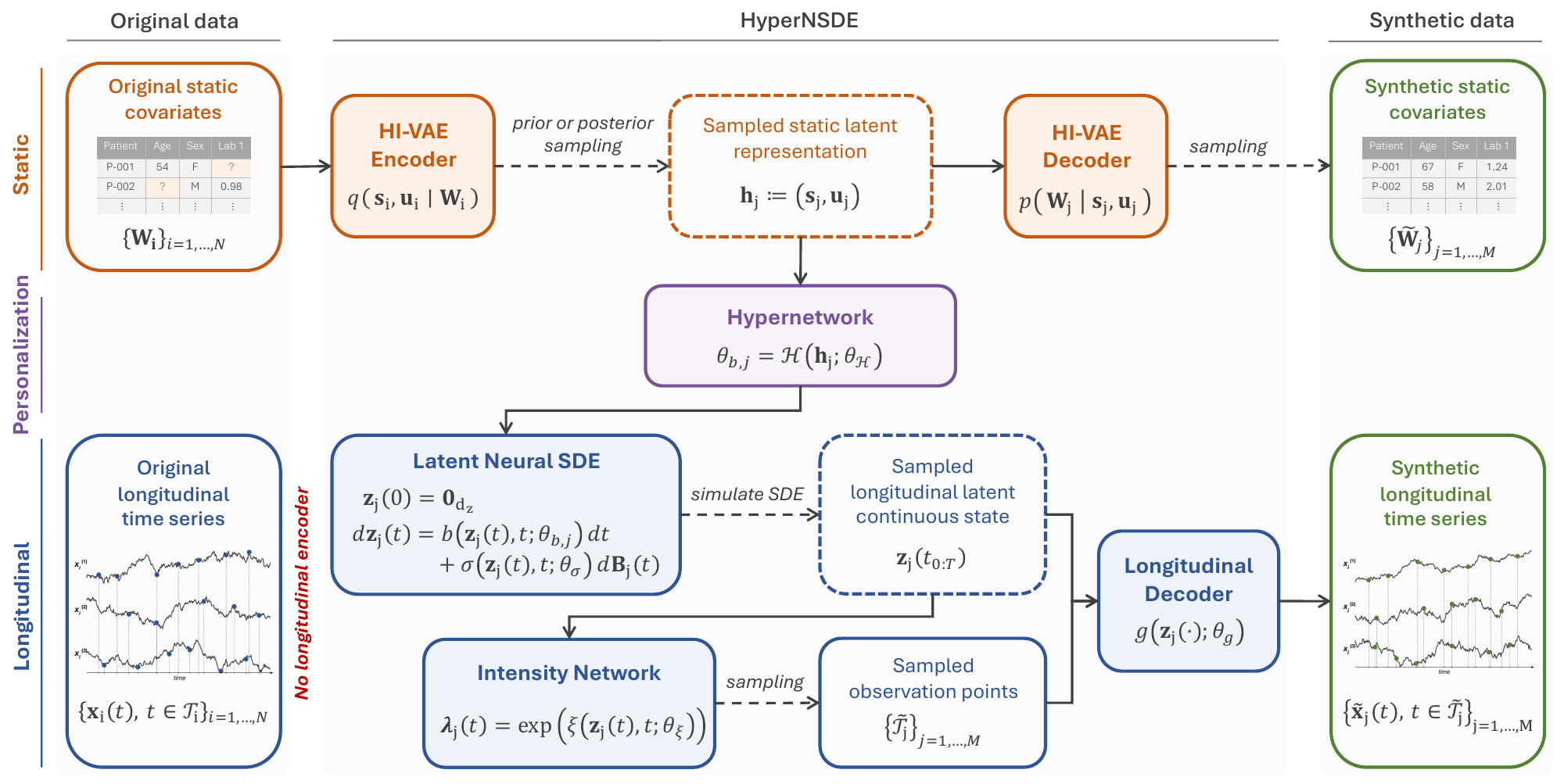}
    \caption{\textbf{HyperNSDE generation pipeline.}
Generating a synthetic subject follows a hierarchical process (Appendix~\ref{app:synthetic_generation}). Static latent representations are sampled from the HI-VAE learned distributions, decoded into synthetic covariates, and passed through the hypernetwork to generate patient-specific SDE parameters. A latent path is then simulated from the personalized Neural SDE and decoded at observation times sampled from the intensity network to produce synthetic irregular longitudinal trajectories. Original longitudinal data are used only at training time via the loss terms, without relying on a longitudinal encoder. Solid arrows denote deterministic computations; dashed arrows denote stochastic steps.}
    \label{fig:model_HyperNSDE}
\end{figure}

\paragraph{Problem formulation}
We consider a cohort of \(N\) subjects. For each subject \(i\), we observe a vector of mixed-type static covariates \(\mathbf{W}_i \in \mathbb{R}^{d_W}\) measured at baseline (\(t=0\)), encoding time-invariant characteristics (such as age, sex...). These covariates may be continuous, positive-valued, categorical, or ordinal.
We also observe \(d_\mathrm{x}\) irregularly sampled and partially observed longitudinal variables, 
\[
\mathbf{x}_i = \bigl\{\mathbf{x}_i^{(1)}, \dots, \mathbf{x}_i^{(d_\mathrm{x})}\bigr\},
\]
where each variable \(\mathbf{x}_i^{(\ell)}\) is recorded on a subject and variable-specific set of time points \(\mathcal{T}_i^{(\ell)} \subset [0,T_i]\). Here, \(T_i \leq T\) denotes the subject-specific last observation time, and \(T\) the overall study horizon.

\subsection{Static latent representation with HI-VAE} \label{sec:static_HIVAE}

To model heterogeneous and incomplete static covariates \(\mathbf{W}_i\), we use a Heterogeneous-Incomplete VAE (HI-VAE) \cite{Nazabal2020}, following the joint modeling strategy of VAMBN \cite{kuhnel2024synthetic} and MultiNODEs \cite{wendland2022generation}. In our framework, this component both provides a generative model for static covariates and a compact latent representation used to condition the longitudinal dynamics.
For each individual \(i\), the HI-VAE introduces a discrete latent variable \(\mathbf{s}_i \in \{0,1\}^L\) and a continuous latent variable \(\mathbf{u}_i \in \mathbb{R}^{d_u}\), with priors
\(p(\mathbf{s}_i)=\mathrm{Cat}(\mathbf{s}_i\mid \boldsymbol{\pi})\) and \(p(\mathbf{u}_i \mid \mathbf{s}_i)=\mathcal{N} \! \bigl(\mathbf{u}_i \mid \boldsymbol{\mu}_p(\mathbf{s}_i), \mathbf{I}\bigr)\).
The decoder uses covariate type-specific likelihoods (e.g., Gaussian, Poisson, or Categorical) to handle heterogeneous variables and missing values. We denote the resulting static latent representation by
\[
\mathbf{h}_i := (\mathbf{s}_i,\mathbf{u}_i),
\]
which is used to condition the longitudinal model. Additional details are provided in Appendix~\ref{app:details_hivae}.

\subsection{Personalized stochastic latent dynamics}

We model the irregular longitudinal trajectories \(\{\mathbf{x}_i(t)\}_{t\in[0,T]}\) conditionally on subject-specific static information summarized by the latent representation \(\mathbf{h}_i=(\mathbf{s}_i,\mathbf{u}_i)\).

\paragraph{Latent stochastic dynamics}
Following a standard latent-variable formulation for continuous-time generative modeling \cite{Kingma2013AutoEncodingVB,murphy2012machine}, we assume that the observed trajectories \(\mathbf{x}_i(t)\in\mathbb{R}^{d_\mathrm{x}}\) arise from a continuous-time latent process \(\mathbf{z}_i(t)\in\mathbb{R}^{d_\mathrm{z}}\), decoded through
\begin{equation}
\hat{\mathbf{x}}_i(t)=g\bigl(\mathbf{z}_i(t);\theta_g\bigr),
\label{eq:obs_decoder}
\end{equation}
where \(g\) is a feed-forward neural network. Since \(\mathbf{z}_i(t)\) is defined in continuous time, the decoder can be evaluated at any observation time \(t\), which allows both reconstruction and generation on arbitrary irregular time grids. 
We model the latent dynamics of each subject \(i\) with a Neural SDE \cite{tzen2019neural,jia2019neural,liu2019neural,kidger2021efficient}:
\begin{equation}
\begin{cases}
d\mathbf{z}_i(t)
=
b_i\bigl(\mathbf{z}_i(t),t\bigr)\,dt
+
\sigma\bigl(\mathbf{z}_i(t),t;\theta_\sigma\bigr)\,d\mathbf{B}_i(t),\\
\, \mathbf{z}_i(0)=\mathbf{0}_{d_\mathrm{z}},
\end{cases}
\label{eq:latent_sde}
\end{equation}
% for each subject \(i\), 
where \(\mathbf{B}_i\) is a standard Brownian motion. The drift \(b_i\) and diffusion \(\sigma\) are neural networks that model, respectively, subject-specific deterministic dynamics and stochastic variability shared across subjects. We assume standard global Lipschitz conditions to ensure well-posedness \cite{oksendal2003stochastic}.
Equation \eqref{eq:latent_sde} provides the conceptual latent-SDE formulation that motivates our model, but it is not trained directly. Instead, we use the deterministic–stochastic decomposition described in Section~\ref{sec:training} and Appendix~\ref{app:details_splitSDEtrain} to separate subject-specific trends from residual stochastic variability.

\paragraph{Personalization via hypernetworks} To personalize the dynamics, we let the static representation \(\mathbf{h}_i\) parameterize the drift via a hypernetwork \(\mathcal H\) \cite{ha2016hypernetworks}:
\begin{equation}
\theta_{b,i}=\mathcal H(\mathbf{h}_i;\theta_{\mathcal H}),
\qquad
b_i(\cdot,\cdot)=b(\cdot,\cdot;\theta_{b,i}).
\label{eq:hypernetwork}
\end{equation}
The hypernetwork maps each patient's static representation to a subject-specific drift parameter $\theta_{b,i}$, yielding individualized vector fields while sharing the hypernetwork weights $\theta_{\mathcal{H}}$ across subjects. The diffusion coefficient $\sigma$ is kept population-shared, separating subject-specific trajectory trends from population-level stochasticity.

Unlike standard latent SDE formulations that infer a distribution over the initial latent state, we fix \(\mathbf{z}_i(0)=\mathbf{0}\) for all subjects. Since baseline measurements are included in the static input \(\mathbf{W}_i\) (see Appendix~\ref{app:preprocessing}) and encoded in \(\mathbf{h}_i\), absolute signal levels are absorbed by the static representation, and \(\mathbf{z}_i(t)\) models deviations from baseline driven by the personalized drift \(b_i\) .

\subsection{Event intensity from the latent path}

In many longitudinal settings, observation times are informative of the underlying state: for instance, patients may be monitored more frequently during periods of deterioration. Following \cite{rubanova2019latent}, we model the observation process as a latent-state-dependent multivariate point process, and we additionally introduce an explicit end-of-sequence (EOS) event. For subject $i$, given latent trajectory $\mathbf{z}_i(t)$, we define
\begin{equation}
\boldsymbol{\lambda}_i(t)
=
\bigl(\lambda_i^{(1)}(t), \dots, \lambda_i^{(d_\mathrm{x})}(t), \lambda_i^{\mathrm{EOS}}(t)\bigr)
\in \mathbb{R}_+^{d_\mathrm{x}+1},
\qquad
\boldsymbol{\lambda}_i(t)=\exp\bigl(\xi(\mathbf{z}_i(t),t;\theta_\xi)\bigr),
\label{eq:event_intensity}
\end{equation}
where \(\xi\) is a neural network with parameters \(\theta_\xi\), and the exponential is applied componentwise. Here, \(\lambda_i^{(\ell)}(t)\) denotes the intensity of observation times for variable \(\ell \in \{1,\dots,d_\mathrm{x}\}\), while \(\lambda_i^{\mathrm{EOS}}(t)\) denotes the hazard that the observation process terminates at time \(t\).

\section{Training procedures}
\label{sec:training}

Training HyperNSDE requires jointly optimizing all interconnected components of the models --- the HI-VAE, hypernetwork, latent dynamics, decoder, and intensity model. Joint training is essential: the static latent representation $\mathbf{h}_i$ must align with the requirements of the hypernetwork $\mathcal{H}$, and independent pretraining of the HI-VAE yields embeddings that fail to capture the features necessary to parameterize subject-specific drift functions, leading to suboptimal personalized dynamics.

% \subsection{Stable training via deterministic--stochastic path decomposition}
\paragraph{Stable training via deterministic--stochastic path decomposition}
In our setting with irregular and sparse longitudinal data, directly training a single latent Neural SDE to capture both smooth trajectory trends and stochastic fluctuations proved unstable. We therefore introduce the following structured latent decomposition:
\begin{equation}
\mathbf{z}_i(t)=\mathbf{z}_i^{\mathrm{det}}(t)+\mathbf{z}_i^{\mathrm{sto}}(t),
\qquad
\mathbf{z}_i^{\mathrm{det}}(0)=\mathbf{z}_i^{\mathrm{sto}}(0)=\mathbf{0}.
\label{eq:latent_decomposition}
\end{equation}
The deterministic component \(\mathbf{z}_i^{\mathrm{det}}\) captures the subject-specific smooth trend and solves a personalized Neural ODE with drift \(b_i^{\mathrm{det}}\), whose parameters are generated by the hypernetwork, \(\theta_{b_i^{\mathrm{det}}}=\mathcal{H}(\mathbf{h}_i;\theta_{\mathcal H})\). The stochastic component \(\mathbf{z}_i^{\mathrm{sto}}\) captures residual temporal variability through a population-shared Neural SDE with drift \(b^{\mathrm{sto}}\) and diffusion \(\sigma^{\mathrm{sto}}\).
For additional training stability, the intensity network $\xi(\cdot, t;\theta_\xi)$ is conditioned only on $\mathbf{z}_i^{\mathrm{det}}(t)$ rather than on the full decomposed latent state
$\mathbf{z}_i(t)$. 
This decomposition is a modeling choice: it provides a structured separation between smooth dynamics and residual stochasticity, but is not, in general, equivalent to the single latent SDE in Eq.~\eqref{eq:latent_sde}. Empirically, this parameterization improves training stability and the identification of static--longitudinal dependencies, as confirmed by the ablation studies in Table~\ref{tab:ablation_studies} and Figure~\ref{fig:simu_ablation_studies}. Further discussion is provided in Appendix~\ref{app:discussion_splitSDEtrain}.

% \subsection{Loss terms}
\paragraph{Loss terms}

The model is trained by minimizing a weighted combination of the four loss terms defined below. Since they differ substantially in scale and nature, we use a convex combination with weights tuned jointly with the other hyperparameters. Details on the weighting strategy and Pareto-based hyperparameter optimization are provided in Appendix~\ref{app:hyperopt}.

\begin{itemize}[leftmargin=0.5cm]
    \item \textbf{Static ELBO:}
The HI-VAE is trained with the standard evidence lower bound \cite{Kingma2013AutoEncodingVB, Nazabal2020}, computed on observed static entries only:
\begin{equation}
\begin{aligned}
\mathcal{L}_{\mathrm{static}}
&=
\sum_{i=1}^N
\mathbb{E}_{q(\mathbf{u}_i,\mathbf{s}_i \mid \mathbf{W}_i)}
\bigl[
\log p(\mathbf{W}_i \mid \mathbf{u}_i,\mathbf{s}_i)
\bigr] \\
&\qquad
-
\mathbb{E}_{q(\mathbf{s}_i \mid \mathbf{W}_i)}
\!\left[
\mathrm{KL}\!\left(q(\mathbf{u}_i \mid \mathbf{W}_i,\mathbf{s}_i)\,\|\,p(\mathbf{u}_i \mid \mathbf{s}_i)\right)
\right]
-
\mathrm{KL}\!\left(q(\mathbf{s}_i \mid \mathbf{W}_i)\,\|\,p(\mathbf{s}_i)\right).
\end{aligned}
\label{eq:loss_static}
\end{equation}
\item \textbf{Point-process likelihood:}
% For subject \(i\) and variable \(\ell\), \(\mathcal{T}_i^{(\ell)} \subset [0,T_i]\) denote the set of observation times of variable \(\ell\). 
Conditioned on the latent path (Eq.~\ref{eq:event_intensity}), the event loss consists of two log-likelihood terms: one for observation times and one for the end-of-sequence event at $T_i$:
\begin{equation}
\mathcal{L}_{\mathrm{event}}\!
=
-\frac{1}{N}\!\sum_{i=1}^N
\!\left[
\sum_{\ell=1}^{d_x} \! \left( \!
\sum_{t \in \mathcal{T}_i^{(\ell)}} \! \log \lambda_i^{(\ell)}\!(t) \!
- \!
\int_0^{T_i}\!\lambda_i^{(\ell)}\!(s)ds\!
\right) \!
+
\log \lambda_i^{\mathrm{EOS}}\!(T_i) \!
-\!
\int_0^{T_i}\!\lambda_i^{\mathrm{EOS}}\!(s)ds 
\right]\!.
\label{eq:loss_event}
\end{equation}
In practice, the integral terms are approximated numerically on the discretization grid used to solve the latent dynamics (see Appendix~\ref{app:details_eventmodel}) \cite{daley2003introduction, rubanova2019latent}. 
% \agathe{on ne voit pas la dependance avec l'état latent comme dans les autres bouts} 
\item \textbf{Deterministic reconstruction loss:}
The deterministic latent part is trained using the mean squared error to reconstruct observed longitudinal values:
\begin{equation}
\mathcal{L}_{\mathrm{long,det}}
=
\frac{1}{N}
\sum_{i=1}^N
\sum_{\ell=1}^{d_x}
\sum_{t \in \mathcal{T}_i^{(\ell)}}
\left\|
\mathbf{x}_i^{(\ell)}(t)
-
\hat{\mathbf{x}}_{i,\mathrm{det}}^{(\ell)}(t)
\right\|_2^2,
\qquad
\hat{\mathbf{x}}_{i,\mathrm{det}}(t)
=
g \bigl(\mathbf{z}_i^{\mathrm{det}}(t);\theta_g\bigr).
\label{eq:loss_det}
\end{equation}
\item \textbf{Stochastic residual loss:}
To train the residual stochastic component, we match reconstructed residual trajectories to empirical residuals using the \emph{signature kernel score} \cite{Issa2023,salvi2021signature}. We denote the empirical residual trajectory and the reconstructed stochastic residual by
\[
\mathbf{r}_i(t)
=
\mathbf{x}_i(t)-g\bigl(\mathbf{z}_i^{\mathrm{det}}(t);\theta_g\bigr),
\qquad
\hat{\mathbf{r}}_i(t)
=
g\bigl(\mathbf{z}_i^{\mathrm{sto}}(t);\theta_g\bigr).
\]
We then define
\begin{equation}
\mathcal{L}_{\mathrm{long,sto}}
=
\widehat{\phi}_{\mathrm{sig}}
\Bigl(
\{\hat{\mathbf{r}}_i\}_{i=1}^N,
\{\mathbf{r}_i\}_{i=1}^N
\Bigr),
\label{eq:loss_sto}
\end{equation}
where \(\widehat{\phi}_{\mathrm{sig}}\) denotes the empirical signature kernel score associated with a (RBF-)signature kernel on path space. This score is well-suited to irregularly sampled trajectories and is evaluated in practice through its mini-batch estimator. Further details are given in Appendix~\ref{app:details_splitSDEtrain}.
\end{itemize}

\section{Evaluation methodology}

% \subsection{Evaluation metrics}

% Evaluating synthetic multimodal clinical data is challenging, and no universally accepted protocol exists, especially for irregularly sampled longitudinal data with missingness and event-driven observation patterns

\paragraph{Evaluation metrics} No universally accepted evaluation protocol exists for irregularly sampled longitudinal data with missingness~\cite{stenger2024evaluation, stenger2025steb, achterberg2024evaluation, kim2024synthesis, loni2025review}, and most existing benchmarks target regularly sampled or tabular data, primarily assess fidelity and utility while overlooking privacy, and rely on interpolation that can distort temporal structure and bias evaluation. 
We assess performance along three complementary axes --- \emph{fidelity}, \emph{utility}, and \emph{privacy} --- using metrics adapted to our multimodal and irregular setting. Full definitions, extended discussion, and limitations are provided in Appendix~\ref{app:metrics} and ~\ref{app:metric_diagnostics}.

\begin{itemize}[leftmargin=0.5cm]
    \item \textbf{Fidelity} assesses whether synthetic samples reproduce the statistical structure of real data while maintaining diversity. We report modality-adapted \emph{MMD}~\cite{gretton2012kernel} and \emph{correlation RMSE (Corr. RMSE)} for both static and longitudinal data, using signature-kernel MMD~\cite{chevyrev2022signature, salvi2021signature} and masked correlation matrices for longitudinal observations without interpolation. We additionally report the \emph{KL divergence on the observation intensities (KL $\lambda$)} and a \emph{discriminative score (Discrim.)}~\cite{yoon2019time} from real-versus-synthetic classification.
    \item \textbf{Utility} measures whether synthetic data can substitute for real data in downstream tasks. We evaluate it through a forecasting task under a Train-on-Synthetic-Test-on-Real (TSTR) protocol~\cite{esteban2017real}: a predictor trained on static covariates and the first part of each longitudinal trajectory forecasts the remaining time points, and performance is reported as the \emph{prediction score (Pred.)} corresponding to the MSE on real test data, and compared against a Train-on-Real-Test-on-Real (TRTR) reference.
    \item \textbf{Privacy} evaluation assesses whether synthetic samples reveal information about the training data beyond what is expected from the underlying population distribution. We report \emph{nearest-neighbor adversarial accuracy (NNAA)}~\cite{yale2020generation}, \emph{membership inference risk (MIR)}~\cite{shokri2017membership}, and a \emph{dependence score (Dep.)}~\cite{stenger2025steb} measuring residual dependence between generated samples and training individuals, computed on joint multimodal representations.
\end{itemize}

\paragraph{Simulated datasets}
\label{sec:simulated_datasets_ref}
To evaluate whether models recover controlled multimodal dependencies, we construct simulated datasets in which patient-specific static covariates drive longitudinal dynamics. For each individual $i$, we generate static variables $\mathbf{W}_i = [\mathrm{W}_{i,\mathrm{num}}, \mathrm{W}_{i,\mathrm{bin}}] \in \mathbb{R}^{d_W}$, 
with one numerical feature sampled from a Gaussian distribution, and one binary feature generated through a probit model. Conditioned on $\mathbf{W}_i$, the longitudinal process $\mathbf{x}_i(t) \in \mathbb{R}^{d_\mathrm{x}}$ follows independent fractional Ornstein--Uhlenbeck dynamics,
\[
d \mathbf{x}_i^{(\ell)}(t) = -a_{i,\ell}\bigl(\mathbf{x}_i^{(\ell)}(t)-b_{i,\ell}\bigr)\,dt + \zeta\, dB^H_{i,\ell}(t),
\]
where $B^H_{i,\ell}(t)$ is a fractional Brownian motion, $\zeta > 0$ is a global diffusion scale, and the subject-specific parameters $a_{i,\ell}$ and $b_{i,\ell}$ are linear functions of the static covariates. Observation times are sampled from a non-homogeneous Poisson process to mimic irregular clinical follow-up. Each dataset contains $N=1000$ individuals with $d_W=2$ static variables, $d_{\mathrm{x}}=3$ longitudinal variables, and trajectories generated over a horizon $T=7$; the reference setting yields on average $n_{\mathrm{obs}}=50$ visits per patient. Full simulation details are provided in Appendix~\ref{app:simulation}.

\paragraph{Real-world datasets} 
We evaluate HyperNSDE on two real clinical datasets covering diverse medical domains and observation regimes (see Table~\ref{tab:dataset_summary} in Appendix~\ref{app:realworld_datasets} for summary statistics). 
{VELOUR}~\cite{velour2021, projectdatasphere} is a phase III randomized trial of aflibercept plus FOLFIRI in metastatic colorectal cancer accessed via Project DataSphere, with dense longitudinal laboratory and vital sign measurements (23 longitudinal variables, 26 static covariates).
{PPMI}~\cite{marek2011parkinson} is a longitudinal Parkinson's disease cohort with clinical scores and vital signs collected over multi-year follow-up (15 longitudinal variables, 20 static covariates). 
Full preprocessing details are provided in Appendix~\ref{app:realworld_datasets}.
% {PhysioNet 2012}~\cite{silva2012predicting} is a benchmark ICU dataset comprising 3936 patients with up to 48 hours of physiological measurements (9 longitudinal variables, 19 static covariates). 
% We evaluate HyperNSDE on four real clinical datasets covering diverse medical domains and observation regimes (see Table~\ref{tab:dataset_summary} in Appendix~\ref{app:realworld_datasets} for summary statistics). {PhysioNet 2012}~\cite{silva2012predicting} is a benchmark ICU dataset comprising 3936 patients with up to 48 hours of physiological measurements (9 longitudinal variables, 19 static covariates). {PPMI}~\cite{marek2011parkinson} is a longitudinal Parkinson's disease cohort with clinical scores and vital signs collected over multi-year follow-up (15 longitudinal variables, 20 static covariates). {MSK-CHORD}~\cite{jee2024automated} is a real-world oncology registry of prostate cancer patients with PSA measurements collected over up to 25 years (2 longitudinal variables, 21 static covariates). {VELOUR}~\cite{velour2021, projectdatasphere} is a phase III randomized trial of aflibercept plus FOLFIRI in metastatic colorectal cancer accessed via Project DataSphere, with dense longitudinal laboratory and vital sign measurements (23 longitudinal variables, 26 static covariates). Full preprocessing details are provided in Appendix~\ref{app:realworld_datasets}.

% \subsection{Baselines and experimental setup}

\paragraph{Baselines and reference}
We compare our method against three baselines: \emph{MultiNODEs} \cite{wendland2022generation},  \emph{RTSGAN} \cite{pei2021rtsgan}, and \emph{DGBFGP} \cite{balik2025bayesian}. These models were selected for their relevance to the joint static and irregular longitudinal setting and the availability of public, sufficiently well-documented implementations, unlike several other models listed in Table~\ref{tab:model_comparison}.
For \emph{DGBFGP}, which does not natively provide a generative model for static covariates, we augment it with an HI-VAE \cite{Nazabal2020} for static data generation (\emph{DGBFGP+HI-VAE}). As \emph{DGBFGP} requires substantial dataset-specific manual configuration, this comparison is restricted to simulated data.
In all tables, the \emph{Reference} row is not a generative model: it is obtained by randomly splitting the real dataset in half and treating one half as synthetic, providing an empirical reference point that a well-fitted generative model should approach.
Additional details are provided in Appendix~\ref{app:baselines}.

\paragraph{Experimental setup} 
\label{sec:experimental_setup}
% For simulated data, results are reported over $N_{\text{MC}} = 30$ independent Monte Carlo runs, each based on a newly generated dataset with a different random seed. In all experiments, the original data are split into training and test sets with an 80/20 ratio. Hyperparameters are selected in a preliminary tuning phase (Appendix~\ref{app:hyperopt}). To robustly assess generative performance, each trained model is used to generate $N_{\text{gen}} = 50$ independent synthetic test sets, each matching the size $N_{\text{test}}$ of the real test set. Reported results are averaged over these repeated generations. Our implementation is publicly available at \perrine{\url{https://github.com/...}}, where signature kernel computations are performed using the \texttt{pySigLib} package~\cite{shmelev2025pysiglib}. All experiments are run on NVIDIA V100 GPUs on the Jean Zay cluster. 
For simulated data, results are reported over $N_{\text{MC}}=30$ independent Monte Carlo runs, each based on a newly generated dataset with a different random seed. In all experiments, original data are split into training and test sets with an 80/20 ratio. Hyperparameters are selected via a preliminary tuning phase (Appendix~\ref{app:hyperopt}). To robustly assess generative performance, each trained model is used to generate $N_{\text{gen}}=50$ independent synthetic test sets, each matching the real test set size $N_{\text{test}}$; reported results are averaged over these repeated generations. All experiments run on NVIDIA V100 GPUs on the Jean Zay cluster (Appendix~\ref{app:exp_setting_resources}). Our implementation is publicly available at \url{https://github.com/perrinechassat/HyperNSDE}; signature-kernel computations are performed using \texttt{pySigLib}~\cite{shmelev2025pysiglib}.

% Training HyperNSDE uses early stopping with patience 10 and a maximum of 1000 epochs. On the reference simulated dataset, one epoch takes approximately 1\,min\,20\,s on a single NVIDIA V100 GPU, with training terminating around epoch 500 ($\approx$11 hours); training time scales with the number of patients and observations for real-world datasets.
% Training HyperNSDE takes approximately \TODO{X} hours per dataset on a single GPU.

\section{Results}

\begin{figure}[h]
    \centering
    \includegraphics[width=1\linewidth]{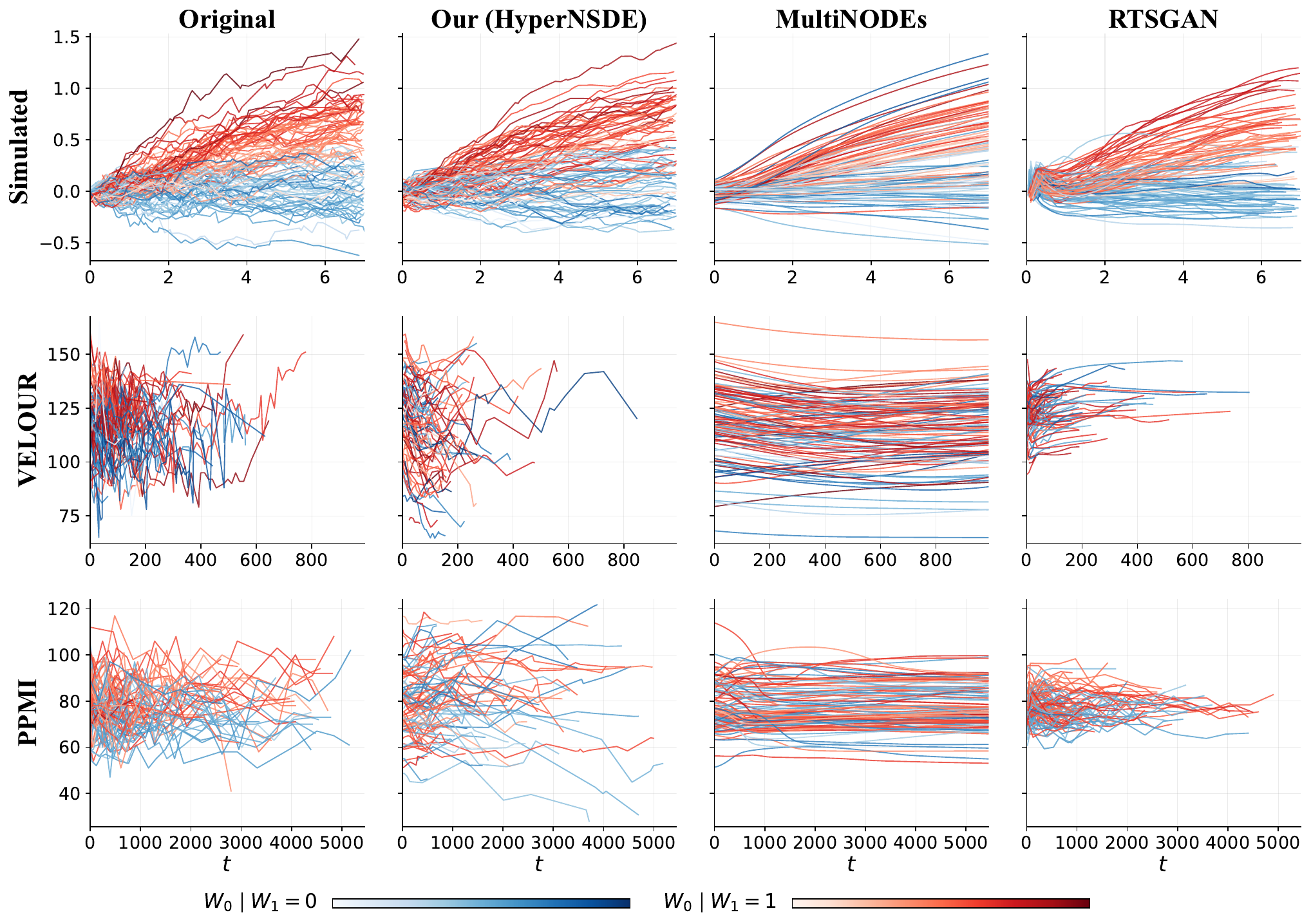}
    % \caption{\textbf{Qualitative comparison of generated trajectories across datasets and methods.} Color encodes a continuous static covariate $W_0$ (blue: $W_1=0$, red: $W_1=1$). Per dataset --- Simulated: $W_0 \in [-0.79, 0.73]$, plotted variable $x_0$; VELOUR: $W_0 = \text{Age} \in [22, 79]$, $W_1 = \text{Sex}$, plotted variable is Hemoglobin; PPMI: $W_0 = \text{Years of education} \in [0, 30]$, $W_1 = \text{Sex}$, plotted variable is Diastolic blood pressure supine.
    % Additional results on simulated datasets are in Appendix~\ref{app:additional_results}.}
    \caption{\textbf{Qualitative comparison of real and generated longitudinal trajectories.} Rows denote datasets and columns the real data and generative methods. Trajectory color represents the static covariates: blue trajectories have $W_1=0$ and red ones have $W_1=1$, with color intensity encoding the continuous covariate $W_0$. For the simulated data, $W_0\in[-0.79,0.73]$ and the plotted variable is $x_0$; for VELOUR, $W_0$ is age ($22$--$79$ years), $W_1$ is sex, and the plotted variable is hemoglobin; for PPMI, $W_0$ is years of education ($0$--$30$), $W_1$ is sex, and the plotted variable is supine diastolic blood pressure. Additional simulated-data results are in Appendix~\ref{app:additional_results}.}
    \label{fig:simu_ref_setting_x0}
\end{figure}

Table~\ref{tab:all_results} and Figure~\ref{fig:simu_ref_setting_x0} report the results in simulated and real-world datasets.
Because longitudinal Corr.\ RMSE depends on the time points jointly observed across variables, and TSTR prediction may be affected by both sampling-grid regularity and trajectory smoothness, we complement the native-grid comparison with controlled reevaluations in Table~\ref{tab:effect_grid_res}. MultiNODEs and HyperNSDE can be evaluated on common time grids, and a drift-only HyperNSDE variant further isolates the effect of trajectory smoothness.

\begin{table*}[h]
    \centering
    % \fontsize{6.5pt}{7.5pt}\selectfont
    % \tiny
    \scriptsize
    \caption{\textbf{Evaluation metrics on simulated and real-world data.} $\downarrow$ lower is 
    better; $\rightarrow 0$ closer to zero is better (NNAA). \textbf{Bold} is the best 
    result excluding the reference, \underline{underlined} is second best.  Standard deviations across $N_{\mathrm{MC}}=30$ Monte Carlo runs and $N_{\text{gen}}=50$ generated datasets from a single trained model are reported in Table~\ref{tab:all_results_std} in Appendix~\ref{app:stat_var_robustness}.}
    \label{tab:all_results}
    \definecolor{rowgray}{gray}{0.9}
    \setlength{\tabcolsep}{4pt}
    \renewcommand{\arraystretch}{1.3}
    \setlength{\heavyrulewidth}{1pt}
    \setlength{\lightrulewidth}{1pt}
    \makebox[\textwidth][c]{%
    %
    % ---- LEFT: dataset label column ----
    \begin{tabular}[t]{r}
    \phantommid
        \\
        \phantomcmid
        \noalign{\vspace{-2pt}}
        \\[-3pt]
        \phantomcmid
        \noalign{\vspace{-2.8pt}}
        \\[-2pt]
        \phantommid
        \multirow{5}{*}{\rotatebox{90}{\textbf{Simulated}}} \\[-2pt]
        \phantomcmid
        \\
        \\
        \\
        \\
        \arrayrulecolor{white}\midrule\arrayrulecolor{black}
        \multirow{4}{*}{\rotatebox{90}{\textbf{VELOUR}}} \\[-2pt]
        \phantomcmid
        \\
        \\
        \\
        \arrayrulecolor{white}\midrule\arrayrulecolor{black}
        \multirow{4}{*}{\rotatebox{90}{\textbf{PPMI}}} \\[-2pt]
        \phantomcmid
        \\
        \\
        \\
        \phantommid
    \end{tabular}%
    %
    % ---- RIGHT: main table ----
    \begin{tabular}[t]{lcccccccccc}
\toprule
  & \multicolumn{6}{c}{\textbf{Fidelity}} & \multicolumn{3}{c}{\textbf{Privacy}} & \textbf{Utility} \\
\cmidrule(lr){2-7} \cmidrule(lr){8-10} \cmidrule(lr){11-11}
\noalign{\vspace{-2pt}}
   & \multicolumn{2}{c}{\textbf{Corr. RMSE} $\downarrow$}
    & \multirow{2}{*}{\textbf{Discrim.} $\downarrow$}
    & \multirow{2}{*}{\textbf{KL $\lambda$} $\downarrow$}
    & \multicolumn{2}{c}{\textbf{MMD} $\downarrow$}
    & \multirow{2}{*}{\textbf{Dep.} $\downarrow$}
    & \multirow{2}{*}{\textbf{MIR} $\downarrow$}
    & \multirow{2}{*}{\textbf{NNAA} $\rightarrow 0$}
    & \multirow{2}{*}{\textbf{Pred.} $\downarrow$} \\[-3pt]
\cmidrule(lr){2-3} \cmidrule(lr){6-7}
\noalign{\vspace{-2.8pt}}
  & \textbf{Long.} & \textbf{Static} & & & \textbf{Long.} & \textbf{Static} & & & & \\[-2pt]
%%% Simulated
\midrule
\rowcolor{white}
 Reference        & $0.061$ & $0.060$ & $0.499$ & $0.139$ & $0.021$   & $0.007$ & $0.001$ & $0.033$ & $0.021$ & $3.759$ \\ [-2pt]
\cmidrule(lr){1-11}
\rowcolor{rowgray}
MultiNODEs & $\mathbf{0.084}$  & $\underline{0.045}$ & $0.998$ & $67.616$ & $816.507$ & $0.006$ & $0.117$ & $\mathbf{0.034}$ & $-0.008$ & $11.328$  \\ [-1pt]
\rowcolor{white}
 RTSGAN           & $\underline{0.103}$ & $0.073$ & $\underline{0.932}$ & $\underline{2.525}$ & $18.427$ & $0.013$ & $0.020$ & $0.037$ & $-0.009$ & $\underline{5.278}$ \\
\rowcolor{rowgray}
 DGBFGP + HIVAE   & $0.167$ & $0.051$ & $0.999$ & $67.616$ & $299.330$ & $\underline{0.006}$ & $0.058$ & $\underline{0.034}$ & $\mathbf{-0.005}$ & $11.603$ \\
 \rowcolor{white}
HyperNSDE & $0.118$  & $\mathbf{0.044}$ & $\mathbf{0.926}$ & $\mathbf{0.687}$ & $\underline{1.150}$ & $\mathbf{0.004}$ & $\mathbf{0.014}$ & $0.039$ & $-0.021$ & $\mathbf{5.226}$\\

%%% VELOUR
\midrule
\rowcolor{white}
 Reference  & $0.091$ & $0.132$ & $0.472$ & $0.844$ & $0.001$ & $0.019$ & $0.063$ & $0.073$ & $-0.049$ & $21.958$ \\[-2pt]
\cmidrule(lr){1-11}
\rowcolor{rowgray}
 MultiNODEs  & $\mathbf{0.139}$  & $\underline{0.154}$ & $\mathbf{0.990}$ & $139.716$ & $\underline{0.144}$ & $\underline{0.019}$ & $0.628$ & $\mathbf{0.048}$ & $\underline{0.011}$ & $\mathbf{5.837}$ \\ 
\rowcolor{white}
 RTSGAN   & $0.842$ & $\mathbf{0.149}$ & $1.000$ & $\mathbf{2.581}$ & $121.596$ & $0.019$ & $\underline{0.118}$ & $\underline{0.050}$ & $0.084$ & $6048.192$ \\
\rowcolor{rowgray}
 HyperNSDE & $\underline{0.239}$ & $0.164$ & $\underline{0.998}$ & $\underline{3.513}$ & $\mathbf{0.029}$ & $\mathbf{0.019}$ & $\mathbf{0.105}$ & $0.063$ & $\mathbf{0.010}$ & $\underline{53.514}$ \\
 
%%% PPMI
\midrule
\rowcolor{white}
 Reference          & $0.044$ & $0.092$ & $0.587$ & $0.064$ & $0.001$ & $0.006$ & $0.052$ & $0.065$ & $-0.032$ & $39.137$ \\[-2pt]
\cmidrule(lr){1-11}
\rowcolor{rowgray}
 MultiNODEs        & $\mathbf{0.108}$ & $\underline{0.133}$ & $1.000$ & $221.398$ & $0.038$ & $0.013$ & $0.391$ & $\mathbf{0.017}$ & $\underline{0.006}$ & $\mathbf{19.948}$ \\ 
\rowcolor{white}
 RTSGAN            & $0.623$ & $\mathbf{0.114}$ & $\mathbf{0.983}$ & $\mathbf{1.042}$ & $\mathbf{0.009}$ & $\mathbf{0.012}$ & $\mathbf{0.057}$ & $0.057$ & $-0.007$ & $\underline{52.954}$ \\
\rowcolor{rowgray}
 HyperNSDE   & $\underline{0.260}$ & $0.135$ & $\underline{0.996}$ & $\underline{1.840}$ & $\underline{0.019}$  & $\underline{0.013}$ & $\underline{0.084}$ & $\underline{0.032}$ & $\mathbf{0.002}$ & $57.697$ \\ 

\bottomrule
    \end{tabular}}
    \end{table*}

\begin{table*}[h]
    \centering
    \scriptsize
    \caption{\textbf{Effect of observation-grid regularity and trajectory smoothness on longitudinal Corr. RMSE and utility metrics.} We reevaluate MultiNODEs and HyperNSDE on HyperNSDE-generated and real observation grids and use drift-only HyperNSDE to assess the effect of trajectory smoothness. $\downarrow$ lower is better. \textbf{Bold} is best and \underline{underlined} second best, excluding the reference.}
    \label{tab:effect_grid_res}
    \definecolor{rowgray}{gray}{0.9}
    \setlength{\tabcolsep}{4pt}
    \renewcommand{\arraystretch}{1.3}
    \setlength{\heavyrulewidth}{1pt}
    \setlength{\lightrulewidth}{1pt}
    \makebox[\textwidth][c]{%
    \begin{tabular}[t]{llcccccc}
\toprule
  & & \multicolumn{2}{c}{\textbf{Simulated}} & \multicolumn{2}{c}{\textbf{VELOUR}} &  \multicolumn{2}{c}{\textbf{PPMI}}  \\
\cmidrule(lr){3-4}  \cmidrule(lr){5-6} \cmidrule(lr){7-8}
\noalign{\vspace{-2pt}}
\multirow{2}{*}{\textbf{Model}} &
\multirow{2}{*}{\textbf{Setting (obs. grid)}} 
   & \multicolumn{1}{c}{\textbf{Corr. RMSE} $\downarrow$}
    & \multirow{2}{*}{\textbf{Pred.} $\downarrow$}
       & \multicolumn{1}{c}{\textbf{Corr. RMSE} $\downarrow$}
    & \multirow{2}{*}{\textbf{Pred.} $\downarrow$}
       & \multicolumn{1}{c}{\textbf{Corr. RMSE} $\downarrow$}
    & \multirow{2}{*}{\textbf{Pred.} $\downarrow$}
    \\[-3pt]
\cmidrule(lr){3-3} \cmidrule(lr){5-5} \cmidrule(lr){7-7}
\noalign{\vspace{-2.8pt}}
 & & \textbf{Long.} & & \textbf{Long.} & & \textbf{Long.} & \\[-2pt]
%%% Simulated
\midrule
% \rowcolor{white}
Reference
& real obs.
& $0.061$ & $3.759$
& $0.091$ & $21.958$
& $0.044$ & $39.137$ \\ 
% [-2pt]
\cmidrule(lr){1-8}

\multirow{3}{*}{MultiNODEs}
& \cellcolor{rowgray} native (regular)
& \cellcolor{rowgray} $\mathbf{0.084}$
& \cellcolor{rowgray} $11.328$
& \cellcolor{rowgray} $\mathbf{0.139}$
& \cellcolor{rowgray} $\mathbf{5.837}$
& \cellcolor{rowgray} $\mathbf{0.108}$
& \cellcolor{rowgray} $\mathbf{19.948}$ \\

% \rowcolor{white}
% \cellcolor{white}
& HyperNSDE grid
& $0.502$ & $5.815$
& $0.940$ & $39.275$
& $0.915$ & $57.621$ \\

&
\cellcolor{rowgray} real obs.
& \cellcolor{rowgray} $0.494$
& \cellcolor{rowgray} $6.067$
& \cellcolor{rowgray} $0.940$
& \cellcolor{rowgray} $7.357$
& \cellcolor{rowgray} $0.914$
& \cellcolor{rowgray} $\underline{21.841}$ \\

\cmidrule(lr){1-8}

\multirow{3}{*}{HyperNSDE}
& native (generated)
& $0.118$ & $5.226$
& $0.239$ & $53.514$
& $0.260$ & $57.697$ \\

&
\cellcolor{rowgray} real obs.
& \cellcolor{rowgray} $\underline{0.115}$
& \cellcolor{rowgray} $\underline{4.379}$
& \cellcolor{rowgray} $0.284$
& \cellcolor{rowgray} $10.633$
& \cellcolor{rowgray} $0.261$
& \cellcolor{rowgray} $36.173$ \\

&
real obs., drift only
& $0.171$ & $\mathbf{3.636}$
& $\underline{0.216}$ & $\underline{7.010}$
& $\underline{0.241}$ & $23.298$ \\
\bottomrule
    \end{tabular}}
    \end{table*}

\paragraph{Simulated data} 
Table~\ref{tab:all_results} and Figure~\ref{fig:simu_ref_setting_x0} show that HyperNSDE performs competitively across the complementary evaluation axes in the reference simulated setting, averaged over $N_{\text{MC}}=30$ Monte Carlo runs and $N_{\text{gen}}=50$ generated datasets per trained model. It achieves the best discriminative, dependence, KL~$\lambda$, and predictive mean scores, along with the lowest longitudinal MMD ($16\times$ lower than RTSGAN). Qualitatively, it is also the only method that simultaneously reproduces (i)~the color stratification induced by $W_0$ and $W_1$, (ii)~stochastic variability across trajectories, and (iii)~irregular observation patterns. By contrast, MultiNODEs captures the $W_1$-driven separation but produces overly smooth deterministic trajectories; RTSGAN shows partial color stratification but collapses near $t=0$; and DGBFGP~+~HI-VAE generates flat trajectories with low variance (Figure~\ref{app:fig_simu_ref_full}). MultiNODEs achieves a lower native-grid Corr.\ RMSE, but matched-grid analysis in Table~\ref{tab:effect_grid_res} shows that this advantage is strongly affected by grid regularity: on the real observation grid, its Corr.\ RMSE increases from $0.084$ to $0.494$, whereas HyperNSDE changes from $0.118$ to $0.115$. Prediction scores are likewise grid-dependent, while the drift-only results show that the MSE objective also favors smoother trajectories.

The discriminative-score decomposition in Appendix~\ref{app:metric_diagnostics} further shows that static covariates are difficult to distinguish from real data across all methods, whereas observation patterns strongly distinguish the baselines; for HyperNSDE, the remaining gap primarily arises from trajectory values. Subgroup analyses restricted to $W_1=0$ and $W_1=1$ (Tables~\ref{tab:simu_ref_setting_W1_0}--\ref{tab:simu_ref_setting_W1_1}) further assess preservation of subpopulation-specific structure. Additional sensitivity analyses in Appendix~\ref{app:sensitivity_analysis_results} examine limited training data, sparse observations, state-dependent observation sampling, prior versus posterior latent sampling, and the state dependence of the modeled observation intensity. Paired Wilcoxon tests across the $30$ Monte
Carlo runs confirm significant improvements in longitudinal and static MMD against all baselines, whereas longitudinal Corr.\ RMSE differences are not significant. For KL~$\lambda$, the paired results are baseline-dependent, with significant improvements over MultiNODEs and DGBFGP+HI-VAE (Appendix~\ref{app:stat_var_robustness}).

\paragraph{Real-world datasets}
Results on VELOUR and PPMI are more mixed than on simulated data, reflecting the added complexity of real clinical trajectories, including heterogeneous missingness, patient-specific follow-up, and informative observation times. HyperNSDE achieves the best NNAA, the best longitudinal MMD on VELOUR (about $5\times$ and $4000\times$ lower than MultiNODEs and RTSGAN), and a competitive longitudinal MMD on PPMI. For observation-intensity fidelity, HyperNSDE yields substantially lower KL~$\lambda$ than MultiNODEs on both datasets, although RTSGAN obtains the lowest values.
MultiNODEs performs better on native-grid Corr.\ RMSE and prediction, but Table~\ref{tab:effect_grid_res} shows that these comparisons are strongly affected by the evaluation setting. On VELOUR, moving to the real observation
grid changes Corr.\ RMSE from $0.139$ to $0.940$ for MultiNODEs and from $0.239$ to $0.284$ for HyperNSDE, while prediction changes from $5.837$ to $7.357$ and from $53.514$ to $10.633$, respectively. HyperNSDE's drift-only prediction further reaches $7.010$ on the same grid, close to MultiNODEs, indicating once again that the MSE forecasting objective favors smoother deterministic trajectories. Similar effects are observed on PPMI. This is consistent with Figure~\ref{fig:simu_ref_setting_x0}, where MultiNODEs produces smoother paths but does not reproduce the stochastic variability of the real trajectories. Finally, unlike the fixed follow-up used in simulation, real datasets exhibit highly variable patient-specific end-of-sequence times, further complicating evaluation on short heterogeneous trajectories.

\paragraph{Ablation studies}
Table~\ref{tab:ablation_studies} and Figure~\ref{fig:simu_ablation_studies} in Appendix~\ref{app:qualitative_results} isolate the contribution of each component. Replacing the Neural SDE with an ODE (AS.1) preserves utility but degrades correlation RMSE and MMD, confirming that stochastic dynamics are necessary to capture trajectory variability beyond a single deterministic path. 
Removing the det--sto decomposition (AS.2) introduces excessive stochastic noise that overwhelms the trend signal, yielding chaotic trajectories with poor color stratification and a severe drop in correlation RMSE and utility, validating the importance of the proposed strategy for stabilizing training.
Among conditioning alternatives (AS.3), initial-state conditioning (MNSDE), continuous concatenation (SNSDE), and FiLM modulation (FNSDE) are competitive on individual metrics, but none dominates HyperNSDE; the hypernetwork achieves the lowest longitudinal MMD and generates visually less noisy, better-separated trajectories while remaining competitive elsewhere. These results support hypernetwork-based personalization as an effective choice without establishing it as uniquely necessary.
Finally, removing the observation model (AS.4) causes a large increase in KL $\lambda$ (from $0.687$ to $67.616$) and MMD, underscoring the importance of explicitly modeling informative sampling patterns.
Additional details on these ablations are provided in Appendix~\ref{app:ablations}.

\begin{table*}[h]
    \centering
    \scriptsize 
    \caption{\textbf{Ablation studies on simulated data.} 
    % $\downarrow$ lower is better; $\rightarrow 0$ closer to zero is better (NNAA). \textbf{Bold} indicates the best result (excluding the reference), \underline{underlined} the second best. 
    Standard deviations across $N_{\mathrm{MC}}=30$ Monte Carlo runs and $N_{\text{gen}}=50$ generated datasets are reported in Table~\ref{tab:ablation_studies_std} in Appendix~\ref{app:stat_var_robustness}.}
    \label{tab:ablation_studies}
    \definecolor{rowgray}{gray}{0.9}
    \setlength{\tabcolsep}{4pt}
    \renewcommand{\arraystretch}{1.3}
    \setlength{\heavyrulewidth}{1pt}
    \setlength{\lightrulewidth}{1pt}
    \makebox[\textwidth][c]{
    \begin{tabular}{lcccccccccc}
\toprule
  & \multicolumn{6}{c}{\textbf{Fidelity}} & \multicolumn{3}{c}{\textbf{Privacy}} & \textbf{Utility}  \\
\cmidrule(lr){2-7} \cmidrule(lr){8-10} \cmidrule(lr){11-11} 
\noalign{\vspace{-2pt}}
   & \multicolumn{2}{c}{\textbf{Corr. RMSE} $\downarrow$} 
    & \multirow{2}{*}{\textbf{Discrim.} $\downarrow$} 
    & \multirow{2}{*}{\textbf{KL $\lambda$} $\downarrow$} 
    & \multicolumn{2}{c}{\textbf{MMD} $\downarrow$} 
    & \multirow{2}{*}{\textbf{Dep.} $\downarrow$} 
    & \multirow{2}{*}{\textbf{MIR} $\downarrow$}  
    & \multirow{2}{*}{\textbf{NNAA} $\rightarrow 0$} 
    & \multirow{2}{*}{\textbf{Pred.} $\downarrow$} \\[-3pt]
\cmidrule(lr){2-3} \cmidrule(lr){6-7}
\noalign{\vspace{-2.8pt}}
  & \textbf{Long.} & \textbf{Static} & & & \textbf{Long.} & \textbf{Static} & & & & \\[-2pt]
\midrule
\rowcolor{white}
  Reference & $0.061$ & $0.060$ & $0.499$ & $0.139$ & $0.021$ & $0.007$ & $0.001$ & $0.033$ & $0.021$ & $3.734$ \\[-2pt]
\cmidrule(lr){1-11}
\rowcolor{rowgray}
  AS.1 only ODE & $0.160$ & $0.040$ & $0.992$ & ${0.773}$ & $92.621$ & $0.004$ & $0.095$ & $0.046$ & $-0.025$ & $\mathbf{4.515}$ \\
% \cmidrule(lr){1-11}
\rowcolor{white}
  AS.2 training & $0.233$ & $0.052$ & $0.964$ & ${0.971}$ & $1.209$ & $0.006$ & $0.042$ & $\mathbf{0.035}$ & $\underline{-0.007}$ & $11.880$ \\
% \cmidrule(lr){1-11}
\rowcolor{rowgray}
  AS.3 MNSDE & $0.154$ & $\mathbf{0.024}$ & $0.954$ & $\mathbf{0.422}$ & $1.555$ & $\mathbf{0.003}$ & $0.023$ & $0.045$ & $-0.027$ & $\underline{4.840}$ \\
\rowcolor{white}
  AS.3 SNSDE & $\underline{0.110}$ & $\underline{0.026}$ & $\mathbf{0.846}$ & $\underline{0.582}$ & $\underline{1.201}$ & $\underline{0.003}$ & $\underline{0.013}$ & $0.047$ & $-0.035$ & $5.485$ \\
\rowcolor{rowgray}
  AS.3 FNSDE & ${0.149}$ & $0.073$ & ${0.926}$ & ${0.584}$ & $1.673$ & $0.010$ & $\mathbf{0.010}$ & $\underline{0.035}$ & $\mathbf{-0.001}$ & $6.295$ \\
% \cmidrule(lr){1-11}
\rowcolor{white}
  AS.4 w/o lbda & $\mathbf{0.101}$ & $0.045$ & $0.986$ & $67.616$ & $149.558$ & $0.004$ & $0.018$ & ${0.037}$ & ${-0.016}$ & $11.563$ \\
% \cmidrule(lr){1-11}
\rowcolor{rowgray}
  HyperNSDE & $0.118$ & $0.044$ & $\underline{0.926}$ & $0.687$ & $\mathbf{1.150}$ & $0.004$ & ${0.014}$ & $0.039$ & $-0.021$ & $5.226$ \\
\bottomrule
    \end{tabular}}
    \end{table*}

\section{Conclusion}
\label{sec:conclusion}

We introduced HyperNSDE, a continuous-time generative model for heterogeneous static covariates, irregular longitudinal trajectories, and informative observation times. By combining a generative model for static covariates with a hypernetwork-conditioned latent Neural SDE, HyperNSDE captures both inter-individual variability driven by baseline characteristics and intra-individual stochastic fluctuations. 
Experiments show improved observation-time fidelity and competitive performance across complementary metrics, while matched-grid analyses reveal important sensitivity of the metrics to sampling regularity and trajectory smoothness.

Several limitations remain. HyperNSDE learns the observational distribution of its training cohort rather than causal counterfactuals, and therefore cannot by itself correct selection bias or extrapolate to unsupported subgroups. It is also computationally more demanding than the baselines. More broadly, our experiments show that existing metrics can disagree with trajectory realism for sparse, irregular, stochastic data, while discriminative scores remain high for all methods, highlighting both the difficulty of the problem and the limits of current approaches. In their present form, such models may already support benchmarking use cases, but further progress is needed before they can be confidently used for cohort augmentation or clinical trial simulation.

These limitations suggest several directions for future work. On the modeling side, richer diffusion parameterizations \cite{oh2024stable}, including covariates-dependent diffusion terms, could improve the representation of subject-specific stochasticity. Extending the framework to typed longitudinal variables (e.g., integer, positive, or bounded) through adapted output distributions would also improve realism for clinical biomarkers. 
Integrating differential privacy during training would provide formal guarantees to complement our empirical privacy evaluation. On the methodological side, a deeper theoretical and empirical study of the deterministic--stochastic decomposition would help clarify when and why it improves training in sparse irregular settings. Finally, improving evaluation protocols and metrics for synthetic irregular longitudinal data remains, in our view, the most important open problem, since current metrics are still the main limiting factor for reliably assessing generated-data quality and for making fair comparisons between models.

% \clearpage

\begin{ack}
For this work, PC and AG have benefited from support from the National Agency for Research under the France 2030 program, reference ANR-22-PESN-0016. PC and AG received funding from the MEDITWIN Bpifrance i-Demo project as part of the France 2030 program. AG has also benefited from the Institut National du Cancer within INCa\_CHAIRE IA-RC26-009.
\end{ack}

% \section*{References}

\clearpage
\bibliographystyle{plain}  % or unsrt, alpha, abbrvnat, neurips_2023, etc.
\bibliography{biblio}

\medskip

%%%%%%%%%%%%%%%%%%%%%%%%%%%%%%%%%%%%%%%%%%%%%%%%%%%%%%%%%%%%
% \clearpage
\appendix

\section{Model details}\label{app:model_detail}

\subsection{Additional details on the HI-VAE} \label{app:details_hivae}

The HI-VAE \cite{Nazabal2020} is designed to model static covariates with heterogeneous data types and missing values within a unified variational framework. In our setting, for each individual \(i\), the observed static vector \(\mathbf{W}_i\) is associated with a discrete latent variable \(\mathbf{s}_i\) and a continuous latent variable \(\mathbf{u}_i\), as introduced in Section~\ref{sec:static_HIVAE}. The decoder defines the conditional distribution
\[
p(\mathbf{W}_i \mid \mathbf{s}_i, \mathbf{u}_i)
=
\prod_{k=1}^{d_\mathrm{W}}
p\!\left(\mathbf{W}_i^{(k)} \mid \gamma_k(\mathbf{y}_i,\mathbf{s}_i)\right),
\qquad \mathbf{y}_i = f(\mathbf{u}_i),
\]
where each feature-specific decoder \(\gamma_k\) outputs the parameters of a likelihood adapted to the type of the \(k\)-th variable.

\paragraph{Feature-wise likelihoods}
Following \cite{Nazabal2020}, we use different observation models depending on the type of each static covariate:
\begin{itemize}
    \item Continuous real-valued variables (Normal): $\gamma_k(\mathbf{y}_i, \mathbf{s}_i) = \big(\mu_k(\mathbf{y}_i, \mathbf{s}_i), \sigma_k^2(\mathbf{y}_i, \mathbf{s}_i)\big)$, 
    \item Positive real-valued variables (Log-Normal): $\gamma_k(\mathbf{y}_i, \mathbf{s}_i) = \big(\mu_k(\mathbf{y}_i, \mathbf{s}_i), \sigma_k^2(\mathbf{y}_i, \mathbf{s}_i)\big)$, 
    \item Count variables (Poisson):  $\gamma_k(\mathbf{y}_i, \mathbf{s}_i) = \lambda_k (\mathbf{y}_i, \mathbf{s}_i)$,
    \item Categorical variables (Multinomial-logit): $\gamma_k(\mathbf{y}_i, \mathbf{s}_i) = \big(\gamma_{k0}(\mathbf{y}_i, \mathbf{s}_i), \hdots, \gamma_{k(R-1)}(\mathbf{y}_i, \mathbf{s}_i)\big)$.
\end{itemize}

\paragraph{Variational inference}
Inference is performed using variational encoders for both latent variables. The posterior factorization follows the HI-VAE construction:
\[
q(\mathbf{s}_i,\mathbf{u}_i \mid \mathbf{W}_i)
=
q(\mathbf{s}_i \mid \mathbf{W}_i)\,
q(\mathbf{u}_i \mid \mathbf{W}_i,\mathbf{s}_i).
\]
The discrete latent variable \(\mathbf{s}_i\) is handled using the Gumbel--Softmax reparameterization \cite{Li2019LearningLS}, while the continuous latent variable \(\mathbf{u}_i \mid \mathbf{s}_i\) is sampled using the standard Gaussian reparameterization trick \cite{Kingma2013AutoEncodingVB}. This enables end-to-end gradient-based optimization of the evidence lower bound.

\subsection{Additional details on the event intensity model} \label{app:details_eventmodel}

The observation process is modeled as a latent-state-dependent multivariate point process with integrated intensity
\[
\boldsymbol{\Lambda}_i(t)
= \int_0^t \boldsymbol{\lambda}_i(s)\,ds = \int_0^t 
\exp \bigl(\xi(\mathbf{z}_i(s),s;\theta_\xi)\bigr)ds,
\]
where the exponential is applied componentwise. In the original Latent ODE framework \cite{rubanova2019latent}, the integrated intensity is introduced as an additional state variable within the ODE system and evolves jointly with the latent trajectory. In our setting, however, we found that augmenting the SDE system with \(\boldsymbol{\Lambda}_i(t)\) increases both memory usage and computational cost, especially when backpropagating through long and irregular trajectories under stochastic dynamics. We therefore adopt a decoupled strategy. We first solve for the latent path \(\mathbf{z}_i(t)\), then evaluate the intensity \(\boldsymbol{\lambda}_i(t)\) at the required time points, and finally approximate the integrated intensity \(\boldsymbol{\Lambda}_i(t)\) by numerical quadrature. This approach preserves the flexibility of the latent-state-dependent point-process model while remaining scalable to high-dimensional latent spaces and large patient cohorts.

\subsection{Extended details on synthetic data generation}
\label{app:synthetic_generation}

As outlined in Figure~\ref{fig:model_HyperNSDE}, generating a fully synthetic individual follows a hierarchical sampling path. This ensures that the generated longitudinal trajectories are intrinsically conditioned on the generated static baseline characteristics. The exact procedure for generating a synthetic subject $j$ is as follows:

\begin{enumerate}[leftmargin=0.5cm]
    \item \textbf{Static covariate generation:} We sample the latent representations ${\mathbf{h}}_j=({\mathbf{s}}_j,{\mathbf{u}}_j)$ from the HI-VAE using either \textit{posterior} or \textit{prior} sampling. Under posterior sampling, latent representations are drawn from individual approximate posteriors $q(\mathbf{s}_i,\mathbf{u}_i \mid \mathbf{W}_i)$, then pooled, shuffled, and subsampled to form a synthetic cohort without preserving patient correspondence. Under prior sampling, they are drawn directly from the learned HI-VAE prior. We then decode these latents to obtain the synthetic static covariates $\tilde{\mathbf{W}}_j$. The two sampling strategies are compared in the sensitivity analysis Table~\ref{tab:prior_posterior_sensitivity}.
    \item \textbf{Conditioning the dynamics:} We map the static latent representation ${\mathbf{h}}_j$ through the hypernetwork $\mathcal{H}$ to generate the individualized drift parameters $\theta_{b,j} = \mathcal{H}(\tilde{\mathbf{h}}_j;\theta_{\mathcal H})$. This is the crucial step that preserves the static--longitudinal dependency.
    % \item \textbf{Latent trajectory simulation:} Using the individualized parameters $\theta_{b,j}$, we simulate a continuous latent path ${\mathbf{z}}_j(t)$ by solving the SDE defined in Equation \eqref{eq:latent_sde}.
    \item \textbf{Latent trajectory simulation:} In practice, rather than simulating the full latent SDE of Eq.~\eqref{eq:latent_sde} directly, we use the deterministic--stochastic decomposition introduced in Section~\ref{sec:training} and simulate the two components separately. The deterministic part $\mathbf{z}_j^{\mathrm{det}}(t)$ is obtained by solving the personalized Neural ODE with individualized drift parameters $\theta_{b,j}$, capturing the patient-specific smooth trend. The stochastic residual $\mathbf{z}_j^{\mathrm{sto}}(t)$ is simulated from the population-shared Neural SDE, capturing intra-individual stochastic fluctuations. The full latent path is $\mathbf{z}_j(t) = \mathbf{z}_j^{\mathrm{det}}(t) + \mathbf{z}_j^{\mathrm{sto}}(t)$, with both components initialized at $\mathbf{0}_{d_\mathrm{z}}$.
    \item \textbf{Irregular sampling simulation:} To replicate the irregular observation patterns of the real data, we compute the event intensities ${\boldsymbol{\lambda}}_j(t)$ based on the latent path via Equation \eqref{eq:event_intensity}, and sample a set of discrete observation times $\tilde{\mathcal{T}}_j$.
    \item \textbf{Longitudinal decoding:} Finally, we evaluate each latent component at the sampled observation times and map these states through the decoder network to obtain the observable longitudinal values: $\tilde{\mathbf{x}}_j(t)=g({\mathbf{z}_j^{\mathrm{det}}(t)};\theta_g) + g({\mathbf{z}_j^{\mathrm{sto}}(t)};\theta_g)$.
\end{enumerate}

This yields synthetic multimodal records preserving static–longitudinal dependence, temporal variability, and irregular observation patterns.

% \clearpage
\FloatBarrier

\section{Training details/hyperparameters} 
\label{app:training_hyperopt}

\subsection{Deterministic--stochastic decomposition: further discussion}
\label{app:discussion_splitSDEtrain}

\paragraph{Full specification}
The deterministic component $\mathbf{z}_i^{\mathrm{det}}$ solves the personalized Neural ODE
\[
d\mathbf{z}_i^{\mathrm{det}}(t)
=
b^{\mathrm{det}}_i\bigl(\mathbf{z}_i^{\mathrm{det}}(t),t;\,\theta_{b^{\mathrm{det}},i}
\bigr)\,dt,
\qquad
\mathbf{z}_i^{\mathrm{det}}(0) = \mathbf{0},
\]
where $\theta_{b^{\mathrm{det}},i} = \mathcal{H}(\mathbf{h}_i;\theta_{\mathcal{H}})$ are the hypernetwork-generated parameters, trained via the reconstruction loss $\mathcal{L}_{\mathrm{long,det}}$ (Eq.~\ref{eq:loss_det}). The stochastic residual $\mathbf{z}_i^{\mathrm{sto}}$ solves the population-shared Neural SDE
\[
d\mathbf{z}_i^{\mathrm{sto}}(t)
=
b^{\mathrm{sto}}\bigl(\mathbf{z}_i^{\mathrm{sto}}(t),t;\,\theta_{b^{\mathrm{sto}}}
\bigr)\,dt
+
\sigma^{\mathrm{sto}}\bigl(\mathbf{z}_i^{\mathrm{sto}}(t),t;\,
\theta_{\sigma^{\mathrm{sto}}}\bigr)\,d\mathbf{B}_i(t),
\qquad
\mathbf{z}_i^{\mathrm{sto}}(0) = \mathbf{0},
\]
where $b^{\mathrm{sto}}$ is a low-capacity drift network and $\sigma^{\mathrm{sto}}$ models residual stochastic fluctuations, trained via the signature kernel score $\mathcal{L}_{\mathrm{long,sto}}$ (Eq.~\ref{eq:loss_sto}).

Then, the decomposition $\mathbf{z}_i = \mathbf{z}_i^{\mathrm{det}} + \mathbf{z}_i^{\mathrm{sto}}$ defines a different model from a single latent SDE on $\mathbf{z}_i$. The total process satisfies
\[
d\mathbf{z}_i(t) = \bigl[b^{\mathrm{det}}_i(\mathbf{z}_i^{\mathrm{det}},t) + 
b^{\mathrm{sto}}(\mathbf{z}_i^{\mathrm{sto}},t)\bigr]\,dt 
+ \sigma^{\mathrm{sto}}(\mathbf{z}_i^{\mathrm{sto}},t)\,d\mathbf{B}_i(t),
\]
which matches the form of a single latent SDE only when the drift decomposes additively across the two components --- a condition satisfied exactly in the linear case and approximately otherwise. The decomposition is therefore a deliberate modeling choice. The low-capacity drift $b^{\mathrm{sto}}$ absorbs any residual trend not captured by $\mathbf{z}_i^{\mathrm{det}}$, ensuring the decomposition remains well-behaved even when the linear approximation breaks down.

\paragraph{Connection to related work}
Separately estimating drift and diffusion from discrete observations is a classical statistical problem~\cite{comte2021nonparametric, mohammadi2024nonparametric, zhao2026drift}, but existing methods typically require the number of observed time points per path to tend to infinity --- an assumption incompatible with our sparse irregular setting. To the best of our knowledge, no prior work proposes a decomposition of the form~\eqref{eq:latent_decomposition} in the context of Neural SDEs. Our decomposition is instead a modeling choice inspired by the 
Doob--Meyer decomposition of a semimartingale into a predictable finite-variation process and a local martingale~\cite{protter2004stochastic}, and by the FDA practice of analyzing residual trajectories separately from a smooth trend~\cite{ramsay1997functional, yao2005functional}.

% \paragraph{Hypothesis on the role of irregular sampling} \perrine{à potentiellement enlever ?}
% We conjecture that the decomposition is particularly beneficial in the irregularly sampled setting, for the following reason. The signature kernel score used to train $\mathbf{z}_i^{\mathrm{sto}}$ is sensitive to the geometric structure of paths. When trajectories are sparsely observed, the signature of a subsampled path may differ substantially from that of the underlying continuous process. Training a single SDE to simultaneously recover trend and stochastic fluctuations from such sparse paths may therefore be ill-conditioned. By first removing the deterministic trend via $\mathbf{z}_i^{\mathrm{det}}$, the remaining residual signal could geometrically closer to the true stochastic component of the latent process, yielding a better-conditioned optimization target for the signature kernel objective. Formalizing this conjecture theoretically --- in particular through the approximation theory of path signatures under subsampling --- is an interesting direction for future work.

\subsection{Additional details on signature-kernel loss for training Neural SDEs}
\label{app:details_splitSDEtrain}

\subsubsection{Signature-kernel score}

To train stochastic path-valued generative models without adversarial optimization, we use the \emph{signature kernel score} introduced in \cite{salvi2021signature,Issa2023}. This score compares probability distributions on paths through a positive-definite kernel defined from path signatures.

\paragraph{Path signature}
Let \(X : [0,T] \to \mathbb{R}^d\) be a continuous path of bounded variation. Its signature is the sequence of iterated integrals
\[
S(X)
=
\Bigl(
1,\;
\int_0^T dX_{t_1},\;
\int_{0<t_1<t_2<T} dX_{t_1}\otimes dX_{t_2},\;
\dots
\Bigr).
\]
The signature provides a nonparametric representation of the path and characterizes it up to mild equivalence relations \cite{salvi2021signature}. Importantly, signature-based representations do not require regularly sampled trajectories: they are naturally defined for continuous paths and can be computed from discretely observed paths on irregular time grids.

\paragraph{Signature kernel}
For two paths \(X\) and \(Y\), the signature kernel is defined as the inner product of their signatures in the tensor algebra,
\[
k_{\mathrm{sig}}(X,Y)
=
\bigl\langle S(X), S(Y) \bigr\rangle.
\]
In practice, normalized or RBF-transformed versions are often used for numerical stability; in our experiments, we use an RBF-signature kernel as in \cite{Issa2023}. This defines a positive-definite kernel on path space and therefore an associated reproducing kernel Hilbert space.

\paragraph{Signature kernel score}
Let \(\mathbb{P}_{\theta}\) denote the law of trajectories generated by the model and let \(X\sim\mathbb{P}_{\mathrm{data}}\) denote a data path. The signature kernel score of \(\mathbb{P}_{\theta}\) at \(X\) is
\begin{equation*}
\phi_{\mathrm{sig}}(\mathbb{P}_{\theta},X)
=
\mathbb{E}_{\hat X,\hat X' \sim \mathbb{P}_{\theta}}
\bigl[k_{\mathrm{sig}}(\hat X,\hat X')\bigr]
-
2\,
\mathbb{E}_{\hat X \sim \mathbb{P}_{\theta}}
\bigl[k_{\mathrm{sig}}(\hat X,X)\bigr],
\label{eq:sig_score_population}
\end{equation*}
where \(\hat X,\hat X'\) are independent samples from the model. The expected training objective is then
\begin{equation*}
\mathcal{L}_{\mathrm{sig}}
=
\mathbb{E}_{X\sim\mathbb{P}_{\mathrm{data}}}
\bigl[
\phi_{\mathrm{sig}}(\mathbb{P}_{\theta},X)
\bigr].
\label{eq:sig_score_expected}
\end{equation*}
This score differs from the full kernel MMD only by a term depending solely on the data distribution, and is therefore equivalent for optimization with respect to the model parameters.

\paragraph{Empirical estimator}
Given two mini-batches of trajectories
\[
\mathcal{B}_{\hat X}=\{\hat X^i\}_{i=1}^B \sim \mathbb{P}_{\theta},
\qquad
\mathcal{B}_{X}=\{X^j\}_{j=1}^B \sim \mathbb{P}_{\mathrm{data}},
\]
a consistent empirical estimator of the signature kernel score is
\begin{equation*}
\widehat{\mathcal{L}}_{\mathrm{sig}}
=
\frac{1}{B(B-1)}
\sum_{i\neq j}
k_{\mathrm{sig}}(\hat X^i,\hat X^j)
-
\frac{2}{B^2}
\sum_{i=1}^B \sum_{j=1}^B
k_{\mathrm{sig}}(\hat X^i,X^j).
\label{eq:sig_score_empirical}
\end{equation*}
This is the form used in practice for training.

\subsection{Hyperparameters optimization method}
\label{app:hyperopt}

\paragraph{Aggregation strategies} Aggregating the four loss terms defined in Section~\ref{sec:training} is non-trivial as they differ in scale and sign (e.g., MSE vs.\ log-likelihood). We compared four standard multi-objective weighting strategies \cite{vandenhende2021multi} to combine the individual losses into \(\mathcal{L}_{\mathrm{total}}\) (Eq.~\ref{eq:loss_total}): \emph{uncertainty-based weighting} \cite{kendall2018multi}, \emph{GradNorm} \cite{chen2018gradnorm}, \emph{initial-loss scaling}, and a \emph{linear weighted sum}. In our experiments, uncertainty weighting and GradNorm were unstable, likely due to the heterogeneous nature of the objectives. Initial-loss scaling gave reasonable results but was too sensitive to initialization. 
We therefore use a linear weighted combination
\begin{equation}
\mathcal{L}_{\mathrm{total}}
= \alpha_1 \mathcal{L}_{\mathrm{static}}
+ \alpha_2 \mathcal{L}_{\mathrm{event}}
+ \alpha_3 \mathcal{L}_{\mathrm{long,det}}
+ \alpha_4 \mathcal{L}_{\mathrm{long,sto}},
\qquad \alpha_j \ge 0,\quad \sum_{j=1}^4 \alpha_j = 1,
\label{eq:loss_total}
\end{equation}
which provided the most stable training and the best validation trade-off.

\paragraph{Pareto hyperparameter optimization} The weights \(\boldsymbol{\alpha}\) are tuned jointly with the other model hyperparameters using Pareto-front optimization with Optuna~\cite{akiba2019optuna}: each trial \(k\) trains the model on $\mathcal{L}_{\mathrm{total}}(\boldsymbol{\alpha}^{(k)})$ and records the unweighted validation losses $\{\mathcal{L}^{\mathrm{val}}_j\}_j$ as separate objectives. After approximately \(100\) trials, we select a configuration from the Pareto front that achieves the best trade-off across all validation objectives.
% , ensuring that longitudinal fidelity is not sacrificed for static reconstruction quality, and vice versa.
This prevents any single component from dominating optimization. 

\begin{figure}[h]
    \centering
    \includegraphics[width=0.55\linewidth]{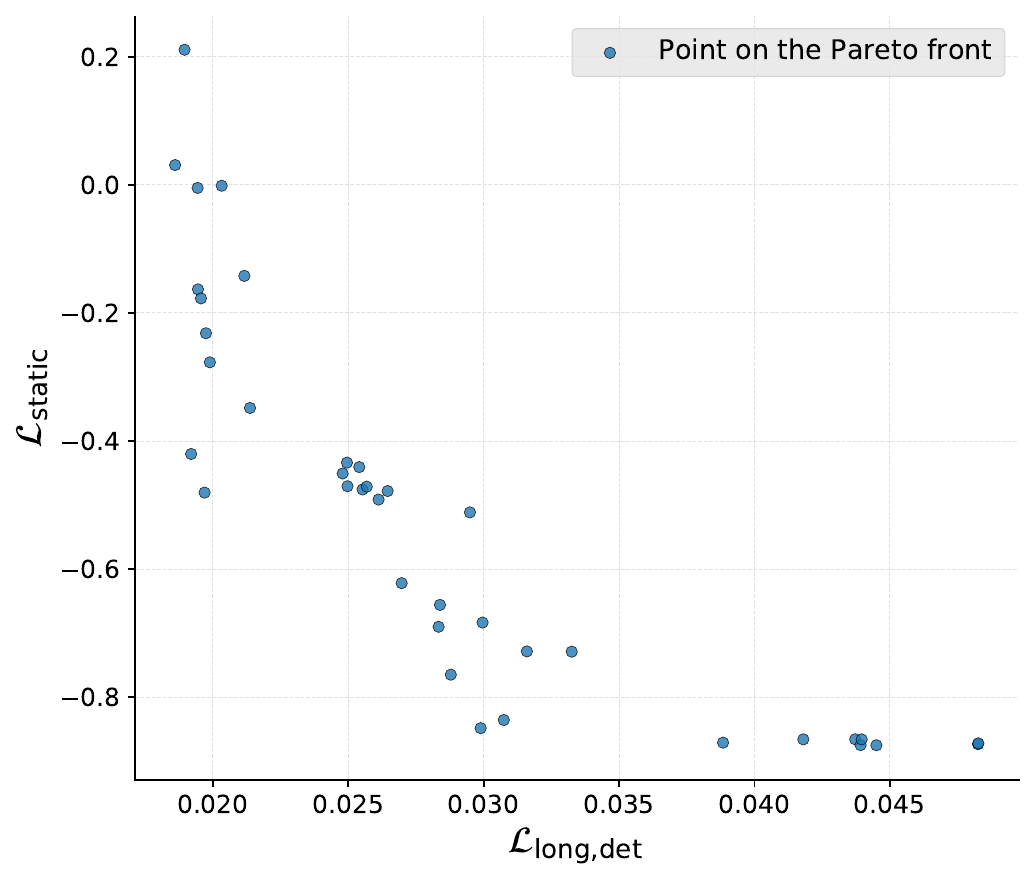}
    \caption{
        Pareto front between the longitudinal deterministic reconstruction loss $\mathcal{L}_{\mathrm{long,det}}$ and the static reconstruction loss $\mathcal{L}_{\mathrm{static}}$.
        Each point corresponds to a non-dominated hyperparameter configuration found by Optuna for our model on the reference simulated-data setting. 
    }
    \label{fig:pareto}
\end{figure}

\paragraph{Trade-off analysis}
Figure~\ref{fig:pareto} shows the Pareto front obtained from the hyperparameter optimization of our model on the reference simulated-data setting, in the $(\mathcal{L}_{\mathrm{long,det}}, \mathcal{L}_{\mathrm{static}})$ plane. It reveals a clear trade-off between the two objectives: configurations that improve longitudinal reconstruction tend to worsen static reconstruction, and vice versa. Its broad coverage of the objective space indicates that the model supports a diverse set of non-dominated solutions, rather than collapsing to a narrow region. We select an operating point on this front that provides a favorable compromise between the two objectives.

\paragraph{Hyperparameter values/search space}
Table~\ref{tab:hyperopt_space} summarizes the search space explored by Optuna across approximately 100 trials per dataset. Other hyperparameters are fixed across all experiments based on preliminary exploration and kept constant during Optuna tuning. The ODE drift and decoder use \texttt{Tanh} activations; the hypernetwork uses \texttt{ReLU}; the diffusion and residual drift networks use \texttt{LipSwish}~\cite{kidger2021neural}, which was found to improve stability of the SDE training. The ODE is solved with the \texttt{dopri5} adaptive-step solver (tolerances $\texttt{rtol}=10^{-3}$, 
$\texttt{atol}=10^{-4}$); the SDE uses Stratonovich integration with diagonal noise. Batch normalization is applied to static latent representations. Observation times are normalized to $[0, 1]$. 

\begin{table}[h]
\centering
\footnotesize
% \small
\caption{Hyperparameter search space.}
\label{tab:hyperopt_space}
\definecolor{rowgray}{gray}{0.9}
\setlength{\tabcolsep}{5pt}
\renewcommand{\arraystretch}{1.1}
\setlength{\heavyrulewidth}{1pt}
\setlength{\lightrulewidth}{1pt}
\begin{tabular}{lll}
\toprule
\textbf{Hyperparameter} & \textbf{Range / Values} & \textbf{Type}  \\
\midrule
Batch size                           & $\{128, 256\}$                                        & categorical \\
Learning rate                        & $\{10^{-4}, 10^{-3}, 2\cdot10^{-3}, 5\cdot10^{-3}, 10^{-2}\}$ & categorical \\
Static latent dims. $d_\mathrm{s}$, $d_\mathrm{u}$     & $[2,\, 8]$                          & integer \\
Longitudinal latent dim. $d_\mathrm{z}$       & $[2,\, d_\mathrm{z}]$                        & integer \\
Hidden dim.                          & $\{16, 32, 64, 128\}$                                 & categorical \\
ODE/SDE layers                       & $[1,\, 4]$                                            & integer \\
Hypernetwork layers                  & $[1,\, 5]$                                            & integer \\
Diffusion MLP layers                 & $[1,\, 3]$                                            & integer \\
Dropout (decoder)                    & $\{0,\, 0.1,\, 0.2\}$                                & categorical \\
Weight decay                         & $[10^{-7},\, 10^{-2}]$                               & log-uniform \\
SDE training kernel $\sigma^{\dagger}$         & $[0.01,\, 1.0]$                                       & log-uniform \\
Loss weight $\alpha_{\mathrm{det}}$  & $[0.1,\, 100]$                                        & log-uniform \\
Loss weight $\alpha_{\mathrm{sto}}$  & $[0.01,\, 100]$                                       & log-uniform \\
Loss weight $\alpha_{\mathrm{event}}$& $[0.001,\, 1]$                                           & log-uniform \\
\bottomrule
\multicolumn{3}{l}{$^{\dagger}$ Kernel bandwidth used in the signature-kernel 
training loss $\mathcal{L}_{\mathrm{sto}}$, distinct from the evaluation MMD bandwidth.}
\end{tabular}
\end{table}

% \clearpage
\FloatBarrier

\section{Evaluation metrics}
\label{app:metrics}

\subsection{Challenges and limitations of existing approaches}

Evaluating synthetic clinical data remains an open problem, particularly for irregularly sampled longitudinal data.
Most existing benchmarking frameworks and toolkits --- including Synthcity \cite{qian2023synthcity} and TSGBench \cite{ang2023tsgbench} --- are designed primarily for regularly sampled time series or tabular data. When applied to irregular longitudinal data, they typically require an interpolation or imputation step as preprocessing, which can substantially distort the original temporal structure and introduce systematic biases into the evaluation results.

Beyond tooling, there is no universally accepted evaluation protocol for synthetic data, even in the simpler tabular setting \cite{stenger2024evaluation, achterberg2024evaluation, kim2024synthesis, loni2025review}. The selection of an appropriate set of measures is recognized as an open challenge, and recent work has shown that results can be highly sensitive to implementation choices --- for instance, \cite{stenger2025steb} demonstrated a notable influence of the embedding strategy on embedding-based metrics for synthetic time series.

In practice, a large proportion of papers on time-series generation report only fidelity and utility metrics, often relying on qualitative visual inspection rather than rigorous quantitative evaluation \cite{alaa2022faithful, van2023generating}. Sampling regularity, event dependence, and cross-modal structure are almost never evaluated. Privacy, despite being a central requirement in healthcare, remains inconsistently assessed \cite{miletic2025synthetic}.

Our work requires evaluating jointly static covariates and irregular longitudinal trajectories, while accounting for their cross-modal dependencies, the event-driven observation process, and missingness.
This combination falls outside the scope of most existing toolkits and metrics. Developing new metrics fully adapted to this setting is itself a research problem beyond the scope of this paper; however, we believe this discussion highlights a real and underappreciated gap in the literature that method papers in this space should explicitly address.

\subsection{Our evaluation strategy}

We evaluate synthetic data quality along three complementary axes ---
\emph{fidelity}, \emph{utility}, and \emph{privacy} --- using metrics that are adapted to the multimodal and irregular nature of our setting.
When standard metrics are available, we use modality-aware variants that operate directly on irregular observations without interpolation or imputation. We describe each metric precisely below.

\subsubsection{Fidelity metrics}

\paragraph{Maximum Mean Discrepancy (MMD)}
We use the Maximum Mean Discrepancy~\cite{gretton2012kernel} to compare the distributions of real and synthetic samples. Given two sample sets $X = \{x_i\}_{i=1}^n$ and $\tilde{X} = \{\tilde{x}_j\}_{j=1}^m$, the empirical squared MMD is
\[
\mathrm{MMD}^2(X, \tilde{X})
=
\frac{1}{n(n-1)} \sum_{i \neq i'} k(x_i, x_{i'})
+ \frac{1}{m(m-1)} \sum_{j \neq j'} k(\tilde{x}_j, \tilde{x}_{j'})
- \frac{2}{nm} \sum_{i,j} k(x_i, \tilde{x}_j),
\]
where $k$ is a positive-definite kernel. For static variables, we use a standard RBF kernel on the static covariates. For longitudinal data, we use a signature-kernel MMD \cite{chevyrev2022signature, salvi2021signature}, which is well suited to irregularly sampled trajectories: paths are time-augmented and compressed according to the observation mask, enabling comparisons directly on irregular trajectories. Lower values indicate better agreement between real and synthetic distributions.

The signature-kernel MMD with an RBF static kernel, computed with \texttt{PySigLib}~\cite{shmelev2025pysiglib}, depends critically on the bandwidth parameter \(\sigma\). If \(\sigma\) is poorly matched to the scale of pairwise path discrepancies in signature space, the kernel values saturate, and the metric loses discriminative power. In this regime, the empirical estimator may also become numerically unstable; in particular, the unbiased estimate of MMD\(^2\) can take negative values at finite sample size, and this effect may be exacerbated in practice by long, sparse, and highly irregular trajectories.
To improve stability, we select \(\sigma\) using a median heuristic applied after time augmentation and path compression, and we linearly interpolate trajectories between observed points before computing the signature kernel. The latter ensures that the evaluated paths are piecewise linear, but sparse and irregular sampling may still make the interpolated paths only rough proxies for the underlying continuous trajectories. As a result, the signature-kernel MMD should be interpreted with caution, especially when comparing models that differ substantially in smoothness or in the induced observation patterns.

\paragraph{Correlation RMSE}
To assess whether synthetic data preserve pairwise dependency structure, we compare real and synthetic correlation matrices:
\[
\mathrm{Corr.\ RMSE}
= \sqrt{
    \frac{1}{|\Omega|}
    \sum_{(p,q) \in \Omega}
    \left( C^{\mathrm{real}}_{pq} - C^{\mathrm{syn}}_{pq} \right)^2
},
\]
where $\Omega$ denotes the set of valid entries. For static variables, correlations are computed after type-aware preprocessing (standardization for continuous, rank encoding for ordinal, integer encoding for categorical). For longitudinal data, we flatten across individuals and time and compute a masked correlation matrix using only overlapping
observed pairs, excluding entries with insufficient overlap. We also report a cross-modal Corr.\ RMSE computed from a joint masked correlation matrix augmented with static covariates. Lower values indicate better preservation of dependency structure. 
% \perrine{Peux-être que le fait qu'il n'y ait pas de valeurs manquantes dans MutliNODEs explique son meilleur résultat, car plus de points overlapped?}

% \paragraph{\perrine{A modifier} KL divergence on the observation process (KL $\Delta_t$)}
% To evaluate whether synthetic data reproduce the irregular observation process, we compare visit pattern distributions. Let $p_{\mathrm{real}}$ and $p_{\mathrm{syn}}$ denote the empirical distributions of inter-visit times $\Delta t$. We compute
% \[
% D_{\mathrm{KL}}\!\left(p_{\mathrm{real}} \,\|\, p_{\mathrm{syn}}\right)
% = \sum_t p_{\mathrm{real}}(t) \log \frac{p_{\mathrm{real}}(t)}{p_{\mathrm{syn}}(t)}.
% \]
% We report the KL divergence on inter-visit times (KL $\Delta_t$), which more directly captures the irregularity of the observation process than absolute event times. Lower values indicate better agreement with the real visit process.

\paragraph{KL divergence on the observation intensities (KL $\lambda$)}
To evaluate whether synthetic data reproduce the irregular and informative observation process, we compare the empirical event intensities of the real and synthetic datasets. For each observed variable and for the end-of-sequence (EOS) process, we estimate a piecewise-constant population intensity as the number of events divided by the corresponding person-time at risk,
\[D_k
=
\sum_b \frac{E_b^R}{N_R}
\left[
\lambda_{k,b}^R
\log\frac{\lambda_{k,b}^R}{\lambda_{k,b}^S}
-\lambda_{k,b}^R+\lambda_{k,b}^S
\right]
\]
where $b$ indexes the time bins, $E_b^{R}$ denotes the total real-data person-time exposure in bin $b$, $N_R$ is the number of real subjects, and $\lambda_{k,b}^{R}$ and $\lambda_{k,b}^{S}$ are the corresponding empirical real and synthetic intensities.
We report the component-averaged Poisson intensity divergence, \(\frac{1}{d_x+1}
\left(
\sum_{\ell=1}^{d_x}D_\ell+D_{\mathrm{EOS}}
\right)\),
as KL $\lambda$. It captures discrepancies in both the magnitude and temporal shape of the observation intensities while explicitly accounting for the EOS termination process. Lower values indicate better agreement between the real and synthetic observation processes.

\paragraph{Discriminative score}
As a joint fidelity measure, we use a classifier two-sample test (C2ST)~\cite{yoon2019time}: a classifier is trained to distinguish real from synthetic samples, taking as input the longitudinal trajectories (with last-observation-carried-forward imputation) augmented with static covariates. The classifier is based on an S4 architecture \cite{gu2022efficiently} with masked global average pooling, ensuring predictions depend only on observed time steps. We report the test AUROC; values closer to $0.5$ indicate that synthetic samples are harder to distinguish from real data, and hence better fidelity.

\subsubsection{Utility metrics}

\paragraph{Predictive score}
We assess downstream utility through a forecasting task under a Train-on-Synthetic-Test-on-Real (TSTR) protocol~\cite{esteban2017real}, compared with the Train-on-Real-Test-on-Real (TRTR) baseline. A predictor based on an S4 architecture \cite{gu2022efficiently, el-laham2025variational}, is trained to predict $m$ future time points ($\mathbf{x}_i(t_{L+1:L+m})$) from a observed prefix ($\mathbf{x}_i(t_{1:L})$) concatenated with static covariates. Missing values are imputed via last-observation-carried-forward. Performance is measured using Mean Squared Error (MSE) on real test data. Synthetic data yields better downstream utility when its TSTR MSE closely matches the Train-on-Real (TRTR) reference.

\subsubsection{Privacy metrics}

\paragraph{Nearest-Neighbor Adversarial Accuracy (NNAA)}
NNAA~\cite{yale2020generation} measures whether the synthetic data distribution has systematically overfitted to the training data. For two datasets $R$ (real) and $S$ (synthetic), the adversarial accuracy (AA) is defined as the proportion of samples that are closer to a neighbor in their own set than to a neighbor in the opposite set:
\[
\mathrm{AA}(R, S) = \frac{1}{2} \left( \frac{1}{|R|}\sum_{r \in R} \mathbb{I}(d_{RS}(r) > d_{RR}(r)) + \frac{1}{|S|}\sum_{s \in S} \mathbb{I}(d_{SR}(s) > d_{SS}(s)) \right)
\]
where $d_{AB}(x)$ is the distance from $x \in A$ to its nearest neighbor in $B$. If $S$ simply memorizes $R$, $\mathrm{AA}$ drops below 0.5. We compute this quantity once using the training set ($\mathrm{AA}_{\text{train}}$) and once using a held-out test set ($\mathrm{AA}_{\text{test}}$). The final metric is $\mathrm{NNAA} = \mathrm{AA}_{\text{test}} - \mathrm{AA}_{\text{train}}$. Large positive values indicate that synthetic samples are systematically closer to training data than to unseen real data, suggesting higher memorization risk. Distances are computed separately for static and longitudinal modalities (using pairwise Euclidean and signature-based distances, respectively). For the multimodal NNAA score, the modality-specific distance matrices are normalized and combined via a weighted sum. Lower NNAA values indicate lower privacy risk.

\paragraph{Membership Inference Risk (MIR)}
MIR~\cite{shokri2017membership} quantifies the vulnerability of the synthetic dataset to a membership inference attack, assessing whether an adversary can distinguish training samples from held-out test samples based strictly on their distance to the synthetic data. For each real sample $x$, we compute an attack score $s(x) = - d_{k}(x, S)$, where $d_{k}$ is the aggregated distance (e.g., the minimum or mean) to the $k$ nearest synthetic neighbors. Treating training samples as the positive class and test samples as the negative class, we evaluate the discriminative power of $s(x)$ using the AUROC. We report the rescaled risk:
\[
\mathrm{MIR} = 2\,|\mathrm{AUROC} - 0.5|
\]
so that $\mathrm{MIR} \in [0, 1]$, where $0$ corresponds to random guessing (no privacy leak). As with NNAA, MIR is computed per modality and then aggregated in the same manner. Lower MIR values indicate lower privacy risk.

\paragraph{Dependence Score}
We report a dependence score based on the temporal autocorrelation structure. For each longitudinal variable and lag $k$, we compute a dataset-level autocorrelation by averaging patient-level Pearson autocorrelations over jointly observed pairs, yielding a matrix $R \in \mathbb{R}^{V \times (K+1)}$. The dependence score is the Mean Squared Error (MSE) between the real and synthetic matrices:
\[
\mathrm{Dep.\ Score}
= \frac{1}{|\Omega|}
\sum_{(v,k) \in \Omega}
\left( R^{\mathrm{real}}_{v,k} - R^{\mathrm{syn}}_{v,k} \right)^2
\]
where $\Omega$ is the set of valid lag entries up to a maximum lag $K$. 
This metric serves as a critical dual-purpose diagnostic. Primarily, it acts as a fidelity measure: generative models that over-smooth the data fail to capture natural trajectory noise, resulting in an artificially high synthetic autocorrelation and, consequently, a large MSE. Conversely, recent benchmarking by Stenger et al.~\cite{stenger2025steb} empirically shows that autocorrelation error is also a top-performing proxy for privacy risk. If the error is suspiciously close to zero, it indicates the model is likely memorizing the exact trajectory noise of specific training samples rather than learning the underlying distribution. Thus, an optimal model must achieve a low autocorrelation error without exactly replicating the training data.

\FloatBarrier

\section{Details on datasets} \label{app:datasets}

\subsection{Formatting and preprocessing of the data} \label{app:preprocessing}

To enable batch processing while preserving irregular sampling patterns, we first define a common time grid as the union of all individual observation times:
\[
\mathcal{T} = \bigcup_{i=1}^N \bigcup_{\ell=1}^{d_\mathrm{x}} \mathcal{T}_i^{(\ell)}.
\]
Each individual trajectory is then represented on this grid, yielding a three-dimensional tensor
\[
\mathbf{X} \in \mathbb{R}^{N \times |\mathcal{T}| \times d_\mathrm{x}},
\]
where unobserved entries are padded with zeros. To explicitly encode missingness, we introduce a binary mask \(\mathbf{M} \) of the same dimension, where \(\mathbf{M}_{i,t,\ell}=1\) if \(\mathbf{x}_i^{(\ell)}(t)\) is observed and \(\mathbf{M}_{i,t,\ell}=0\) otherwise.
Baseline measurements of the longitudinal variables, \(\mathbf{x}_i^{(\ell)}\bigl(t^{(\ell)}_{i,0}\bigr)\), are extracted and appended to the static covariates, resulting in an augmented static feature vector of dimension \(d_\mathrm{W}+d_\mathrm{x}\). These baseline values are then subtracted from the corresponding longitudinal trajectories, so that each time series represents deviations from baseline.
Finally, each longitudinal variable is rescaled by its standard deviation computed on the training set. This normalization puts variables on comparable scales and improves numerical stability during training.

\subsection{Simulated datasets} \label{app:simulation}

We simulate multimodal datasets in which static covariates explicitly control longitudinal dynamics. 

\paragraph{Static variables} For each individual $i$, we generate static variables
\[
\mathbf{W}_i = [\mathbf{W}_{i,\mathrm{num}}, \mathbf{W}_{i,\mathrm{bin}}],
\]
where $\mathbf{W}_{i,\mathrm{num}}$ denotes numerical variables and $\mathbf{W}_{i,\mathrm{bin}}$ binary variables. 
The numerical components are sampled as
\[
\mathbf{W}_{i,\mathrm{num}} \sim \mathcal{N}(\mathbf{0}, \sigma_{\mathrm{num}}^2 \mathbf{I}).
\]
The binary components are generated through a probit model. Let
\[
\tilde{\mathbf{W}}_{i,\mathrm{bin}} \sim \mathcal{N}(\mathbf{0}, \sigma_{\mathrm{bin}}^2 \mathbf{I}),
\qquad
u_{i,j} \sim \mathcal{U}(0,1),
\]
and let $c_j \sim \mathcal{N}(0,1)$ denote fixed coefficients. Then, for each binary dimension $j$,
\[
\mathbf{W}_{i,\mathrm{bin}}^{(j)} = \mathbf{1}\!\left[u_{i,j} < \Phi\!\left(\tilde{\mathbf{W}}_{i,\mathrm{bin}}^{(j)} c_j\right)\right],
\]
where $\Phi$ is the standard normal cumulative distribution function.
In the experiments reported in the main paper, we use $d_W=2$, corresponding to one numerical and one binary feature, $\sigma_\mathrm{num}^2 = 0.07$ and $\sigma_\mathrm{bin}^2 = 0.01$.

\paragraph{Longitudinal variables} Conditioned on $\mathbf{W}_i$, the longitudinal process $\mathbf{x}_i(t) \in \mathbb{R}^{d_\mathrm{x}}$ follows independent fractional Ornstein--Uhlenbeck dynamics,
\[ d \mathbf{x}_i^{(\ell)}(t) = -a_{i,\ell}\bigl(\mathbf{x}_i^{(\ell)}(t)-b_{i,\ell}\bigr)\,dt + \zeta\, dB^H_{i,\ell}(t), 
\qquad \ell \in \{1,\dots,d_\mathrm{x}\},
\]
where $B^H_{i,\ell}(t)$ is a fractional Brownian motion with Hurst parameter $H \in (0,1)$. The subject-specific parameters are obtained from the static covariates through linear maps:
\[
a_{i,\ell} = s\,(\Omega^{(a)} \mathbf{W}_i)_\ell,
\qquad
b_{i,\ell} = s\,(\Omega^{(b)} \mathbf{W}_i)_\ell,
\]
with $\Omega^{(a)}, \Omega^{(b)} \in \mathbb{R}^{d_\mathrm{x} \times d_W}$ random weight matrices and $s \in \mathbb{R}$ a fixed scaling factor. In our experiments, we set $d_\mathrm{x} = 3, \zeta = 0.1$ and $H = 0.4$.

\paragraph{Observation-time points}
To mimic irregular clinical monitoring, observation times are generated from an inhomogeneous Poisson process on \([0,T]\) with intensity \(\lambda_i(t)\). We consider two observation schemes.

\emph{Time-dependent sampling.} In the baseline setting, the observation process depends only on time:
\begin{equation} \label{eq:time_dep_sampling_lambda}
    \lambda_i(t)
= \alpha + (1-\alpha)\frac{1}{2}\sin\!\left(\frac{-2\pi t}{T}\right),
\end{equation}
where \(T=7\) in our experiments, and the constant \(\alpha\) is calibrated to obtain the desired average number of observations per individual. 

\emph{State-dependent sampling.} In the informative-sampling setting, the observation intensity depends on the underlying trajectory:
\begin{equation} \label{eq:state_dep_sampling_lambda}
    \lambda_i(t)
=
c\,\exp\!\left(-\sum_{\ell=1}^{d_\mathrm{x}} \mathbf{x}_i^{(\ell)}(t)\right),
\end{equation}
where \(c>0\) is chosen so that the expected number of observations matches a prescribed target. 

In both settings, visit times are generated by a discrete thinning algorithm applied on the regular grid, and latent trajectories are first simulated on a fine regular grid and then observed only at the sampled visit times.

In the main experiments, repeated Monte Carlo runs correspond to independently generated datasets obtained with different random seeds.

\subsection{Real-world datasets}
\label{app:realworld_datasets}

\subsubsection{Preprocessing pipeline}

All real-world datasets were preprocessed using the same general pipeline, with minor dataset-specific adaptations. We first separated baseline static covariates from longitudinal variables using dataset documentation and empirical missingness patterns. Static variables correspond to time-invariant or quasi-time-invariant patient characteristics, while longitudinal variables correspond to repeatedly measured clinical quantities.

We then applied the following steps. \emph{(i) Static variables:} sentinel values were replaced by missing values; nominally static variables recorded multiple times were aggregated at the patient level using the median; variables observed in fewer than $50\%$ of patients were discarded. \emph{(ii) Longitudinal variables:} original observation times were preserved; duplicate patient-time entries were averaged; rows with no observed longitudinal variable were removed; variables with overall presence below a dataset-specific threshold were discarded (typically $30\%$ or $50\%$ depending on the dataset). \emph{(iii) Patient filtering:} we retained only patients with both static and longitudinal information and removed patients with fewer than 5 longitudinal observation times after filtering. \emph{(iv) Time normalization:} observation times were shifted so that each patient's first visit occurred at time zero (or study start for VELOUR), and a common fine temporal grid was defined as the union of all observed times. Longitudinal data were finally represented on this grid together with a binary observation mask. When needed, skewed positive variables were log-transformed before standardization.

The final numbers of retained patients, static variables, longitudinal variables, and grid points are reported in Table~\ref{tab:dataset_summary}.

\begin{table*}[h]
\centering
\footnotesize
\caption{Summary statistics of the real-world datasets after preprocessing.}
\label{tab:dataset_summary}
\definecolor{rowgray}{gray}{0.9}
\setlength{\tabcolsep}{5pt}
\renewcommand{\arraystretch}{1.1}
\setlength{\heavyrulewidth}{1pt}
\setlength{\lightrulewidth}{1pt}
\begin{tabular}{lcccccc}
\toprule
\multirow{2}{*}{\textbf{Dataset}}  & \multirow{2}{*}{\textbf{Patients}} & \textbf{Long.} & \textbf{Static} & \textbf{Common} & \textbf{Avg.\ observed} & \textbf{Time span} \\
\noalign{\vspace{-2pt}}
 &  & \textbf{variables} & \textbf{variables} & \textbf{grid points} & \textbf{visits/patient} & \textbf{(days)} \\
 \midrule
 VELOUR       &  526 & 23 & 26 &  673 & 21.3 &   23--990 \\
\rowcolor{rowgray}
PPMI          & 1930 & 15 & 20 &  686 & 11.2 & 153--5448 \\
% PhysioNet 2012 & 3936 &  9 & 19 & 2881 & 62.3 &     0--2 \\
% \rowcolor{rowgray}
% MSK-Chord     & 2924 &  2 & 21 & 7306 & 28.8 &  47--9164 \\
\bottomrule
\end{tabular}
\end{table*}

\subsubsection{Dataset-specific notes}

\paragraph{VELOUR \cite{velour2021, projectdatasphere}}
We use data from the VELOUR phase III randomized trial in metastatic colorectal cancer. Static covariates include demographics (sex, race, geographic region, age category), baseline ECOG performance status, height, and binary indicators for the 10 most frequent adverse events and concomitant medications recorded at baseline. Retained longitudinal variables correspond to standard oncology laboratory panels (hepatic enzymes, blood counts, metabolic markers) and vital signs. Observations prior to treatment initiation were excluded, and highly right-skewed laboratory variables were log-transformed before standardization. VELOUR can be accessed via Project DataSphere (\url{https://data.projectdatasphere.org}), which requires registration and a data use agreement.

\paragraph{PPMI \cite{marek2011parkinson}}
We use participants from the Parkinson's Progression Markers Initiative, a multi-year observational cohort on Parkinson's disease progression. Static covariates include demographics (age, sex), enrollment cohort, genetic risk factors (LRRK2, SNCA status), and family history of Parkinson's disease. Retained longitudinal variables include vital signs (temperature, supine and standing blood pressure, and heart rate) and clinical rating scales (UPDRS parts 1--3, ESS, SCOPA-AUT, STAI). PPMI is available 
upon registration at \url{https://www.ppmi-info.org}.

% \paragraph{PhysioNet 2012 \cite{silva2012predicting}}
% We use set A of the PhysioNet/CinC 2012 challenge, comprising ICU patients monitored over up to 48 hours. Static covariates include demographics (age, sex, height, weight, ICU type) and admission laboratory values (BUN, creatinine, glucose, HCO$_3$, HCT, K, Mg, Na, platelets, WBC, PaCO$_2$, PaO$_2$, pH, lactate). Longitudinal variables retained after filtering include heart rate, invasive and non-invasive blood pressure (systolic, diastolic, mean), temperature, and urine output.

% \paragraph{MSK-Chord \cite{jee2024automated}}
% We restrict the MSK-CHORD oncology registry to prostate cancer patients. Static covariates include demographics (sex, race, ethnicity, age), disease characteristics (cancer stage, Gleason scores, tumor sites), and survival-related information. The longitudinal variable retained after filtering is PSA; CA-15, CA-19, and CEA are excluded due to very low presence.

% \clearpage
\FloatBarrier

\section{Experimental setup} \label{app:exp_setup}

\subsection{Experiments setting and compute resources} \label{app:exp_setting_resources}
All models are trained using AdamW~\cite{Loshchilov2017DecoupledWD}. HyperNSDE is trained with early stopping (patience 10, maximum 1000 epochs). All experiments were run on the Jean Zay cluster using single-GPU jobs on  NVIDIA V100 GPU with 32\,GB of memory, 10 CPU cores per task. and a wall-clock limit of 20 hours per job. On the reference simulated dataset, one epoch takes approximately 1\,min\,20\,s, with training terminating around epoch 500 ($\approx$11 hours). Training time scales with the number of patients and observations for real-world datasets.

\subsection{Baseline implementations and tuning} \label{app:baselines}

We used the public implementations of MultiNODEs \cite{wendland2022generation}, RTSGAN \cite{pei2021rtsgan}, and DGBFGP \cite{balik2025bayesian}, with dataset-specific preprocessing adapted to our setting.\footnote{Official repositories: MultiNODEs \url{https://github.com/SCAI-BIO/MultiNODEs/tree/main}, RTSGAN \url{https://github.com/acphile/RTSGAN}, and DGBFGP \url{https://github.com/YigitBalik/DGBFGP}.}

\paragraph{Fair hyperparameter tuning}
To ensure a fair comparison with our model, we did not rely solely on the default hyperparameter settings provided in the public implementations. Whenever such a procedure was not already implemented in the original code, we performed an additional hyperparameter search using Optuna \cite{akiba2019optuna}. Each baseline was tuned using its native training objective, evaluated on the validation set, as a single optimization target. This additional tuning step was introduced to reduce the risk that performance differences arise from unequal optimization effort rather than from the underlying modeling assumptions.

\paragraph{DGBFGP + HI-VAE}
Since DGBFGP does not provide a generative model for static covariates, we coupled its longitudinal component with an HI-VAE trained separately on the static variables. As a result, static and longitudinal generation are not jointly learned in this baseline, unlike in our model and MultiNODEs. For this reason, and because DGBFGP requires substantial dataset-specific manual configuration, we restrict this baseline to the simulated-data experiments.

\subsection{Details on the sensitivity analysis}
\label{app:sensitivy_analysis}

We assess the robustness of all methods under several controlled modifications of the reference setting. We consider the following sensitivity scenarios. 
\begin{itemize}[leftmargin=0.5cm]
    \item[] \textbf{\emph{(i) Limited training data:}} to evaluate sensitivity to training sample size, we reduce the training set to $N_{\mathrm{train}}=200$ while keeping the test set fixed at $N_{\mathrm{test}}=200$. 
    \item[] \textbf{\emph{(ii) Sparse observation regime:}} to test robustness to increased irregularity and missingness, we keep the standard 80/20 split but reduce the average number of visits to $n_{\mathrm{obs}}=15$ per patient. 
    \item[] \textbf{\emph{(iii) State-dependent observation process:}} we replace the baseline time-dependent intensity by a more complex event model in which the visit intensity $\lambda$ depends on the longitudinal state $\mathbf{x}$ defined in Equation~\ref{eq:state_dep_sampling_lambda}, thereby introducing informative sampling. Unless otherwise stated, all other simulation parameters are kept identical to the reference setting.
    \item[] \textbf{\emph{(iv) Prior versus posterior sampling:}} to assess sensitivity to the static latent sampling strategy, we compare \emph{posterior sampling (post.)} and \emph{prior sampling (prior)}. Under posterior sampling, static latent representations are drawn from individual approximate posteriors, pooled, shuffled, and subsampled to form a synthetic cohort without preserving patient correspondence. Under prior sampling, they are drawn directly from the learned HI-VAE prior. We report posterior sampling in the main experiments for consistency with MultiNODEs and use this sensitivity analysis to assess whether the results depend on this choice.
    \item[] \textbf{\emph{(v) State dependence of the observation intensity:}} to assess the contribution of conditioning the observation process on the latent state, we compare the standard HyperNSDE intensity $\lambda_i(t)=\exp\{\xi(\mathbf{z}_i(t),t;\theta_\xi)\}$ with a state-independent variant in which the intensity depends only on time, $\lambda_i(t)=\exp\{\xi(t;\theta_\xi)\}$. For the simulated data case, we use the simulated data of the sensitivity analysis \emph{(iii)}. All other model components and training procedures are unchanged.
\end{itemize}

\subsection{Details on the ablation studies} \label{app:ablations}

All ablation studies are conducted on the reference simulated-data setting described in Section~\ref{sec:simulated_datasets_ref}. Unless otherwise stated, all remaining architectural choices, hyperparameters, and evaluation protocols are kept unchanged.

\begin{itemize}[leftmargin=0.5cm]
    \item[] \textbf{\emph{(AS.1) Deterministic latent dynamics}.}
In this variant, we replace the latent Neural SDE by a latent Neural ODE, thereby removing the stochastic component of the latent dynamics. The latent trajectory satisfies
\[
d\mathbf{z}_i(t)=b_i\bigl(\mathbf{z}_i(t),t\bigr)\,dt,
\qquad \mathbf{z}_i(0)=\mathbf{0}.
\]
This ablation evaluates whether explicitly modeling stochasticity in latent space improves generative performance compared with a purely deterministic trajectory model. It is close in spirit to MultiNODEs, but adapted to our framework without an encoder and with the same event-intensity module.

    \item[] \textbf{\emph{(AS.2) Direct stochastic training without decomposition}.}
In the proposed model, the latent path is decomposed into a deterministic component and a stochastic residual, each trained with a separate objective. In this ablation, we remove this decomposition and model the full latent path directly as a Neural SDE. The stochastic model is then trained directly from the signature-kernel discrepancy loss on the full latent trajectory. This ablation tests whether the proposed decomposition improves optimization stability and disentangles deterministic dynamics from stochastic fluctuations.

    \item[] \textbf{\emph{(AS.3) Alternative conditioning of longitudinal dynamics}.}
We compare the proposed hypernetwork based personalization with two simpler strategies for incorporating the static latent representation \(\mathbf{h}_i\) into the longitudinal dynamics.

\begin{itemize}[leftmargin=0.5cm]
\item \textbf{\emph{MNSDE:}} Inspired by MultiNODEs, we concatenate the static latent representation only to the initial latent state:
\[
\mathbf{z}_{i,0}^{\mathrm{aug}}
=
\begin{pmatrix}
\mathbf{z}_i(0) \\
\mathbf{h}_i
\end{pmatrix}
\in \mathbb{R}^{d_z+d_h}.
\]
The augmented process is then evolved according to
\[
d\mathbf{z}_i^{\mathrm{aug}}(t)
=
b\bigl(\mathbf{z}_i^{\mathrm{aug}}(t),t;\theta_b\bigr)\,dt
+
\sigma\bigl(\mathbf{z}_i^{\mathrm{aug}}(t), t;\theta_\sigma\bigr)\,dB_t.
\]
In this formulation, static information is only injected at \(t=0\).

\item \textbf{\emph{SNSDE:}} Inspired by modulated dynamics models such as \cite{auzina2023modulated}, we concatenate the static latent representation to the latent state throughout the trajectory:
\[
d\mathbf{z}_i(t)
=
b\bigl(\mathbf{z}_i(t),t,\mathbf{h}_i;\theta_b\bigr)\,dt
+
\sigma\bigl(\mathbf{z}_i(t),t;\theta_\sigma\bigr)\,dB_t.
\]
In practice, this amounts to providing \(\mathbf{h}_i\) as an input to the drift network at all time points. This ablation evaluates whether the proposed hypernetwork parameterization is more effective than direct concatenation-based conditioning.

\item \textbf{\emph{FNSDE:}} We additionally consider FiLM-based personalization, where \(\mathbf{h}_i\) generates layer-wise scale and shift parameters that modulate each hidden layer of the shared drift network. Unlike the hypernetwork, which generates the full drift parameters, FiLM only generates feature-wise affine modulations, providing a lower-dimensional personalization mechanism.
\end{itemize}

    \item[] \textbf{\emph{(AS.4) No observation-process modeling}.}
In this variant, we remove the latent-state-dependent event-intensity model, including the explicit end-of-sequence (EOS) component. This ablation isolates the contribution of explicitly modeling the observation process.
\end{itemize}

\FloatBarrier

\section{Additional results}
\label{app:additional_results}

% \subsection{Additional results on simulated data}

\subsection{Statistical variability and robustness}
\label{app:stat_var_robustness}

We complement the mean results reported in the main text with measures of variability, paired significance tests, and median-based aggregation. These analyses assess the stability of the reported comparisons across Monte Carlo runs and generated datasets.

\begin{table*}[h]
    \centering
    \scriptsize
    \caption{\textbf{Evaluation metrics on simulated and real-world data --- standard deviations.} 
    Standard deviations computed across $N_{\mathrm{MC}}=30$ Monte Carlo runs (for simulated data only) and $N_{\text{gen}}=50$ generated datasets from a single trained model. Corresponding means are reported in Table~\ref{tab:all_results}.}
    \label{tab:all_results_std}
    \definecolor{rowgray}{gray}{0.9}
    \setlength{\tabcolsep}{4pt}
    \renewcommand{\arraystretch}{1.3}
    \setlength{\heavyrulewidth}{1pt}
    \setlength{\lightrulewidth}{1pt}
    \makebox[\textwidth][c]{%
    %
    % ---- LEFT: dataset label column ----
    \begin{tabular}[t]{r}
    \phantommid
        \\
        \phantomcmid
        \noalign{\vspace{-2pt}}
        \\[-3pt]
        \phantomcmid
        \noalign{\vspace{-2.8pt}}
        \\[-2pt]
        \phantommid
        \multirow{5}{*}{\rotatebox{90}{\textbf{Simulated}}} \\[-2pt]
        \phantomcmid
        \\
        \\
        \\
        \\
        \arrayrulecolor{white}\midrule\arrayrulecolor{black}
        \multirow{3}{*}{\rotatebox{90}{\textbf{VELOUR}}} \\
        \\
        \\
        \arrayrulecolor{white}\midrule\arrayrulecolor{black}
        \multirow{3}{*}{\rotatebox{90}{\textbf{PPMI}}} \\
        \\
        \\
        % \arrayrulecolor{white}\midrule\arrayrulecolor{black}
        % \multirow{4}{*}{\rotatebox{90}{\textbf{PhysioNet 2012}}} \\[-2pt]
        % \phantomcmid
        % \\
        % \\
        % \\
        % \arrayrulecolor{white}\midrule\arrayrulecolor{black}
        % \multirow{4}{*}{\rotatebox{90}{\textbf{MSK-CHORD}}} \\[-2pt]
        % \phantomcmid
        % \\
        % \\
        % \\
        \phantommid
    \end{tabular}%
    %
    % ---- RIGHT: main table ----
    \begin{tabular}[t]{lcccccccccc}
\toprule
  & \multicolumn{6}{c}{\textbf{Fidelity}} & \multicolumn{3}{c}{\textbf{Privacy}} & \textbf{Utility} \\
\cmidrule(lr){2-7} \cmidrule(lr){8-10} \cmidrule(lr){11-11}
\noalign{\vspace{-2pt}}
   & \multicolumn{2}{c}{\textbf{Corr. RMSE}}
    & \multirow{2}{*}{\textbf{Discrim.}}
    & \multirow{2}{*}{\textbf{KL $\lambda$}}
    & \multicolumn{2}{c}{\textbf{MMD}}
    & \multirow{2}{*}{\textbf{Dep.}}
    & \multirow{2}{*}{\textbf{MIR}}
    & \multirow{2}{*}{\textbf{NNAA}}
    & \multirow{2}{*}{\textbf{Pred.}} \\[-3pt]
\cmidrule(lr){2-3} \cmidrule(lr){6-7}
\noalign{\vspace{-2.8pt}}
  & \textbf{Long.} & \textbf{Static} & & & \textbf{Long.} & \textbf{Static} & & & & \\[-2pt]
%%% Simulated
\midrule
\rowcolor{white} 
 Reference        & $0.047$ & $0.037$ & $0.069$ & $0.031$ & $0.321$   & $0.004$ & $0.000$ & $0.027$ & $0.044$ & $1.785$ \\ [-2pt]
\cmidrule(lr){1-11}
\rowcolor{rowgray}
 MultiNODEs    & $0.070$ & $0.038$ & $0.006$ & $0.413$ & $1717.726$ & $0.004$ & $0.044$ & $0.026$ & $0.035$ & $5.251$ \\
\rowcolor{white}
 RTSGAN        & $0.071$ & $0.057$ & $0.120$ & $12.336$ & $66.257$   & $0.024$ & $0.033$ & $0.029$ & $0.040$ & $2.784$ \\
\rowcolor{rowgray}
 DGBFGP + HIVAE & $0.068$ & $0.039$ & $0.003$ & $0.413$ & $1150.507$ & $0.004$ & $0.023$ & $0.027$ & $0.033$ & $5.430$ \\
\rowcolor{white}
 HyperNSDE & $0.070$ & $0.043$ & $0.134$ & $0.213$ & $3.382$ & $0.003$ & $0.011$ & $0.029$ & $0.034$ & $2.114$ \\
%%% VELOUR
\midrule
\rowcolor{white}
 MultiNODEs  & ${0.007}$ & ${0.005}$ & $0.010$ & ${0.000}$ & ${0.192}$ & ${0.000}$ & ${0.004}$ & $0.030$ & $0.044$ & ${0.162}$    \\
\rowcolor{rowgray}
 RTSGAN   & $0.008$ & $0.006$ & ${0.000}$ & ${1.003}$ & $736.300$ & $0.000$ & $0.020$ & ${0.028}$ & ${0.031}$ & $1427.366$   \\
\rowcolor{white}
 HyperNSDE & ${0.007}$ & ${0.005}$ & ${0.000}$ & $1.293$ & ${0.005}$ & ${0.000}$ & ${0.007}$ & ${0.029}$ & ${0.033}$ & ${1.120}$   \\

%%% PPMI
\midrule
\rowcolor{white}
 MultiNODEs      & $0.006$ & ${0.003}$ & ${0.000}$ & ${0.000}$ & $0.004$ & ${0.000}$ & ${0.001}$ & ${0.013}$ & ${0.011}$ &  $0.957$ \\
\rowcolor{rowgray}
 RTSGAN  & ${0.004}$ & $0.009$ & $0.009$ & $0.132$ & $0.002$ & $0.001$ & ${0.003}$ & ${0.012}$ & $0.019$ & $3.747$ \\
\rowcolor{white}
 HyperNSDE & ${0.006}$ & ${0.004}$ & ${0.004}$ & ${0.210}$ & $0.001$ & ${0.000}$ & $0.006$ & $0.019$ & ${0.009}$ &  $3.490$  \\

\bottomrule
    \end{tabular}}
    \end{table*}
    
\begin{table*}[h]
    \centering
    \scriptsize 
    \caption{\textbf{Ablation studies on simulated data --- standard deviations.} 
    Standard deviations computed across $N_{\mathrm{MC}}=30$ Monte Carlo runs and $N_{\text{gen}}=50$ generated datasets from a single trained model. Corresponding means are reported in Table~\ref{tab:ablation_studies}.}
    \label{tab:ablation_studies_std}
    \definecolor{rowgray}{gray}{0.9}
    \setlength{\tabcolsep}{4pt}
    \renewcommand{\arraystretch}{1.3}
    \setlength{\heavyrulewidth}{1pt}
    \setlength{\lightrulewidth}{1pt}
    \makebox[\textwidth][c]{
    \begin{tabular}{lcccccccccc}
\toprule
  & \multicolumn{6}{c}{\textbf{Fidelity}} & \multicolumn{3}{c}{\textbf{Privacy}} & \textbf{Utility}  \\
\cmidrule(lr){2-7} \cmidrule(lr){8-10} \cmidrule(lr){11-11}
\noalign{\vspace{-2pt}}
   & \multicolumn{2}{c}{\textbf{Corr. RMSE}}
    & \multirow{2}{*}{\textbf{Discrim.}}
    & \multirow{2}{*}{\textbf{KL $\lambda$}}
    & \multicolumn{2}{c}{\textbf{MMD}}
    & \multirow{2}{*}{\textbf{Dep.}}
    & \multirow{2}{*}{\textbf{MIR}}
    & \multirow{2}{*}{\textbf{NNAA}}
    & \multirow{2}{*}{\textbf{Pred.}} \\[-3pt]
\cmidrule(lr){2-3} \cmidrule(lr){6-7}
\noalign{\vspace{-2.8pt}}
  & \textbf{Long.} & \textbf{Static} & & & \textbf{Long.} & \textbf{Static} & & & & \\[-2pt]
\midrule
\rowcolor{white}
  Reference        & $0.047$ & $0.037$ & $0.069$ & $0.031$ & $0.321$   & $0.004$ & $0.000$ & $0.027$ & $0.044$ & $1.785$ \\
\cmidrule(lr){1-11}
\rowcolor{rowgray}
  AS.1 only ODE   & $0.077$ & $0.036$ & $0.032$ & $0.347$ & $241.458$ & $0.003$ & $0.053$ & $0.035$ & $0.037$ & $1.852$ \\
\rowcolor{white}
  AS.2 training   & $0.156$ & $0.039$ & $0.053$ & $0.361$ & $3.329$   & $0.004$ & $0.056$ & $0.026$ & $0.036$ & $5.339$ \\
\rowcolor{rowgray}
  AS.3 MNSDE      & $0.071$ & $0.024$ & $0.094$ & $0.091$ & $3.564$   & $0.004$ & $0.014$ & $0.034$ & $0.037$ & $1.641$ \\
\rowcolor{white}
  AS.3 SNSDE      & $0.069$ & $0.024$ & $0.189$ & $0.137$ & $3.647$   & $0.003$ & $0.009$ & $0.035$ & $0.039$ & $2.197$ \\
\rowcolor{rowgray}
  AS.3 FNSDE    & $0.103$ & $0.058$ & $0.112$ & $0.134$ & $24.650$ & $0.008$ & $0.007$ & $0.026$ & $0.034$ & $2.953$ \\  
\rowcolor{white}
  AS.4 w/o lbda   & $0.044$ & $0.037$ & $0.024$ & $0.413$ & $864.737$ & $0.002$ & $0.011$ & $0.028$ & $0.033$ & $5.513$ \\
\rowcolor{rowgray}
  HyperNSDE & $0.070$ & $0.043$ & $0.134$ & $0.213$ & $3.382$  & $0.003$ & $0.011$ & $0.029$ & $0.034$ & $2.114$ \\
\bottomrule
    \end{tabular}}
    \end{table*}

\begin{table}[h]
\centering
\caption{\textbf{Paired Wilcoxon signed-rank tests, HyperNSDE vs each baseline.} $N_{\mathrm{MC}}=30$ seeds (the 50 generations averaged within each seed before pairing), two-sided, Holm-corrected across all 30 tests. Cells show the median difference (HyperNSDE vs. baseline); the sign is oriented per metric so that \cmark{} marks where HyperNSDE is significantly better and \xmark{} where significantly worse. $^{*}\,p<.05$, $^{**}\,p<.01$, $^{***}\,p<.001$; n.s.\ = not significant after correction.}
\scriptsize
\setlength{\tabcolsep}{6pt}
\renewcommand{\arraystretch}{1.3}
\begin{tabular}{lccc}
\toprule
\textbf{Metric} & \textbf{vs MultiNODEs} & \textbf{vs RTSGAN} & \textbf{vs DGBFGP+HIVAE} \\
\midrule
MMD longi $\downarrow$        & $-2378.8^{***}$ \cmark & $-1.242^{***}$ \cmark & $-103.6^{***}$ \cmark \\
MMD static $\downarrow$       & $-0.0026^{**}$ \cmark  & $-0.0048^{***}$ \cmark & $-0.0026^{***}$ \cmark \\
Corr. RMSE longi $\downarrow$ & $+0.020$ n.s.          & $+0.007$ n.s.         & $-0.064$ n.s. \\
Corr. RMSE static $\downarrow$& $-0.004$ n.s.          & $-0.027^{*}$ \cmark   & $-0.014$ n.s. \\
Discrim. Score $\downarrow$   & $-0.035^{***}$ \cmark  & $-0.002$ n.s.         & $-0.037^{***}$ \cmark \\
KL $\lambda$ $\downarrow$    & $-66.86^{***}$ \cmark  & $0.138^{*}$ \xmark & $-66.86^{***}$ \cmark \\
Dep. Score $\downarrow$       & $-0.107^{***}$ \cmark  & $+0.001$ n.s.         & $-0.041^{***}$ \cmark \\
MIR $\downarrow$              & $+0.0007$ n.s.         & $+0.001$ n.s.         & $+0.002$ n.s. \\
NNAA $\rightarrow 0$          & $+0.009$ n.s.          & $-0.004$ n.s.         & $+0.013^{*}$ \xmark \\
Pred. Score $\downarrow$      & $-5.229^{***}$ \cmark  & $-0.098$ n.s.         & $-5.656^{***}$ \cmark \\
\bottomrule
\end{tabular}
\end{table}

\begin{table}[h]
    \centering
    \scriptsize
    \caption{\textbf{Evaluation metrics on simulated and real-world data --- medians.} $\downarrow$ lower is 
    better; $\rightarrow 0$ closer to zero is better (NNAA). \textbf{Bold} is the best 
    result excluding the reference, \underline{underlined} is second best.}
    \label{tab:all_results_medians}
    \definecolor{rowgray}{gray}{0.9}
    \setlength{\tabcolsep}{4pt}
    \renewcommand{\arraystretch}{1.3}
    \setlength{\heavyrulewidth}{1pt}
    \setlength{\lightrulewidth}{1pt}
    \makebox[\textwidth][c]{%
    %
    % ---- LEFT: dataset label column ----
    \begin{tabular}[t]{r}
    \phantommid
        \\
        \phantomcmid
        \noalign{\vspace{-2pt}}
        \\[-3pt]
        \phantomcmid
        \noalign{\vspace{-2.8pt}}
        \\[-2pt]
        \phantommid
        \multirow{5}{*}{\rotatebox{90}{\textbf{Simulated}}} \\[-2pt]
        \phantomcmid
        \\
        \\
        \\
        \\
        \arrayrulecolor{white}\midrule\arrayrulecolor{black}
        \multirow{4}{*}{\rotatebox{90}{\textbf{VELOUR}}} \\[-2pt]
        \phantomcmid
        \\
        \\
        \\
        \arrayrulecolor{white}\midrule\arrayrulecolor{black}
        \multirow{4}{*}{\rotatebox{90}{\textbf{PPMI}}} \\[-2pt]
        \phantomcmid
        \\
        \\
        \\
        \phantommid
    \end{tabular}%
    %
    % ---- RIGHT: main table ----
    \begin{tabular}[t]{lcccccccccc}
\toprule
  & \multicolumn{6}{c}{\textbf{Fidelity}} & \multicolumn{3}{c}{\textbf{Privacy}} & \textbf{Utility} \\
\cmidrule(lr){2-7} \cmidrule(lr){8-10} \cmidrule(lr){11-11}
\noalign{\vspace{-2pt}}
   & \multicolumn{2}{c}{\textbf{Corr. RMSE} $\downarrow$}
    & \multirow{2}{*}{\textbf{Discrim.} $\downarrow$}
    & \multirow{2}{*}{\textbf{KL $\lambda$} $\downarrow$}
    & \multicolumn{2}{c}{\textbf{MMD} $\downarrow$}
    & \multirow{2}{*}{\textbf{Dep.} $\downarrow$}
    & \multirow{2}{*}{\textbf{MIR} $\downarrow$}
    & \multirow{2}{*}{\textbf{NNAA} $\rightarrow 0$}
    & \multirow{2}{*}{\textbf{Pred.} $\downarrow$} \\[-3pt]
\cmidrule(lr){2-3} \cmidrule(lr){6-7}
\noalign{\vspace{-2.8pt}}
  & \textbf{Long.} & \textbf{Static} & & & \textbf{Long.} & \textbf{Static} & & & & \\[-2pt]
%%% Simulated
\midrule
\rowcolor{white}
 Reference  & $0.062$ & $0.062$ & $0.493$ & $0.133$ & $0.000$ & $0.006$ & $0.001$ & $0.021$ & $0.047$ & $3.449$
 \\ [-2pt]
\cmidrule(lr){1-11}
\rowcolor{rowgray}
 MultiNODEs & $\mathbf{0.053}$ & $\underline{0.035}$ & $1.000$ & $67.522$ & $35.926$ & $0.006$ & $0.118$ & $\underline{0.028}$ & $\underline{-0.014}$ & $9.527$ \\
\rowcolor{white}
 RTSGAN & $\underline{0.079}$ & $0.060$ & $\mathbf{0.978}$ & $\mathbf{0.460}$ & $\underline{1.175}$ & $0.008$ & $\mathbf{0.011}$ & $0.030$ & $-0.027$ & $\underline{4.957}$ \\
\rowcolor{rowgray}
 DGBFGP + HIVAE & $0.162$ & $0.042$ & $1.000$ & $67.522$ & $12.198$ & $\underline{0.005}$ & $0.058$ & $\mathbf{0.028}$ & $\mathbf{-0.009}$ & $9.730$ \\
\rowcolor{white}
 HyperNSDE & $0.098$ & $\mathbf{0.031}$ & $\underline{0.983}$ & $\underline{0.668}$ & $\mathbf{0.145}$ & $\mathbf{0.003}$ & $\underline{0.013}$ & $0.032$ & $-0.028$ & $\mathbf{4.777}$ \\
 
%%% VELOUR
\midrule
\rowcolor{white}
 Reference & $0.091$ & $0.132$ & $0.300$ & $0.845$ & $0.001$ & $0.019$ & $0.063$ & $0.073$ & $-0.049$ & $21.118$  \\[-2pt]
\cmidrule(lr){1-11}
\rowcolor{rowgray}
 MultiNODEs & $\mathbf{0.139}$ & $\underline{0.154}$ & ${1.000}$ & $139.716$ & $\underline{0.140}$ & $\mathbf{0.019}$ & $0.628$ & $\underline{0.052}$ & $\mathbf{-0.003}$ & $\mathbf{5.846}$ \\
\rowcolor{white}
 RTSGAN & $0.842$ & $\mathbf{0.149}$ & ${1.000}$ & $\mathbf{2.533}$ & $24.844$ & $0.019$ & $\underline{0.116}$ & $\mathbf{0.048}$ & $0.083$ & $5821.113$ \\
\rowcolor{rowgray}
 HyperNSDE  & $\underline{0.239}$ & $0.165$ & ${1.000}$ & $\underline{3.227}$ & $\mathbf{0.029}$ & $\underline{0.019}$ & $\mathbf{0.106}$ & $0.061$ & $\underline{0.013}$ & $\underline{53.447}$ \\
 
%%% PPMI
\midrule
\rowcolor{white}
 Reference       & $0.044$ & $0.092$ & $0.587$ & $0.064$ & $0.000$ & $0.006$ & $0.052$ & $0.065$ & $-0.032$ & $39.137$ \\
 [-2pt]
\cmidrule(lr){1-11}
\rowcolor{rowgray}
 MultiNODEs & $\mathbf{0.107}$ & $\underline{0.133}$ & $1.000$ & $221.398$ & $0.037$ & $0.013$ & $0.391$ & $\mathbf{0.013}$ & $0.007$ & $\mathbf{19.841}$ \\
\rowcolor{white}
 RTSGAN & $0.623$ & $\mathbf{0.113}$ & $\mathbf{0.983}$ & $\mathbf{1.013}$ & $\mathbf{0.009}$ & $\mathbf{0.012}$ & $\mathbf{0.057}$ & $0.057$ & $\underline{-0.006}$ & $\underline{52.895}$ \\
\rowcolor{rowgray}
 HyperNSDE   & $\underline{0.260}$ & $0.136$ & $\underline{0.997}$ & $\underline{1.796}$ & $\underline{0.019}$ & $\underline{0.013}$ & $\underline{0.083}$ & $\underline{0.028}$ & $\mathbf{0.002}$ & $58.470$ \\
\bottomrule
    \end{tabular}}
\end{table}

% \clearpage
\FloatBarrier

\subsection{Additional analyses of evaluation metrics}
\label{app:metric_diagnostics}

We complement the controlled analyses reported in the main text (Table~\ref{tab:effect_grid_res}) with additional diagnostics of the discriminative score. 
% In particular, we decompose the discriminative score by input component.

\paragraph{Discriminative score decomposition}
To identify which components of the synthetic records drive the discriminative score, we repeat the real-versus-synthetic classification on simulated data using static covariates, observation patterns, and trajectory values separately. Static covariates are difficult to distinguish from real data across methods, whereas observation patterns strongly distinguish the baselines. For HyperNSDE, the remaining discriminability mainly arises from trajectory values.

\begin{table}[h]
\centering
\caption{\textbf{C2ST test AUROC by classifier input (simulated
data).} Values closer to 0.5 are better (harder to distinguish from real). Best non-reference value per column in \textbf{bold}.}
\scriptsize
\renewcommand{\arraystretch}{1.3}
\setlength{\tabcolsep}{6pt}
\begin{tabular}{lcccc}
\toprule
\textbf{Model} & \textbf{Static only} & \textbf{Pattern only} & \textbf{Values only} & \textbf{All (Table 2)} \\
\midrule
Reference   & $0.493$ & $0.511$ & $0.480$ & $0.499$ \\
\cmidrule(lr){1-5}
MultiNODEs  & $0.486$ & $1.000$ & $0.998$ & $0.998$ \\
RTSGAN      & $\mathbf{0.522}$ & $1.000$ & $0.947$ & $0.932$ \\
DGBFGP + HIVAE& $0.487$ & $1.000$ & $0.999$ & $0.999$ \\
HyperNSDE  & $0.462$ & $\mathbf{0.760}$ & $\mathbf{0.942}$ & $\mathbf{0.926}$ \\
\bottomrule
\end{tabular}
\end{table}

\FloatBarrier

\subsection{Sensitivity analysis}
\label{app:sensitivity_analysis_results}

We further assess robustness under controlled variations described in Appendix~\ref{app:sensitivy_analysis}: limited training data, sparse observations, state-dependent sampling, and prior versus posterior latent sampling.
% \perrine{ajouter lambda ici}

\begin{table*}[h]
    \centering
    \scriptsize 
    \caption{\textbf{Sensitivity analysis: \emph{(i) Limited training data}.} Performance comparison on simulated data when restricting the number of available training samples. $\downarrow$ lower is better; $\rightarrow 0$ closer to zero is better (NNAA). \textbf{Bold} is the best result excluding the reference, \underline{underlined} is second best.}
    \label{tab:simu_SE_limited_data}
    \definecolor{rowgray}{gray}{0.9}
    \setlength{\tabcolsep}{4pt}
    \renewcommand{\arraystretch}{1.3}
    \setlength{\heavyrulewidth}{1pt}
    \setlength{\lightrulewidth}{1pt}
    \makebox[\textwidth][c]{
    \begin{tabular}{lcccccccccc}
\toprule
  & \multicolumn{6}{c}{\textbf{Fidelity}} & \multicolumn{3}{c}{\textbf{Privacy}} & \textbf{Utility}  \\
\cmidrule(lr){2-7} \cmidrule(lr){8-10} \cmidrule(lr){11-11} 
\noalign{\vspace{-2pt}}
   & \multicolumn{2}{c}{\textbf{Corr. RMSE} $\downarrow$} 
    & \multirow{2}{*}{\textbf{Discrim.} $\downarrow$} 
    & \multirow{2}{*}{\textbf{KL $\lambda$} $\downarrow$} 
    & \multicolumn{2}{c}{\textbf{MMD} $\downarrow$} 
    & \multirow{2}{*}{\textbf{Dep.} $\downarrow$} 
    & \multirow{2}{*}{\textbf{MIR} $\downarrow$}  
    & \multirow{2}{*}{\textbf{NNAA} $\rightarrow 0$} 
    & \multirow{2}{*}{\textbf{Pred.} $\downarrow$} \\[-3pt]
\cmidrule(lr){2-3} \cmidrule(lr){6-7}
\noalign{\vspace{-2.8pt}}
  & \textbf{Long.} & \textbf{Static} & & & \textbf{Long.} & \textbf{Static} & & & & \\[-2pt]
\midrule
\rowcolor{white}
 Reference & $0.061$ & $0.060$ & $0.499$ & $0.139$ & $0.021$ & $0.007$ & $0.001$ & $0.033$ & $0.021$ & $3.759$ \\
\cmidrule(lr){1-11}
\rowcolor{rowgray}
 MultiNODEs & $0.190$ & $0.085$ & $0.998$ & $67.616$ & $897.653$ & $0.010$ & $0.128$ & $\underline{0.044}$ & $\mathbf{-0.000}$ & $11.394$ \\
\rowcolor{white}
 RTSGAN & $0.115$ & $0.070$ & $0.973$ & $1.754$ & $282.023$ & $0.013$ & $0.016$ & $\mathbf{0.044}$ & $0.004$ & $5.680$ \\
\rowcolor{rowgray}
 DGBFGP + HIVAE & $0.138$ & $\underline{0.053}$ & $0.999$ & $67.616$ & $238.883$ & $\underline{0.007}$ & $0.058$ & $0.045$ & $\underline{-0.000}$ & $11.663$ \\
% \cmidrule(lr){1-11}
\rowcolor{white}
 HyperNSDE (post.) & $\mathbf{0.086}$ & $\mathbf{0.035}$ & $\mathbf{0.765}$ & $\mathbf{0.749}$ & $\mathbf{2.427}$ & $\mathbf{0.003}$ & $\mathbf{0.009}$ & $0.048$ & $-0.014$ & $\mathbf{4.764}$ \\
\rowcolor{rowgray}
 HyperNSDE (prior) & $\underline{0.105}$ & $0.086$ & $\underline{0.790}$ & $\underline{0.762}$ & $\underline{2.570}$ & $0.019$ & $\underline{0.010}$ & $0.045$ & $-0.001$ & $\underline{5.019}$ \\
\bottomrule
    \end{tabular}}
    \end{table*}

 \begin{table*}[h]
    \centering
    \scriptsize 
    \caption{\textbf{Sensitivity analysis: \emph{(ii) Sparse observation regime}.} Performance comparison on simulated data under reduced observation frequencies. $\downarrow$ lower is better; $\rightarrow 0$ closer to zero is better (NNAA). \textbf{Bold} is the best result excluding the reference, \underline{underlined} is second best.}
    \label{tab:simu_SE_sparse}
    \definecolor{rowgray}{gray}{0.9}
    \setlength{\tabcolsep}{4pt}
    \renewcommand{\arraystretch}{1.3}
    \setlength{\heavyrulewidth}{1pt}
    \setlength{\lightrulewidth}{1pt}
    \makebox[\textwidth][c]{
    \begin{tabular}{lcccccccccc}
\toprule
  & \multicolumn{6}{c}{\textbf{Fidelity}} & \multicolumn{3}{c}{\textbf{Privacy}} & \textbf{Utility}  \\
\cmidrule(lr){2-7} \cmidrule(lr){8-10} \cmidrule(lr){11-11} 
\noalign{\vspace{-2pt}}
   & \multicolumn{2}{c}{\textbf{Corr. RMSE} $\downarrow$} 
    & \multirow{2}{*}{\textbf{Discrim.} $\downarrow$} 
    & \multirow{2}{*}{\textbf{KL $\lambda$} $\downarrow$} 
    & \multicolumn{2}{c}{\textbf{MMD} $\downarrow$} 
    & \multirow{2}{*}{\textbf{Dep.} $\downarrow$} 
    & \multirow{2}{*}{\textbf{MIR} $\downarrow$}  
    & \multirow{2}{*}{\textbf{NNAA} $\rightarrow 0$} 
    & \multirow{2}{*}{\textbf{Pred.} $\downarrow$} \\[-3pt]
\cmidrule(lr){2-3} \cmidrule(lr){6-7}
\noalign{\vspace{-2.8pt}}
  & \textbf{Long.} & \textbf{Static} & & & \textbf{Long.} & \textbf{Static} & & & & \\[-2pt]
\midrule
\rowcolor{white}
 Reference & $0.065$ & $0.060$ & $0.504$ & $0.209$ & $-0.001$ & $0.007$ & $0.006$ & $0.032$ & $0.041$ & $12.147$ \\
\cmidrule(lr){1-11}
\rowcolor{rowgray}
 MultiNODEs & $0.141$ & $0.055$ & $0.997$ & $104.096$ & $876.022$ & $\mathbf{0.006}$ & $0.182$ & $0.034$ & $\underline{-0.008}$ & $32.967$ \\
\rowcolor{white}
 RTSGAN & $\mathbf{0.077}$ & $0.064$ & $\mathbf{0.851}$ & $1.890$ & $0.051$ & $0.009$ & $0.053$ & $\mathbf{0.033}$ & $-0.045$ & $\mathbf{13.975}$ \\
\rowcolor{rowgray}
 DGBFGP + HIVAE & $0.307$ & $\underline{0.052}$ & $0.998$ & $104.096$ & $1187.980$ & $\underline{0.006}$ & $0.115$ & $\underline{0.033}$ & $\mathbf{-0.007}$ & $32.964$ \\
% \cmidrule(lr){1-11}
\rowcolor{white}
 HyperNSDE (post.) & $\underline{0.093}$ & $\mathbf{0.046}$ & $\underline{0.926}$ & $\underline{1.607}$ & $\mathbf{0.013}$ & $0.009$ & $\mathbf{0.033}$ & $0.036$ & $-0.054$ & $15.265$ \\
\rowcolor{rowgray}
 HyperNSDE (prior) & $0.114$ & $0.072$ & $0.929$ & $\mathbf{1.584}$ & $\underline{0.016}$ & $0.021$ & $\underline{0.037}$ & $0.034$ & $-0.044$ & $\underline{15.134}$ \\
\bottomrule
    \end{tabular}}
    \end{table*}

\begin{table*}[h]
    \centering
    \scriptsize 
    \caption{\textbf{Sensitivity analysis: \emph{(iii) State-dependent observation process}.} Evaluation on simulated data where observation times are influenced by the underlying system state. $\downarrow$ lower is better; $\rightarrow 0$ closer to zero is better (NNAA). \textbf{Bold} is the best result excluding the reference, \underline{underlined} is second best.}
    \label{tab:simu_SE_depobs_process}
    \definecolor{rowgray}{gray}{0.9}
    \setlength{\tabcolsep}{4pt}
    \renewcommand{\arraystretch}{1.3}
    \setlength{\heavyrulewidth}{1pt}
    \setlength{\lightrulewidth}{1pt}
    \makebox[\textwidth][c]{
    \begin{tabular}{lcccccccccc}
\toprule
  & \multicolumn{6}{c}{\textbf{Fidelity}} & \multicolumn{3}{c}{\textbf{Privacy}} & \textbf{Utility}  \\
\cmidrule(lr){2-7} \cmidrule(lr){8-10} \cmidrule(lr){11-11} 
\noalign{\vspace{-2pt}}
   & \multicolumn{2}{c}{\textbf{Corr. RMSE} $\downarrow$} 
    & \multirow{2}{*}{\textbf{Discrim.} $\downarrow$} 
    & \multirow{2}{*}{\textbf{KL $\lambda$} $\downarrow$} 
    & \multicolumn{2}{c}{\textbf{MMD} $\downarrow$} 
    & \multirow{2}{*}{\textbf{Dep.} $\downarrow$} 
    & \multirow{2}{*}{\textbf{MIR} $\downarrow$}  
    & \multirow{2}{*}{\textbf{NNAA} $\rightarrow 0$} 
    & \multirow{2}{*}{\textbf{Pred.} $\downarrow$} \\[-3pt]
\cmidrule(lr){2-3} \cmidrule(lr){6-7}
\noalign{\vspace{-2.8pt}}
  & \textbf{Long.} & \textbf{Static} & & & \textbf{Long.} & \textbf{Static} & & & & \\[-2pt]
\midrule
\rowcolor{white}
 Reference & $0.067$ & $0.060$ & $0.494$ & $0.187$ & $0.079$ & $0.007$ & $0.001$ & $0.037$ & $0.030$ & $3.714$ \\
\cmidrule(lr){1-11}
\rowcolor{rowgray}
 MultiNODEs & $\mathbf{0.094}$ & $\underline{0.044}$ & $0.993$ & $67.708$ & $1122.343$ & $0.006$ & $0.147$ & $0.037$ & $\underline{-0.008}$ & $9.828$ \\
\rowcolor{white}
 RTSGAN & $0.099$ & $0.058$ & $\underline{0.899}$ & $\underline{0.362}$ & $\underline{3.903}$ & $0.009$ & $\underline{0.020}$ & $0.038$ & $-0.014$ & $\mathbf{5.148}$ \\
\rowcolor{rowgray}
 DGBFGP + HIVAE & $0.158$ & $0.052$ & $0.998$ & $67.708$ & $204.745$ & $0.006$ & $0.077$ & $\mathbf{0.036}$ & $\mathbf{-0.005}$ & $10.001$ \\
% \cmidrule(lr){1-11}
\rowcolor{white}
 AS.4 w/o lbda & $\underline{0.096}$ & $0.049$ & $0.946$ & $66.857$ & $287.777$ & $\underline{0.004}$ & $0.035$ & $\underline{0.037}$ & $-0.016$ & $9.887$ \\
% \cmidrule(lr){1-11}
\rowcolor{rowgray}
 HyperNSDE (post.) & $0.157$ & $\mathbf{0.036}$ & $\mathbf{0.896}$ & $\mathbf{0.722}$ & $\mathbf{0.275}$ & $\mathbf{0.004}$ & $\mathbf{0.017}$ & $0.048$ & $-0.028$ & $\underline{5.726}$ \\
\bottomrule
    \end{tabular}}
    \end{table*}

\begin{table*}[h]
    \centering
    \scriptsize
    \caption{\textbf{Sensitivity analysis: \emph{(iv) Prior versus posterior sampling}.} Comparison of HyperNSDE under posterior and prior sampling of the static latent representation on simulated and real-world datasets. $\downarrow$ lower is better; $\rightarrow 0$ closer to zero is better (NNAA). \textbf{Bold} is the best result excluding the reference.}
    \label{tab:prior_posterior_sensitivity}
    \definecolor{rowgray}{gray}{0.9}
    \setlength{\tabcolsep}{4pt}
    \renewcommand{\arraystretch}{1.3}
    \setlength{\heavyrulewidth}{1pt}
    \setlength{\lightrulewidth}{1pt}
    \makebox[\textwidth][c]{%
    %
    % ---- LEFT: dataset label column ----
    \begin{tabular}[t]{r}
    \phantommid
        \\
        \phantomcmid
        \noalign{\vspace{-2pt}}
        \\[-3pt]
        \phantomcmid
        \noalign{\vspace{-2.8pt}}
        \\[-2pt]
        \phantommid
        \multirow{3}{*}{\rotatebox{90}{\textbf{Simulated}}} \\[-2pt]
        \phantomcmid
        \\
        \\
        \arrayrulecolor{white}\midrule\arrayrulecolor{black}
        \multirow{3}{*}{\rotatebox{90}{\textbf{VELOUR}}} \\[-2pt]
        \phantomcmid
        \\
        \\
        \arrayrulecolor{white}\midrule\arrayrulecolor{black}
        \multirow{3}{*}{\rotatebox{90}{\textbf{PPMI}}} \\[-2pt]
        \phantomcmid
        \\
        \\
        % \arrayrulecolor{white}\midrule\arrayrulecolor{black}
        % \multirow{4}{*}{\rotatebox{90}{\textbf{PhysioNet 2012}}} \\[-2pt]
        % \phantomcmid
        % \\
        % \\
        % \\
        % \arrayrulecolor{white}\midrule\arrayrulecolor{black}
        % \multirow{4}{*}{\rotatebox{90}{\textbf{MSK-CHORD}}} \\[-2pt]
        % \phantomcmid
        % \\
        % \\
        % \\
        \phantommid
    \end{tabular}%
    %
    % ---- RIGHT: main table ----
    \begin{tabular}[t]{lcccccccccc}
\toprule
  & \multicolumn{6}{c}{\textbf{Fidelity}} & \multicolumn{3}{c}{\textbf{Privacy}} & \textbf{Utility} \\
\cmidrule(lr){2-7} \cmidrule(lr){8-10} \cmidrule(lr){11-11}
\noalign{\vspace{-2pt}}
   & \multicolumn{2}{c}{\textbf{Corr. RMSE} $\downarrow$}
    & \multirow{2}{*}{\textbf{Discrim.} $\downarrow$}
    & \multirow{2}{*}{\textbf{KL $\lambda$} $\downarrow$}
    & \multicolumn{2}{c}{\textbf{MMD} $\downarrow$}
    & \multirow{2}{*}{\textbf{Dep.} $\downarrow$}
    & \multirow{2}{*}{\textbf{MIR} $\downarrow$}
    & \multirow{2}{*}{\textbf{NNAA} $\rightarrow 0$}
    & \multirow{2}{*}{\textbf{Pred.} $\downarrow$} \\[-3pt]
\cmidrule(lr){2-3} \cmidrule(lr){6-7}
\noalign{\vspace{-2.8pt}}
  & \textbf{Long.} & \textbf{Static} & & & \textbf{Long.} & \textbf{Static} & & & & \\[-2pt]
%%% Simulated
\midrule
\rowcolor{white}
 Reference        & $0.061$ & $0.060$ & $0.499$ & $0.031$ & $0.021$   & $0.007$ & $0.001$ & $0.033$ & $0.021$ & $3.759$ \\ [-2pt]
\cmidrule(lr){1-11}
\rowcolor{rowgray}
 HyperNSDE (post.) & $\mathbf{0.118}$ & $\mathbf{0.044}$ & $\mathbf{0.926}$ & $\mathbf{0.687}$ & ${1.150}$ & $\mathbf{0.004}$ & $\mathbf{0.014}$ & $0.039$ & $-0.021$ & $\mathbf{5.226}$ \\
\rowcolor{white}
 HyperNSDE (prior) & $0.123$ & $0.090$ & $0.932$ & ${0.705}$ & $\mathbf{1.119}$ & $0.018$ & ${0.015}$ & $\mathbf{0.035}$ & $\mathbf{-0.007}$ & $5.612$ \\
%%% VELOUR
\midrule
\rowcolor{white}
 Reference  & $0.091$ & $0.132$ & $0.472$ & $0.845$ & $0.001$ & $0.019$ & $0.063$ & $0.073$ & $-0.049$ & $21.958$ \\[-2pt]
\cmidrule(lr){1-11}
\rowcolor{rowgray}
 HyperNSDE (post.) & ${0.239}$ & $\mathbf{0.164}$ & $\mathbf{0.998}$ & $\mathbf{3.513}$ & ${0.029}$ & ${0.019}$ & $\mathbf{0.105}$ & $\mathbf{0.063}$ & ${0.010}$ & $\mathbf{53.514}$ \\
 \rowcolor{white}
HyperNSDE (prior)   &  $0.239$ &  $0.165$ &  $1.000$ & {$3.708$}  &  $\mathbf{0.027}$ &  $0.019$ & ${0.107}$ &  $0.065$ &   $\mathbf{0.007}$ & $53.745$ \\
%%% PPMI
\midrule
\rowcolor{white}
 Reference          & $0.044$ & $0.092$ & $0.587$ & $0.064$ & $0.001$ & $0.006$ & $0.052$ & $0.065$ & $-0.032$ & $39.137$ \\[-2pt]
\cmidrule(lr){1-11}
\rowcolor{rowgray}
 HyperNSDE (post.)   & ${0.260}$ & $0.135$ & $\mathbf{0.996}$ & ${1.840}$ & $\mathbf{0.019}$  & ${0.013}$ & ${0.084}$ & ${0.032}$ & $\mathbf{0.002}$ & $\mathbf{57.697}$  \\
\rowcolor{white}
  HyperNSDE (prior)   & $\mathbf{0.259}$  & $0.135$ & $0.997$  &  $\mathbf{1.834}$ & $0.020$  & $0.013$  & $\mathbf{0.082}$ &  $\mathbf{0.017}$ & $0.011$  &  $65.961$ \\
\bottomrule
    \end{tabular}}
    \end{table*}

\begin{table*}[h]
    \centering
    \scriptsize
     \caption{\textbf{Sensitivity analysis: \emph{(v) State dependence of the observation intensity}.} Comparison of HyperNSDE with a latent-state-dependent intensity and a state-independent intensity. $\downarrow$ lower is better; $\rightarrow 0$ closer to zero is better (NNAA). \textbf{Bold} is the best result excluding the reference.}
    \label{tab:indep_lambda_sensitivity}
    \definecolor{rowgray}{gray}{0.9}
    \setlength{\tabcolsep}{4pt}
    \renewcommand{\arraystretch}{1.3}
    \setlength{\heavyrulewidth}{1pt}
    \setlength{\lightrulewidth}{1pt}
    \makebox[\textwidth][c]{%
    %
    % ---- LEFT: dataset label column ----
    \begin{tabular}[t]{r}
    \phantommid
        \\
        \phantomcmid
        \noalign{\vspace{-2pt}}
        \\[-3pt]
        \phantomcmid
        \noalign{\vspace{-2.8pt}}
        \\[-2pt]
        \phantommid
        \multirow{3}{*}{\rotatebox{90}{\textbf{Simulated}}} \\[-2pt]
        \phantomcmid
        \\
        \\
        \arrayrulecolor{white}\midrule\arrayrulecolor{black}
        \multirow{3}{*}{\rotatebox{90}{\textbf{VELOUR}}} \\[-2pt]
        \phantomcmid
        \\
        \\
        \arrayrulecolor{white}\midrule\arrayrulecolor{black}
        \multirow{3}{*}{\rotatebox{90}{\textbf{PPMI}}} \\[-2pt]
        \phantomcmid
        \\
        \\
        % \arrayrulecolor{white}\midrule\arrayrulecolor{black}
        % \multirow{4}{*}{\rotatebox{90}{\textbf{PhysioNet 2012}}} \\[-2pt]
        % \phantomcmid
        % \\
        % \\
        % \\
        % \arrayrulecolor{white}\midrule\arrayrulecolor{black}
        % \multirow{4}{*}{\rotatebox{90}{\textbf{MSK-CHORD}}} \\[-2pt]
        % \phantomcmid
        % \\
        % \\
        % \\
        \phantommid
    \end{tabular}%
    %
    % ---- RIGHT: main table ----
    \begin{tabular}[t]{lcccccccccc}
\toprule
  & \multicolumn{6}{c}{\textbf{Fidelity}} & \multicolumn{3}{c}{\textbf{Privacy}} & \textbf{Utility} \\
\cmidrule(lr){2-7} \cmidrule(lr){8-10} \cmidrule(lr){11-11}
\noalign{\vspace{-2pt}}
   & \multicolumn{2}{c}{\textbf{Corr. RMSE} $\downarrow$}
    & \multirow{2}{*}{\textbf{Discrim.} $\downarrow$}
    & \multirow{2}{*}{\textbf{KL $\lambda$} $\downarrow$}
    & \multicolumn{2}{c}{\textbf{MMD} $\downarrow$}
    & \multirow{2}{*}{\textbf{Dep.} $\downarrow$}
    & \multirow{2}{*}{\textbf{MIR} $\downarrow$}
    & \multirow{2}{*}{\textbf{NNAA} $\rightarrow 0$}
    & \multirow{2}{*}{\textbf{Pred.} $\downarrow$} \\[-3pt]
\cmidrule(lr){2-3} \cmidrule(lr){6-7}
\noalign{\vspace{-2.8pt}}
  & \textbf{Long.} & \textbf{Static} & & & \textbf{Long.} & \textbf{Static} & & & & \\[-2pt]
%%% Simulated
\midrule
\rowcolor{white}
 Reference        & $0.067$ & $0.060$ & $0.494$ & $0.187$ & $0.079$ & $0.007$ & $0.001$ & $0.037$ & $0.030$ & $3.714$ \\ [-2pt]
\cmidrule(lr){1-11}
\rowcolor{rowgray}
 HyperNSDE (state-dep. $\lambda$) & $\mathbf{0.157}$ & $\mathbf{0.036}$ & $\mathbf{0.896}$ & ${0.722}$ & $\mathbf{0.275}$ & $\mathbf{0.004}$ & $\mathbf{0.017}$ & $0.048$ & $-0.028$ & ${5.726}$ \\
\rowcolor{white}
 HyperNSDE (state-indep. $\lambda$)  & $0.249$  & $0.090$ & $0.975$  &  $\mathbf{0.427}$ & $1.600$  & $0.009$  & $0.024$ &  $\mathbf{0.036}$ & $\mathbf{-0.003}$  &  $\mathbf{2.553}$  \\
 
%%% VELOUR
\midrule
\rowcolor{white}
 Reference  & $0.091$ & $0.132$ & $0.472$ & $0.845$ & $0.001$ & $0.019$ & $0.063$ & $0.073$ & $-0.049$ & $21.958$ \\[-2pt]
\cmidrule(lr){1-11}
\rowcolor{rowgray}
 HyperNSDE (state-dep. $\lambda$) & ${0.239}$ & $\mathbf{0.164}$ & $\mathbf{0.998}$ & $\mathbf{3.513}$ & ${0.029}$ & ${0.019}$ & $\mathbf{0.105}$ & ${0.063}$ & $\mathbf{0.010}$ & $\mathbf{53.514}$ \\
 \rowcolor{white}
HyperNSDE (state-indep. $\lambda$)  & $\mathbf{0.221}$  & $0.165$ & $1.000$  &  $7.190$ & $0.029$  & $0.019$  & $0.138$ &  $\mathbf{0.057}$ & $0.014$  &  $54.773$  \\

%%% PPMI
\midrule
\rowcolor{white}
 Reference          & $0.044$ & $0.092$ & $0.587$ & $0.064$ & $0.001$ & $0.006$ & $0.052$ & $0.065$ & $-0.032$ & $39.137$ \\[-2pt]
\cmidrule(lr){1-11}
\rowcolor{rowgray}
 HyperNSDE (state-dep. $\lambda$)  & $\mathbf{0.260}$ & $0.135$ & ${0.996}$ & ${1.840}$ & $\mathbf{0.019}$  & ${0.013}$ & ${0.084}$ & ${0.032}$ & $\mathbf{0.002}$ & $\mathbf{57.697}$  \\
\rowcolor{white}
  HyperNSDE (state-indep. $\lambda$)  & $0.264$  & $\mathbf{0.134}$ & $\mathbf{0.963}$  &  $\mathbf{1.339}$ & $0.020$  & $0.013$  & $\mathbf{0.079}$ &  $\mathbf{0.021}$ & $0.008$  & $66.257$ \\
\bottomrule
    \end{tabular}}
    \end{table*}

% \clearpage
\FloatBarrier

\subsection{Subgroup and cohort analyses}
\label{app:subgroup_cohort}

We complement the aggregate evaluation with analyses of subgroup-specific fidelity on simulated data and source-cohort representativeness on VELOUR. 

HyperNSDE learns the observational distribution represented in its training cohort. To assess how closely the generated population reproduces the source cohort, we compare real and synthetic static covariates on VELOUR using standardized mean differences (SMDs). Across the 26 static covariates, the median SMD is $0.092$, indicating close agreement with the observed VELOUR cohort. This assesses fidelity to the source population and should not be interpreted as a correction of selection bias or representativeness with respect to a broader target population.

\begin{table*}[h]
    \centering
    \scriptsize 
    \caption{\textbf{Results on the subset $W_1 = 0$ of simulated data.} Evaluation in the reference setting, restricted exclusively to the subpopulation of samples where $W_1 = 0$. $\downarrow$ lower is better; $\rightarrow 0$ closer to zero is better (NNAA). \textbf{Bold} is the best result excluding the reference, \underline{underlined} is second best.}
    \label{tab:simu_ref_setting_W1_0}
    \definecolor{rowgray}{gray}{0.9}
    \setlength{\tabcolsep}{4pt}
    \renewcommand{\arraystretch}{1.3}
    \setlength{\heavyrulewidth}{1pt}
    \setlength{\lightrulewidth}{1pt}
    \makebox[\textwidth][c]{
    \begin{tabular}{lcccccccccc}
\toprule
  & \multicolumn{6}{c}{\textbf{Fidelity}} & \multicolumn{3}{c}{\textbf{Privacy}} & \textbf{Utility}  \\
\cmidrule(lr){2-7} \cmidrule(lr){8-10} \cmidrule(lr){11-11} 
\noalign{\vspace{-2pt}}
   & \multicolumn{2}{c}{\textbf{Corr. RMSE} $\downarrow$} 
    & \multirow{2}{*}{\textbf{Discrim.} $\downarrow$} 
    & \multirow{2}{*}{\textbf{KL $\lambda$} $\downarrow$} 
    & \multicolumn{2}{c}{\textbf{MMD} $\downarrow$} 
    & \multirow{2}{*}{\textbf{Dep.} $\downarrow$} 
    & \multirow{2}{*}{\textbf{MIR} $\downarrow$}  
    & \multirow{2}{*}{\textbf{NNAA} $\rightarrow 0$} 
    & \multirow{2}{*}{\textbf{Pred.} $\downarrow$} \\[-3pt]
\cmidrule(lr){2-3} \cmidrule(lr){6-7}
\noalign{\vspace{-2.8pt}}
  & \textbf{Long.} & \textbf{Static} & & & \textbf{Long.} & \textbf{Static} & & & & \\[-2pt]
\midrule
\rowcolor{white}
 Reference & $0.091$ & $0.000$ & $0.482$ & $0.288$ & $-0.001$ & $0.010$ & $0.002$ & $0.057$ & $0.016$ & $8.912$ \\
\cmidrule(lr){1-11}
\rowcolor{rowgray}
 MultiNODEs & $\underline{0.180}$ & $0.000$ & $0.992$ & $67.746$ & $78.209$ & $0.011$ & $0.178$ & $\underline{0.050}$ & $\mathbf{-0.009}$ & $\mathbf{8.829}$ \\
\rowcolor{white}
 RTSGAN & $\mathbf{0.150}$ & $\underline{0.000}$ & $\underline{0.876}$ & $8.007$ & $0.456$ & $0.022$ & $0.032$ & $0.052$ & $\underline{-0.010}$ & $15.541$ \\
\rowcolor{rowgray}
 DGBFGP + HIVAE & $0.322$ & $0.000$ & $0.997$ & $67.746$ & $60.341$ & $0.009$ & $0.077$ & $0.050$ & $-0.010$ & $\underline{8.862}$ \\
% \cmidrule(lr){1-11}
\rowcolor{white}
 AS.3 MNSDE & $0.249$ & $0.000$ & $0.914$ & $\mathbf{0.605}$ & $0.157$ & $\mathbf{0.006}$ & $0.026$ & $0.054$ & $-0.024$ & $53.848$ \\
\rowcolor{rowgray}
 AS.3 SNSDE & $0.217$ & $\mathbf{0.000}$ & $\mathbf{0.796}$ & ${0.746}$ & $\mathbf{0.087}$ & $\underline{0.006}$ & $\mathbf{0.014}$ & $0.054$ & $-0.029$ & $166.064$ \\
% \cmidrule(lr){1-11}
\rowcolor{white}
 AS.3 FNSDE & $0.277$ & $0.000$ & $0.880$ & $\underline{0.722}$ & $1.645$ & $0.012$ & $0.024$ & $0.050$ & $-0.010$ & $129.664$ \\
\rowcolor{rowgray}
 HyperNSDE (post.) & $0.244$ & $0.000$ & $0.890$ & $0.918$ & $\underline{0.101}$ & $0.011$ & $\underline{0.017}$ & $0.052$ & $-0.020$ & $52.438$ \\
\rowcolor{white}
 HyperNSDE (prior) & $0.254$ & $0.000$ & $0.893$ & ${0.913}$ & ${0.121}$ & $0.020$ & $0.018$ & $\mathbf{0.049}$ & $-0.014$ & $54.613$ \\
\bottomrule
    \end{tabular}}
    \end{table*}

\begin{table*}[h]
    \centering
    \scriptsize 
    \caption{\textbf{Results on the subset $W_1 = 1$ of simulated data.} Evaluation in the reference setting, restricted exclusively to the subpopulation of samples where $W_1 = 1$. $\downarrow$ lower is better; $\rightarrow 0$ closer to zero is better (NNAA). \textbf{Bold} is the best result excluding the reference, \underline{underlined} is second best.}
    \label{tab:simu_ref_setting_W1_1}
    \definecolor{rowgray}{gray}{0.9}
    \setlength{\tabcolsep}{4pt}
    \renewcommand{\arraystretch}{1.3}
    \setlength{\heavyrulewidth}{1pt}
    \setlength{\lightrulewidth}{1pt}
    \makebox[\textwidth][c]{
    \begin{tabular}{lcccccccccc}
\toprule
  & \multicolumn{6}{c}{\textbf{Fidelity}} & \multicolumn{3}{c}{\textbf{Privacy}} & \textbf{Utility}  \\
\cmidrule(lr){2-7} \cmidrule(lr){8-10} \cmidrule(lr){11-11} 
\noalign{\vspace{-2pt}}
   & \multicolumn{2}{c}{\textbf{Corr. RMSE} $\downarrow$} 
    & \multirow{2}{*}{\textbf{Discrim.} $\downarrow$} 
    & \multirow{2}{*}{\textbf{KL $\lambda$} $\downarrow$} 
    & \multicolumn{2}{c}{\textbf{MMD} $\downarrow$} 
    & \multirow{2}{*}{\textbf{Dep.} $\downarrow$} 
    & \multirow{2}{*}{\textbf{MIR} $\downarrow$}  
    & \multirow{2}{*}{\textbf{NNAA} $\rightarrow 0$} 
    & \multirow{2}{*}{\textbf{Pred.} $\downarrow$} \\[-3pt]
\cmidrule(lr){2-3} \cmidrule(lr){6-7}
\noalign{\vspace{-2.8pt}}
  & \textbf{Long.} & \textbf{Static} & & & \textbf{Long.} & \textbf{Static} & & & & \\[-2pt]
\midrule
\rowcolor{white}
 Reference & $0.077$ & $0.000$ & $0.519$ & $0.282$ & $0.418$ & $0.010$ & $0.001$ & $0.052$ & $0.028$ & $2.904$ \\
\cmidrule(lr){1-11}
\rowcolor{rowgray}
 MultiNODEs & $\mathbf{0.105}$ & $0.000$ & $0.983$ & $67.611$ & $827.129$ & $0.013$ & $0.085$ & $\underline{0.049}$ & $-0.008$ & $11.267$ \\
\rowcolor{white}
 RTSGAN & $0.182$ & $0.000$ & $0.853$ & $3.609$ & $34.010$ & $0.023$ & $0.025$ & $0.051$ & $-0.013$ & $4.479$ \\
\rowcolor{rowgray}
 DGBFGP + HIVAE & $0.190$ & $0.000$ & $0.995$ & $67.611$ & $257.408$ & $0.009$ & $0.065$ & $0.049$ & $\mathbf{-0.000}$ & $11.738$ \\
% \cmidrule(lr){1-11}
\rowcolor{white}
 AS.3 MNSDE & $0.157$ & ${0.000}$ & $0.889$ & $\mathbf{0.567}$ & $7.498$ & $0.004$ & $0.026$ & $0.064$ & $-0.034$ & $\mathbf{3.916}$ \\
\rowcolor{rowgray}
 AS.3 SNSDE & $\underline{0.130}$ & ${0.000}$ & $\mathbf{0.789}$ & $\underline{0.718}$ & $5.656$ & $\mathbf{0.004}$ & $0.021$ & $0.066$ & $-0.043$ & $\underline{4.280}$ \\
\rowcolor{white}
 AS.3 FNSDE & $0.181$ & $0.000$ & $0.859$ & $0.720$ & $5.446$ & $0.020$ & $\mathbf{0.017}$ & $0.050$ & $\underline{0.004}$ & $5.403$ \\
% \cmidrule(lr){1-11}
\rowcolor{rowgray}
 HyperNSDE (post.) & $0.136$ & $0.000$ & $\underline{0.833}$ & ${0.791}$ & $\underline{5.144}$ & $\underline{0.004}$ & $\underline{0.019}$ & $0.055$ & $-0.026$ & $4.414$ \\
\rowcolor{white}
 HyperNSDE (prior) & $0.142$ & $0.000$ & $0.840$ & ${0.809}$ & $\mathbf{4.835}$ & $0.025$ & $0.020$ & $\mathbf{0.049}$ & ${-0.006}$ & $4.738$ \\
\bottomrule
    \end{tabular}}
    \end{table*}

\begin{table}[h]
\centering
\caption{\textbf{Real-vs-synthetic static-covariate fidelity on VELOUR for HyperNSDE}. For each covariate, the standardized mean difference (SMD; $<0.1$ indicates negligible differences) and the real-vs-synthetic test p-value are computed against the real cohort and summarised across the 50 generated cohorts (median [Q1, Q3]). High p-values indicate the synthetic covariate is statistically indistinguishable from the real one.}
\scriptsize
\setlength{\tabcolsep}{6pt}
\renewcommand{\arraystretch}{1.3}
\begin{tabular}{lcc}
\toprule
\textbf{Static covariate} & \textbf{SMD (median)} & \textbf{p-value (median [Q1, Q3])} \\
\midrule
SEX                     & $0.152$ & $0.334$ $[0.213, 0.582]$ \\
RACE                    & $0.186$ & $0.424$ $[0.184, 0.595]$ \\
REGION                  & $0.221$ & $0.635$ $[0.462, 0.889]$ \\
Age category            & $0.010$ & $0.594$ $[0.343, 0.812]$ \\
ECOG Performance Status & $0.136$ & $0.625$ $[0.352, 0.790]$ \\
Height                  & $0.013$ & $0.560$ $[0.422, 0.855]$ \\
\midrule
\multicolumn{3}{l}{\emph{Adverse-event and concomitant-medication indicators (20 covariates)}} \\
median across the $20$ & $0.091$ & $0.677$ $[0.41, 0.90]$ \\
\bottomrule
\end{tabular}
\vspace{2pt}
\\[2pt]
{\small Across all 26 covariates: median SMD $0.092$, with $65\%$ below $0.1$;
p-value quartiles [Q1, Q3] $= [0.40, 0.87]$.}
\end{table}

% \clearpage
\FloatBarrier

\subsection{Qualitative results}
\label{app:qualitative_results}

The following figures complement the quantitative analyses with trajectory-level comparisons across methods, generation modes, and ablation variants.

\begin{figure}[h]
    \centering
    \includegraphics[width=1\linewidth]{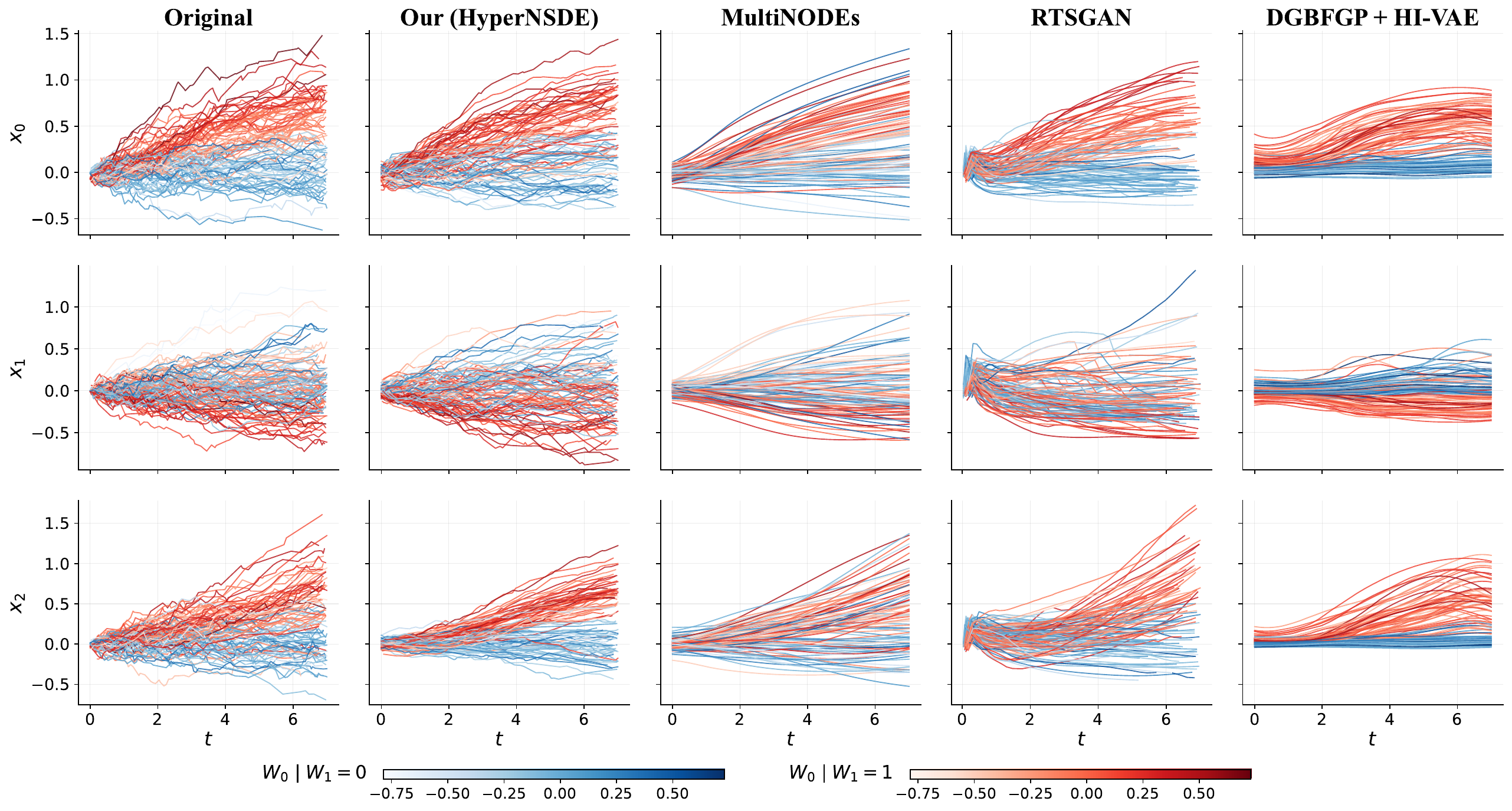}
    \caption{\textbf{Qualitative comparison of HyperNSDE and baselines on simulated data.} Rows correspond to the three longitudinal variables $x_0$, $x_1$, $x_2$; columns compare the original data against HyperNSDE and baselines. $y$-axis limits are shared across columns per row. The color encodes $W_0$ separately for subjects with $W_1=0$ (blue gradient) and $W_1=1$ (red gradient).}
    \label{app:fig_simu_ref_full}
\end{figure}

\begin{figure}[h]
    \centering
    \includegraphics[width=1\linewidth]{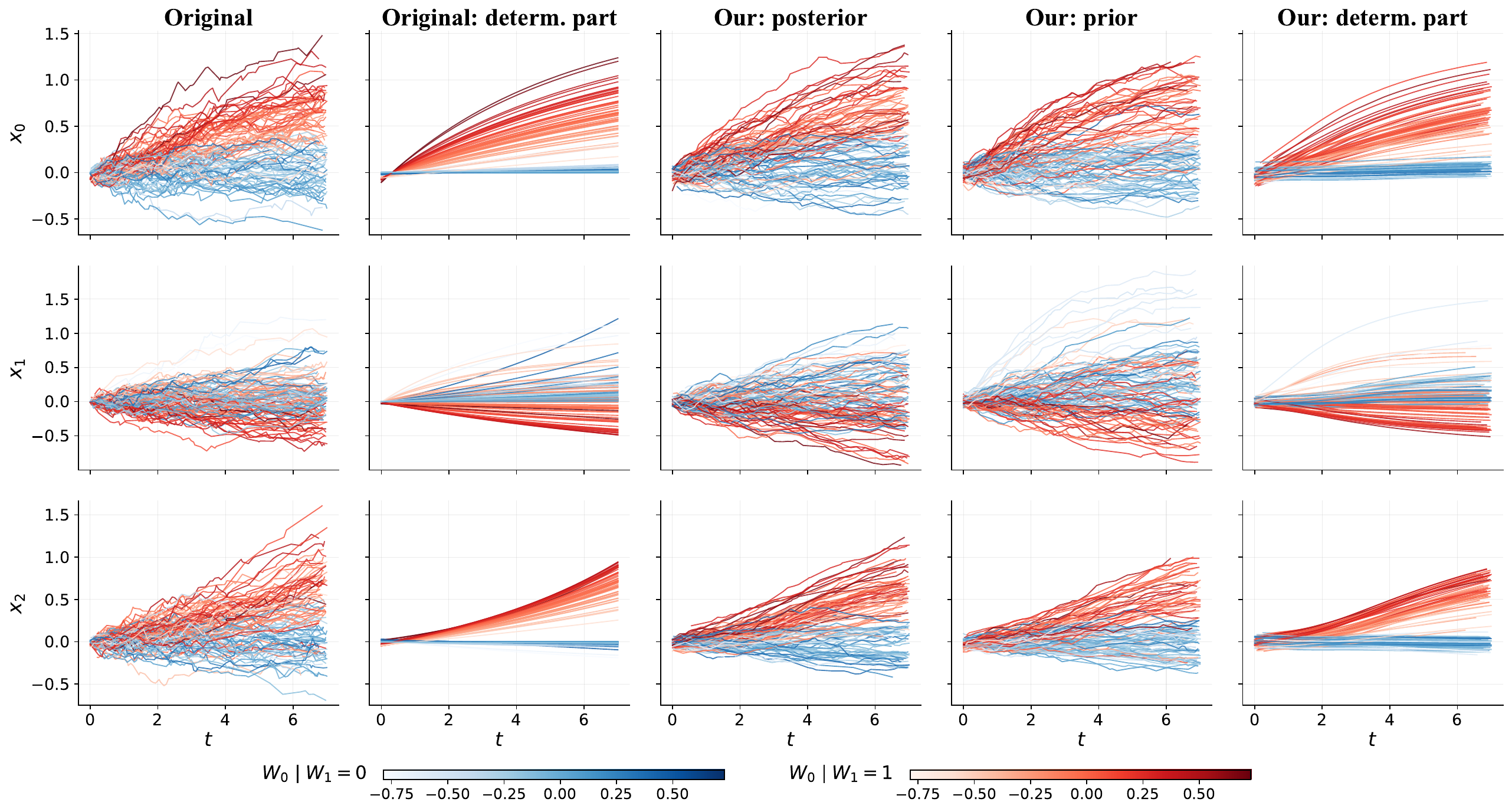}
    \caption{\textbf{Qualitative comparison of HyperNSDE generation modes on simulated data.} Rows correspond to the three longitudinal variables $x_0$, $x_1$, $x_2$; columns compare the original data and its deterministic component against HyperNSDE under posterior sampling, prior sampling, and estimated deterministic component. The color encodes $W_0$ separately for subjects with $W_1=0$ (blue gradient) and $W_1=1$ (red gradient).}
    \label{fig:simu_ref_setting_comp_our}
\end{figure}

% \begin{figure}[h]
\begin{sidewaysfigure}
    \centering
    \includegraphics[width=0.9\linewidth]{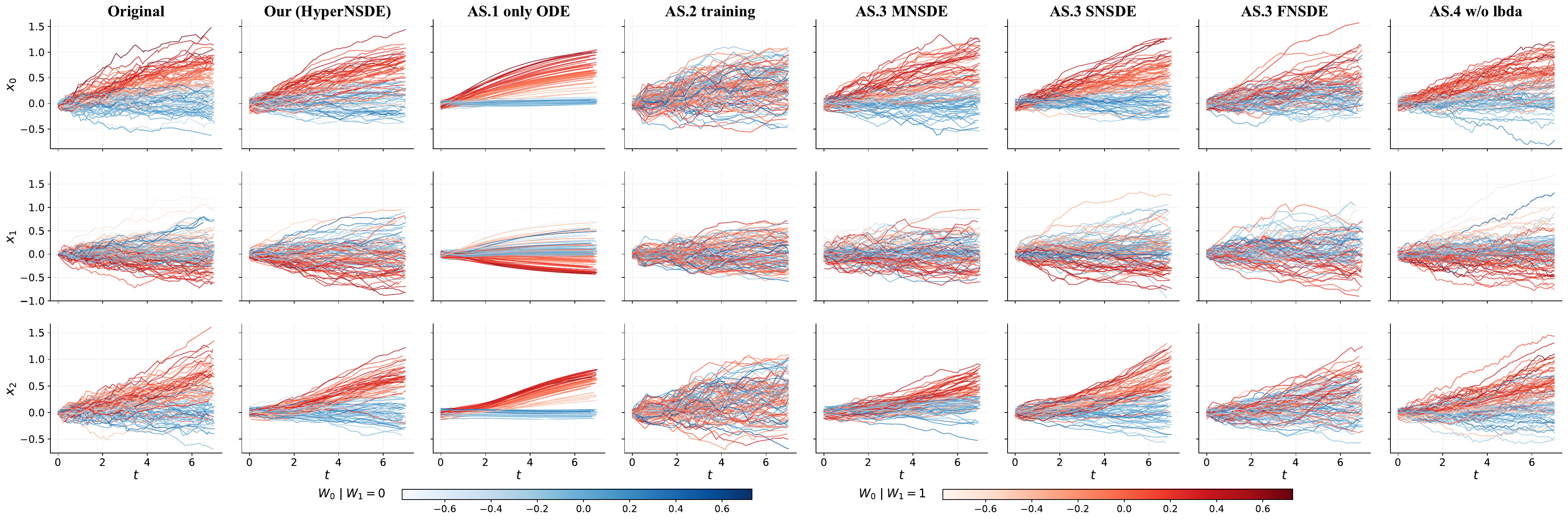}
    \caption{\textbf{Qualitative comparison of ablation study variants on simulated data.} Rows correspond to the three longitudinal variables $x_0$, $x_1$, $x_2$; columns compare the original data and HyperNSDE against the ablation variants described in Appendix~\ref{app:ablations}. The color encodes $W_0$ separately for subjects with $W_1=0$ (blue gradient) and $W_1=1$ (red gradient).}
    \label{fig:simu_ablation_studies}
\end{sidewaysfigure}
% \end{figure}

%%%%%%%%%%%%%%%%%%%%%%%%%%%%%%%%%%%%%%%%%%%%%%%%%%%%%%%%%%%%

\clearpage
\section*{NeurIPS Paper Checklist}

\begin{enumerate}

\item {\bf Claims}
    \item[] Question: Do the main claims made in the abstract and introduction accurately reflect the paper's contributions and scope?
    \item[] Answer: \answerYes{} % \answerTODO{} % Replace by \answerYes{}, \answerNo{}, or \answerNA{}.
    \item[] Justification: The abstract and introduction clearly state the three main contributions (unified generative model, hypernetwork-based personalization, tractable stochastic latent dynamics) and the evaluation scope across three dimensions: fidelity, utility, and privacy (simulated and real clinical datasets). 

\item {\bf Limitations}
    \item[] Question: Does the paper discuss the limitations of the work performed by the authors?
    \item[] Answer: \answerYes{} % Replace by \answerYes{}, \answerNo{}, or \answerNA{}.
    \item[] Justification: Limitations are discussed in the Conclusion (Section~\ref{sec:conclusion}), covering computational cost, scope of the empirical evaluation on real-world datasets, and the limitations of current evaluation metrics for irregular multimodal clinical data.

\item {\bf Theory assumptions and proofs}
    \item[] Question: For each theoretical result, does the paper provide the full set of assumptions and a complete (and correct) proof?
    \item[] Answer: \answerNA{} % Replace by \answerYes{}, \answerNo{}, or \answerNA{}.
    \item[] Justification: The paper does not include formal theoretical results. The deterministic--stochastic decomposition is discussed as a modeling choice with theoretical motivation provided in Appendix~\ref{app:discussion_splitSDEtrain}, but no theorems or proofs are claimed.

    \item {\bf Experimental result reproducibility}
    \item[] Question: Does the paper fully disclose all the information needed to reproduce the main experimental results of the paper to the extent that it affects the main claims and/or conclusions of the paper (regardless of whether the code and data are provided or not)?
    \item[] Answer: \answerYes{} % Replace by \answerYes{}, \answerNo{}, or \answerNA{}.
    \item[] Justification: The model architecture, training procedure, evaluation protocol ($N_{\text{MC}}=30$, $N_{\text{gen}}=50$, 80/20 splits), hyperparameter tuning strategy, model of the simulated datasets, and dataset preprocessing pipelines are described in the main paper and Appendices~\ref{app:training_hyperopt},\ref{app:datasets},\ref{app:exp_setup}. Code is publicly available (Section~\ref{sec:experimental_setup}).

\item {\bf Open access to data and code}
    \item[] Question: Does the paper provide open access to the data and code, with sufficient instructions to faithfully reproduce the main experimental results, as described in supplemental material?
    \item[] Answer: \answerYes{} % Replace by \answerYes{}, \answerNo{}, or \answerNA{}.
    \item[] Justification: Code is released at \url{https://github.com/perrinechassat/HyperNSDE}, with scripts to reproduce simulated experiments. Real datasets require individual data access agreements: VELOUR is available via Project DataSphere~\cite{projectdatasphere} and PPMI via \url{https://www.ppmi-info.org/}; preprocessing details are provided in Appendix~\ref{app:realworld_datasets}.

\item {\bf Experimental setting/details}
    \item[] Question: Does the paper specify all the training and test details (e.g., data splits, hyperparameters, how they were chosen, type of optimizer) necessary to understand the results?
    \item[] Answer: \answerYes{} % Replace by \answerYes{}, \answerNo{}, or \answerNA{}.
    \item[] Justification: Data splits, the number of Monte Carlo runs, generation repetitions, optimizer, early stopping, and hyperparameter selection are described in Section~\ref{sec:experimental_setup} and Appendix~\ref{app:training_hyperopt}--\ref{app:exp_setup}. 

\item {\bf Experiment statistical significance}
    \item[] Question: Does the paper report error bars suitably and correctly defined or other appropriate information about the statistical significance of the experiments?
    \item[] Answer: \answerYes{} % Replace by \answerYes{}, \answerNo{}, or \answerNA{}.
    \item[] Justification:  Results are averaged over $N_{\text{MC}}=30$ Monte Carlo runs (independently generated datasets with different random seeds) and $N_{\text{gen}}=50$ independent synthetic datasets per trained model for simulated data, and over $N_{\text{gen}}=50$ generations for real-world datasets. Standard deviations across these runs are reported in Tables~\ref{tab:all_results_std} and~\ref{tab:ablation_studies_std} in Appendix~\ref{app:additional_results}. No normality assumption is made; standard deviations are reported as a measure of variability across runs.

\item {\bf Experiments compute resources}
    \item[] Question: For each experiment, does the paper provide sufficient information on the computer resources (type of compute workers, memory, time of execution) needed to reproduce the experiments?
    \item[] Answer: \answerYes{}  % Replace by \answerYes{}, \answerNo{}, or \answerNA{}.
    \item[] Justification: The Appendix~\ref{app:exp_setting_resources} reports the main compute resources used for the experiments, including single-GPU jobs on NVIDIA V100 32\,GB GPUs on the Jean Zay cluster, with 10 CPU cores per task and a typical 20-hour wall-clock allocation. We also provide indicative training times on the reference simulated setting and note how runtime scales with the number of patients and observations. While we do not provide a full accounting of all preliminary or failed runs, the paper gives sufficient information to reproduce the reported experiments in practice.
    
\item {\bf Code of ethics}
    \item[] Question: Does the research conducted in the paper conform, in every respect, with the NeurIPS Code of Ethics \url{https://neurips.cc/public/EthicsGuidelines}?
    \item[] Answer: \answerYes{} % Replace by \answerYes{}, \answerNo{}, or \answerNA{}.
    \item[] Justification: To the best of our knowledge, the research presented in this paper conforms to the NeurIPS Code of Ethics. The work addresses privacy-preserving synthetic data generation, aligned with responsible data sharing in healthcare. All datasets are accessed through official channels under their respective terms of use. No patient-level data is released.

\item {\bf Broader impacts}
    \item[] Question: Does the paper discuss both potential positive societal impacts and negative societal impacts of the work performed?
    \item[] Answer: \answerYes{} % Replace by \answerYes{}, \answerNo{}, or \answerNA{}.
    \item[] Justification: The paper discusses both positive and negative societal impacts in the conclusion (Section~\ref{sec:conclusion}). On the positive side, synthetic clinical data may support privacy-aware data sharing, benchmarking, and method development in healthcare. On the negative side, the paper explicitly notes the current limits of realism and the risks of premature use for high-stakes applications such as cohort augmentation or clinical trial simulation. It also discusses privacy-related evaluation and the need for stronger safeguards such as differential privacy in future work.
    
\item {\bf Safeguards}
    \item[] Question: Does the paper describe safeguards that have been put in place for responsible release of data or models that have a high risk for misuse (e.g., pre-trained language models, image generators, or scraped datasets)?
    \item[] Answer: \answerNA{} % Replace by \answerYes{}, \answerNo{}, or \answerNA{}.
    \item[] Justification: The contribution is methodological, and the released asset is limited to research code for reproducibility.

\item {\bf Licenses for existing assets}
    \item[] Question: Are the creators or original owners of assets (e.g., code, data, models), used in the paper, properly credited and are the license and terms of use explicitly mentioned and properly respected?
    \item[] Answer: \answerYes{} % Replace by \answerYes{}, \answerNo{}, or \answerNA{}.
    \item[] Justification: All datasets are properly cited with their respective access conditions. Baseline implementations are cited and used under their respective open-source 
    licenses. The \texttt{pySigLib} package~\cite{shmelev2025pysiglib} is cited and used under its open-source license.

\item {\bf New assets}
    \item[] Question: Are new assets introduced in the paper well documented and is the documentation provided alongside the assets?
    \item[] Answer: \answerYes{} % Replace by \answerYes{}, \answerNo{}, or \answerNA{}.
    \item[] Justification: The paper introduces new research code implementing HyperNSDE. The manuscript and appendix describe the main modeling components, training setup, and implementation details needed to understand and reproduce the method.

\item {\bf Crowdsourcing and research with human subjects}
    \item[] Question: For crowdsourcing experiments and research with human subjects, does the paper include the full text of instructions given to participants and screenshots, if applicable, as well as details about compensation (if any)? 
    \item[] Answer: \answerNA{} % Replace by \answerYes{}, \answerNo{}, or \answerNA{}.
    \item[] Justification: The paper does not involve crowdsourcing experiments or research with human subjects conducted by the authors.

\item {\bf Institutional review board (IRB) approvals or equivalent for research with human subjects}
    \item[] Question: Does the paper describe potential risks incurred by study participants, whether such risks were disclosed to the subjects, and whether Institutional Review Board (IRB) approvals (or an equivalent approval/review based on the requirements of your country or institution) were obtained?
    \item[] Answer: \answerNA{} % Replace by \answerYes{}, \answerNo{}, or \answerNA{}.
    \item[] Justification: The paper does not involve crowdsourcing experiments or human-subject research conducted by the authors, and therefore no IRB approval or equivalent review was required for the work presented here.

\item {\bf Declaration of LLM usage}
    \item[] Question: Does the paper describe the usage of LLMs if it is an important, original, or non-standard component of the core methods in this research? Note that if the LLM is used only for writing, editing, or formatting purposes and does \emph{not} impact the core methodology, scientific rigor, or originality of the research, declaration is not required.
    %this research? 
    \item[] Answer: \answerNA{} % Replace by \answerYes{}, \answerNo{}, or \answerNA{}.
    \item[] Justification: LLMs are not used as an important, original, or non-standard component of the proposed method, experiments, or scientific claims.

\end{enumerate}

\end{document}